\documentclass[sigconf, screen]{acmart}
\AtBeginDocument{%
  }

\renewcommand\footnotetextcopyrightpermission[1]{} 
 \setcopyright{none}

\usepackage{hyperref}
\usepackage{graphicx}
\usepackage{multicol}
\usepackage{makecell}
\usepackage{multirow}
\usepackage{listings}
\usepackage{color}
\usepackage{pifont}
\usepackage{xcolor}
\usepackage{algorithm}
\usepackage{algorithmic}
\usepackage{booktabs}
\usepackage{amsmath}
\usepackage{booktabs}
\usepackage{bbm}
\usepackage{subcaption}
\definecolor{SeaGreen}{rgb}{0.18, 0.55, 0.34}
\begin{document}

\title{How Much To Guide: Revisiting Adaptive Guidance in Classifier-Free Guidance Text-to-Vision Diffusion Models}

\author{Huixuan Zhang}
\affiliation{%
  \institution{Wangxuan Institute of Computer Technology, Peking University}
  \city{Beijing}
  \country{China}}
\email{zhanghuixuan@stu.pku.edu.cn}

\author{Junzhe Zhang}
\affiliation{%
  \institution{Wangxuan Institute of Computer Technology, Peking University}
  \city{Beijing}
  \country{China}}
\email{junzhezhang@stu.pku.edu.cn}

\author{Xiaojun Wan$^{*}$}
\affiliation{%
  \institution{Wangxuan Institute of Computer Technology, Peking University}
  \city{Beijing}
  \country{China}}
\email{wanxiaojun@pku.edu.cn}


\begin{abstract}
  With the rapid development of text-to-vision generation diffusion models, classifier-free guidance has emerged as the most prevalent method for conditioning. However, this approach inherently requires twice as many steps for model forwarding compared to unconditional generation, resulting in significantly higher costs. While previous study has introduced the concept of adaptive guidance, it lacks solid analysis and empirical results, making previous method unable to be applied to general diffusion models. In this work, we present another perspective of applying adaptive guidance and propose \textit{Step AG}, which is a simple, universally applicable adaptive guidance strategy. Our evaluations focus on both image quality and image-text alignment. whose results indicate that restricting classifier-free guidance to the first several denoising steps is sufficient for generating high-quality, well-conditioned images, achieving an average speedup of 20\% to 30\%. Such improvement is consistent across different settings such as inference steps, and various models including video generation models, highlighting the superiority of our method.
\end{abstract}


\begin{teaserfigure}
  \begin{subfigure}[b]{0.095\textwidth}
        \includegraphics[width=1.0\textwidth]{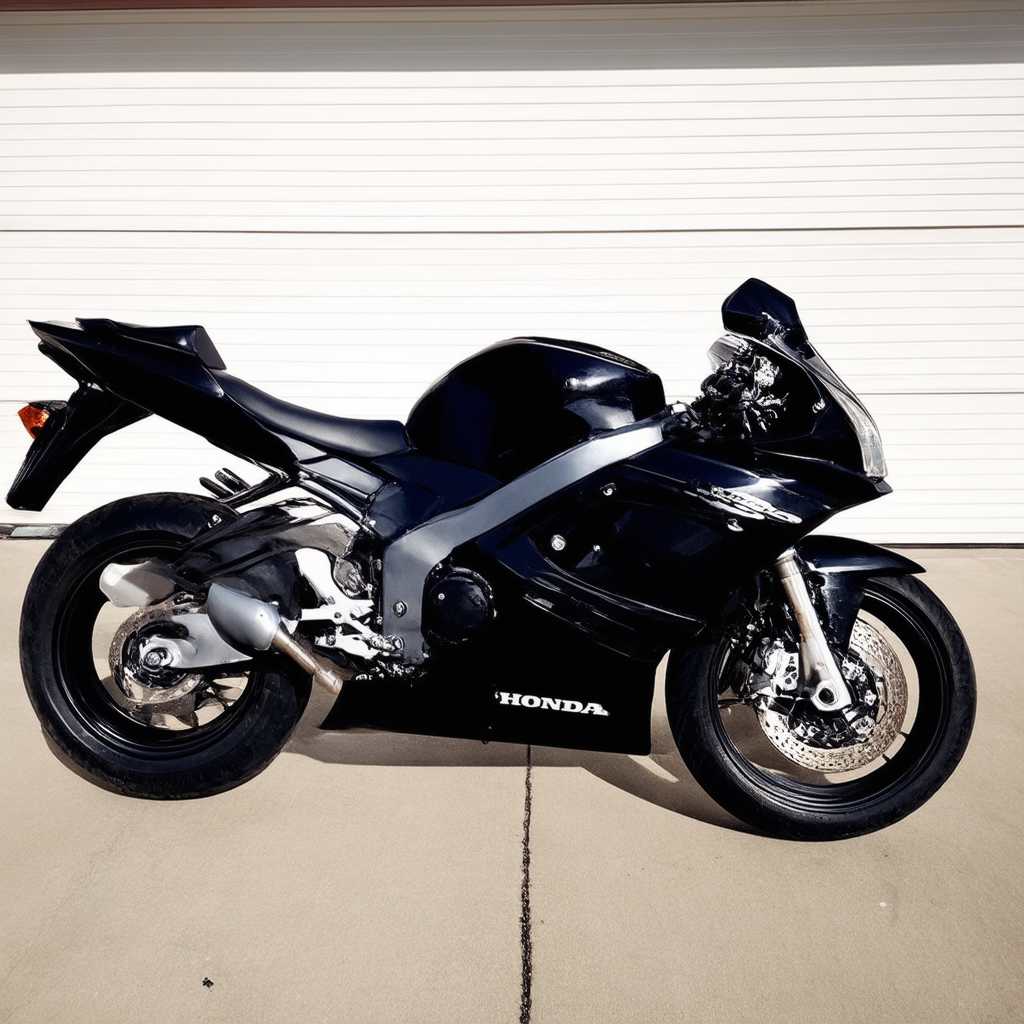}
    \end{subfigure}
    \begin{subfigure}[b]{0.095\textwidth}
        \includegraphics[width=1.0\textwidth]{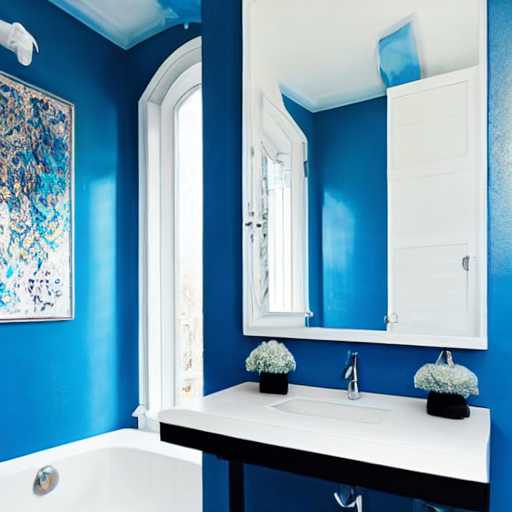}
    \end{subfigure}
     \begin{subfigure}[b]{0.095\textwidth}
        \includegraphics[width=1.0\textwidth]{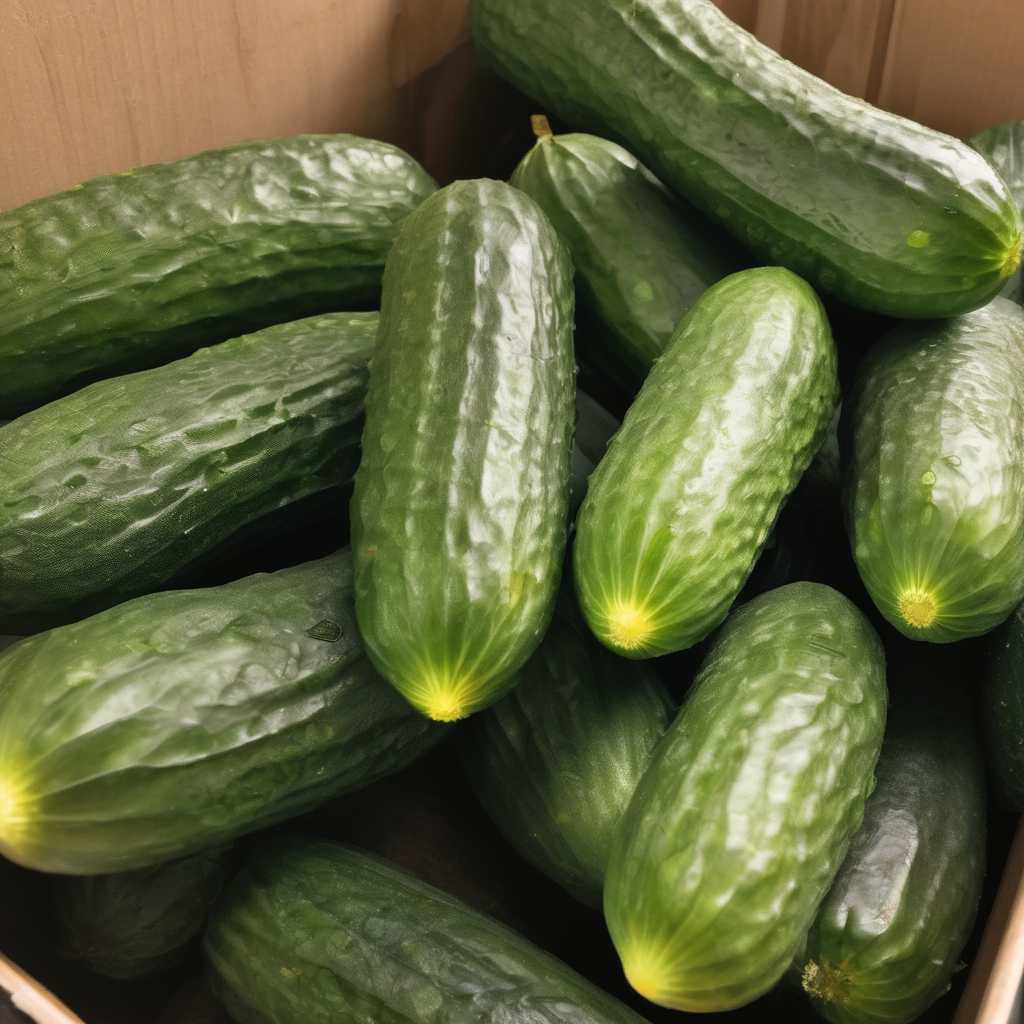}
    \end{subfigure}   
    \begin{subfigure}[b]{0.095\textwidth}
        \includegraphics[width=1.0\textwidth]{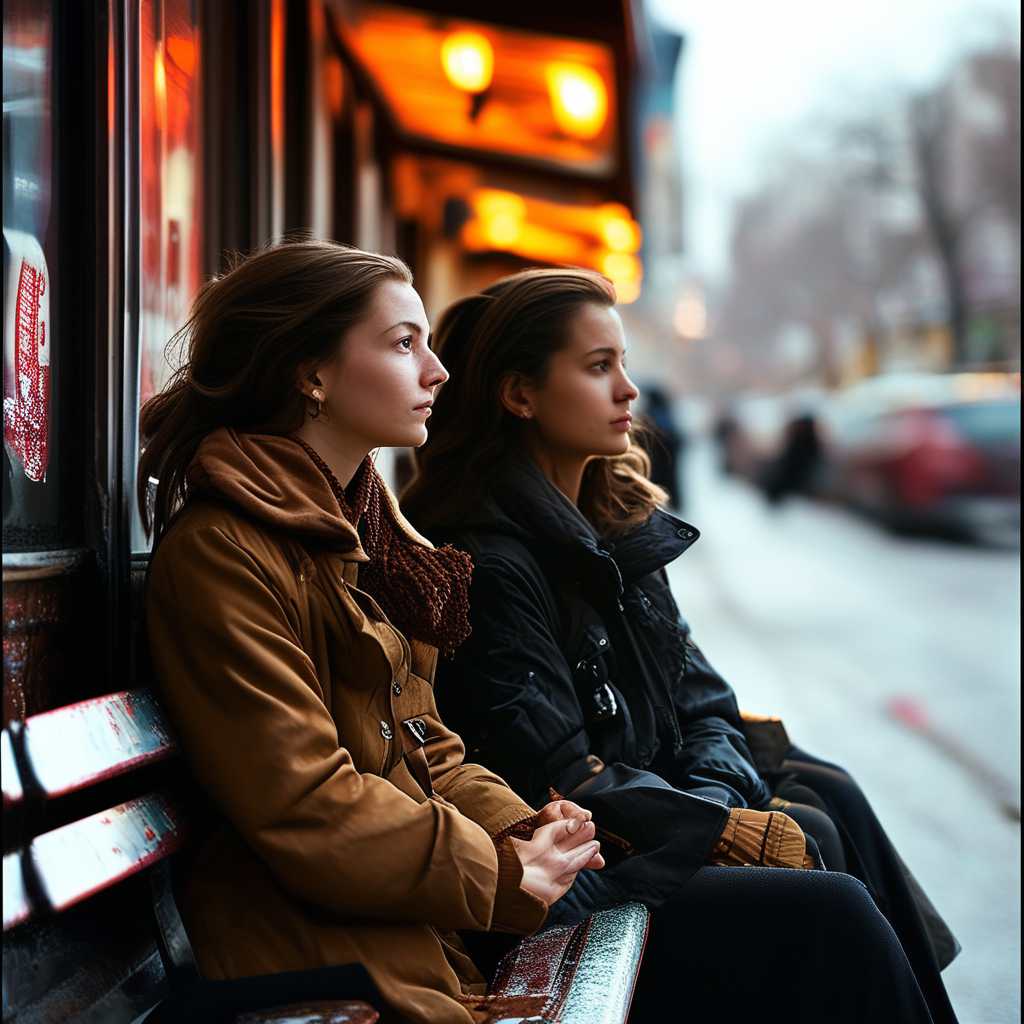}
    \end{subfigure}   
    \begin{subfigure}[b]{0.095\textwidth}
        \includegraphics[width=1.0\textwidth]{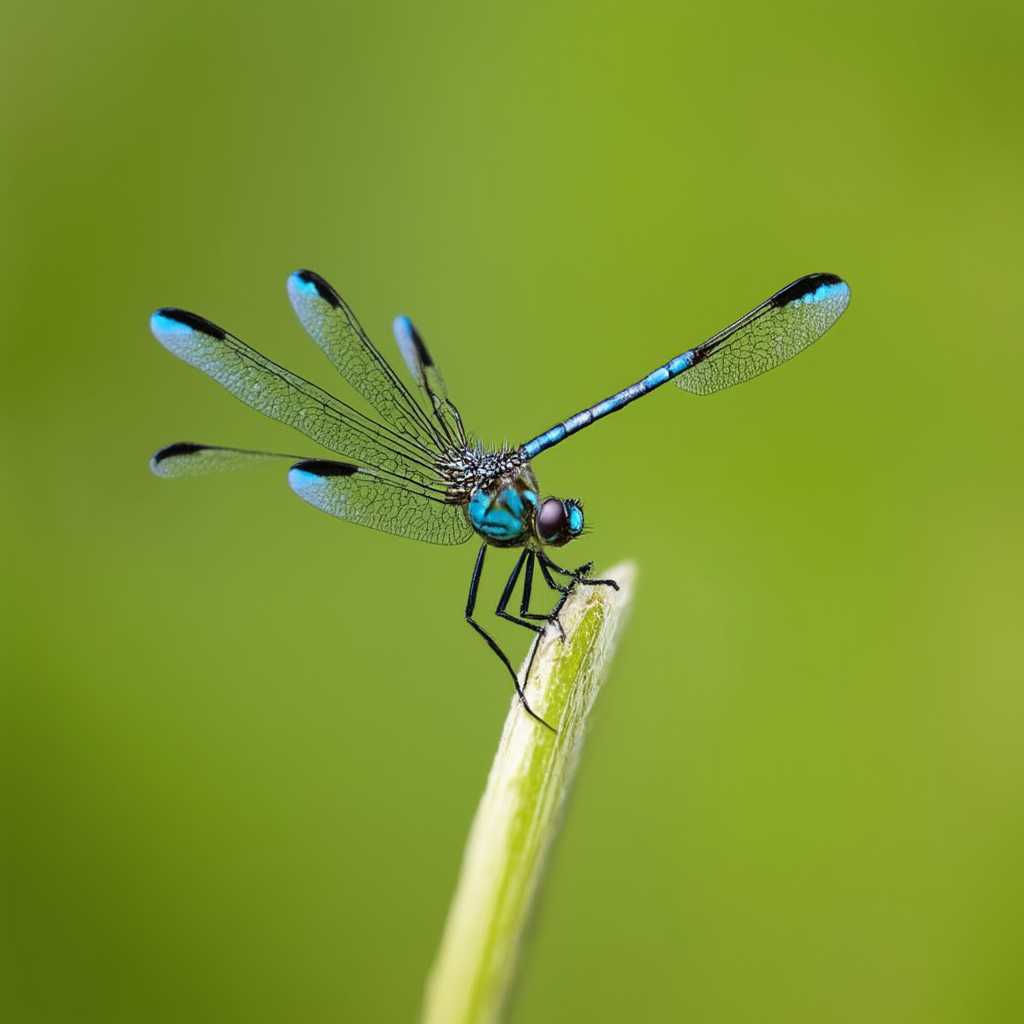}
    \end{subfigure}
    \begin{subfigure}[b]{0.095\textwidth}
        \includegraphics[width=1.0\textwidth]{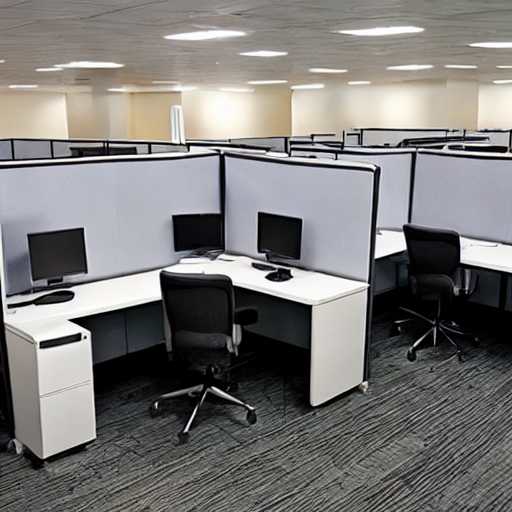}
    \end{subfigure}
    \begin{subfigure}[b]{0.095\textwidth}
        \includegraphics[width=1.0\textwidth]{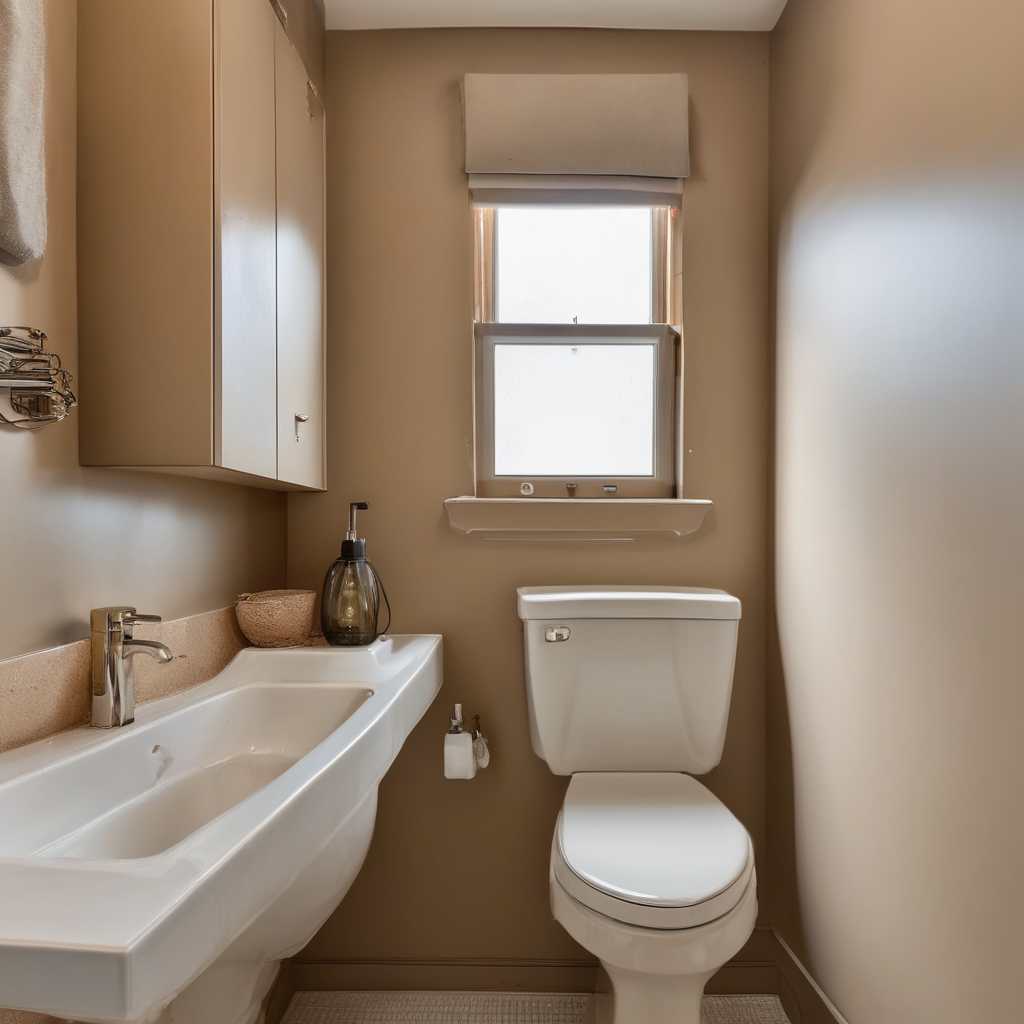}
    \end{subfigure}
    \begin{subfigure}[b]{0.095\textwidth}
        \includegraphics[width=1.0\textwidth]{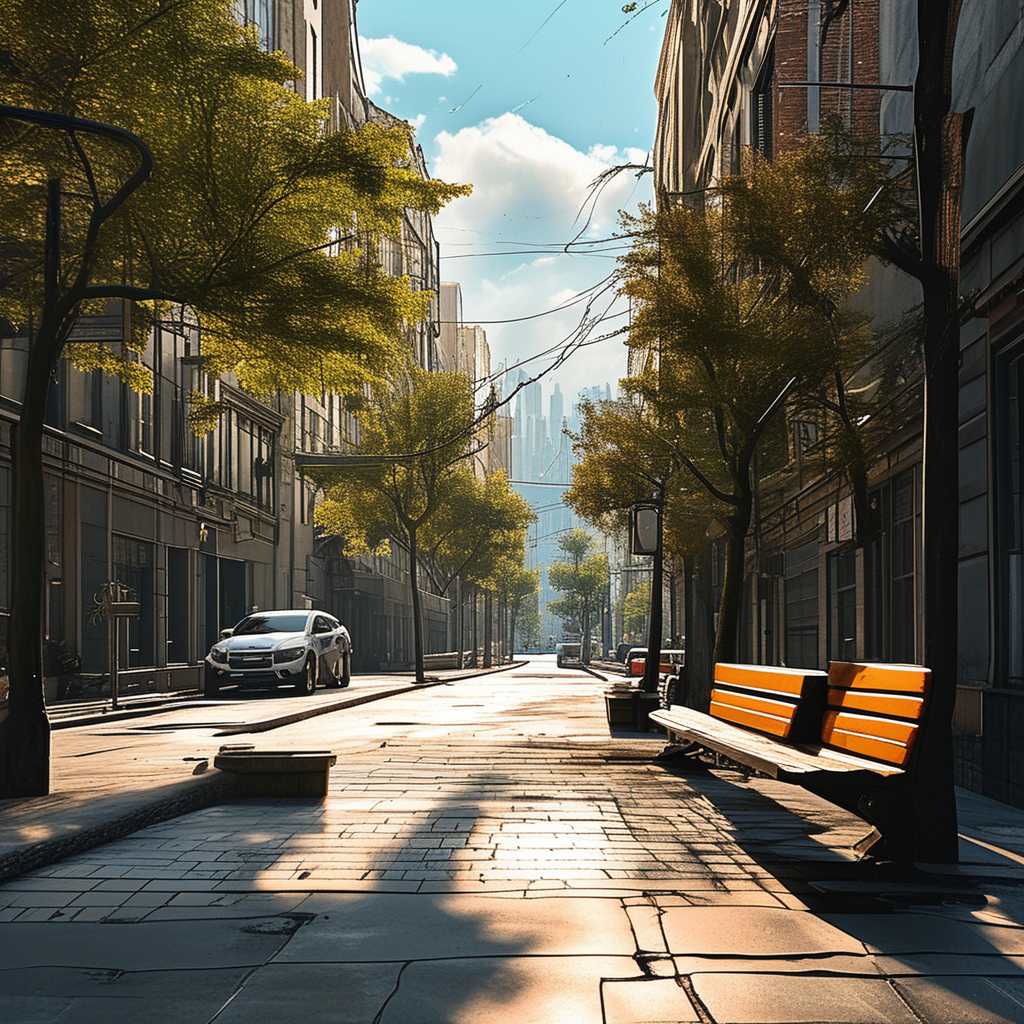}
    \end{subfigure}
    \begin{subfigure}[b]{0.095\textwidth}
        \includegraphics[width=1.0\textwidth]{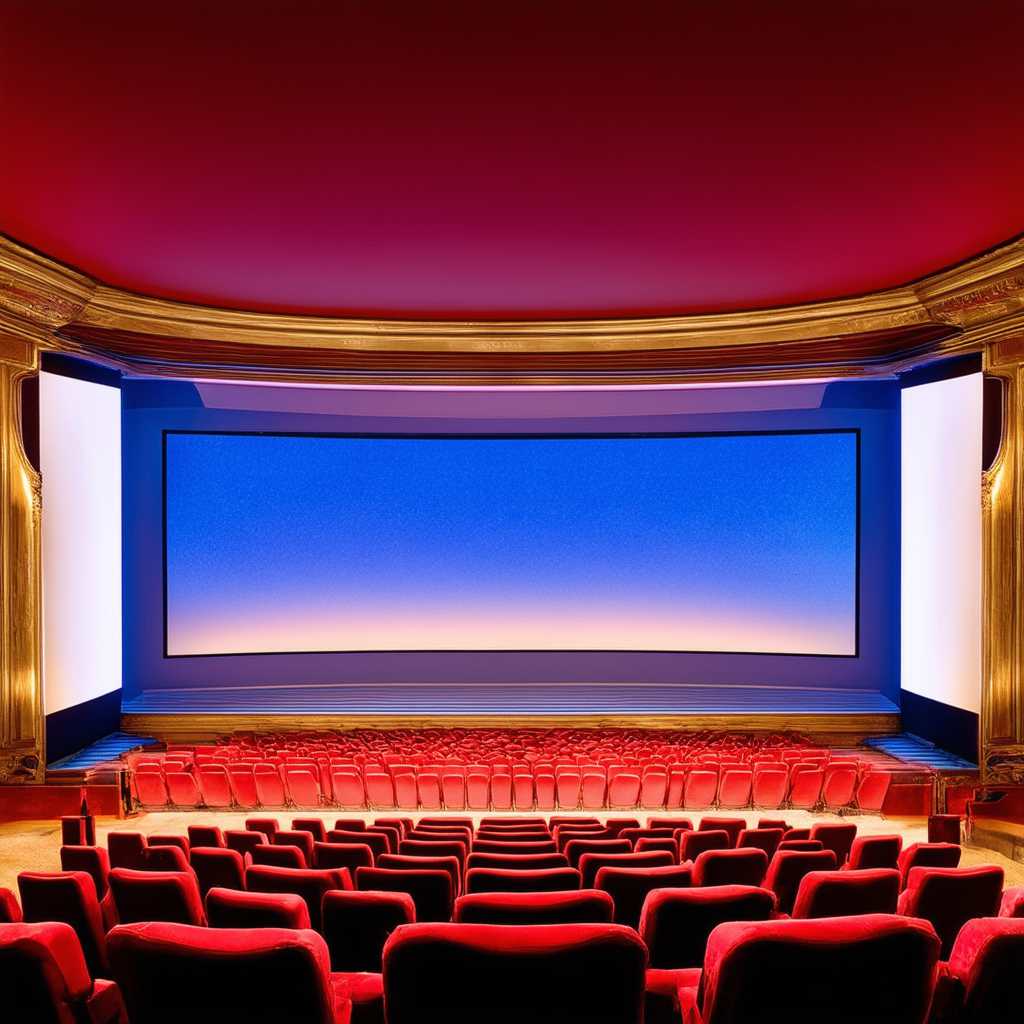}
    \end{subfigure}
    \begin{subfigure}[b]{0.095\textwidth}
        \includegraphics[width=1.0\textwidth]{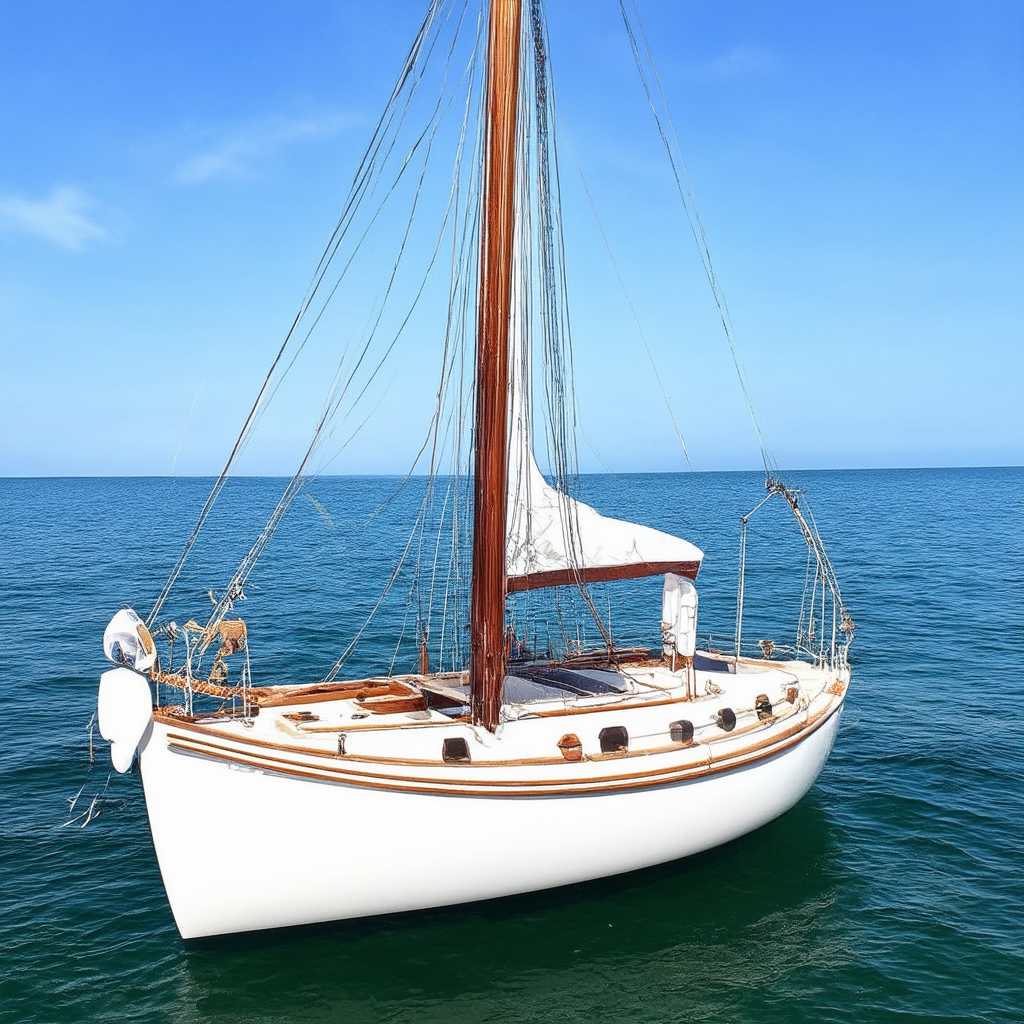}
    \end{subfigure}
    \vfill
    \begin{subfigure}[b]{0.095\textwidth}
        \includegraphics[width=1.0\textwidth]{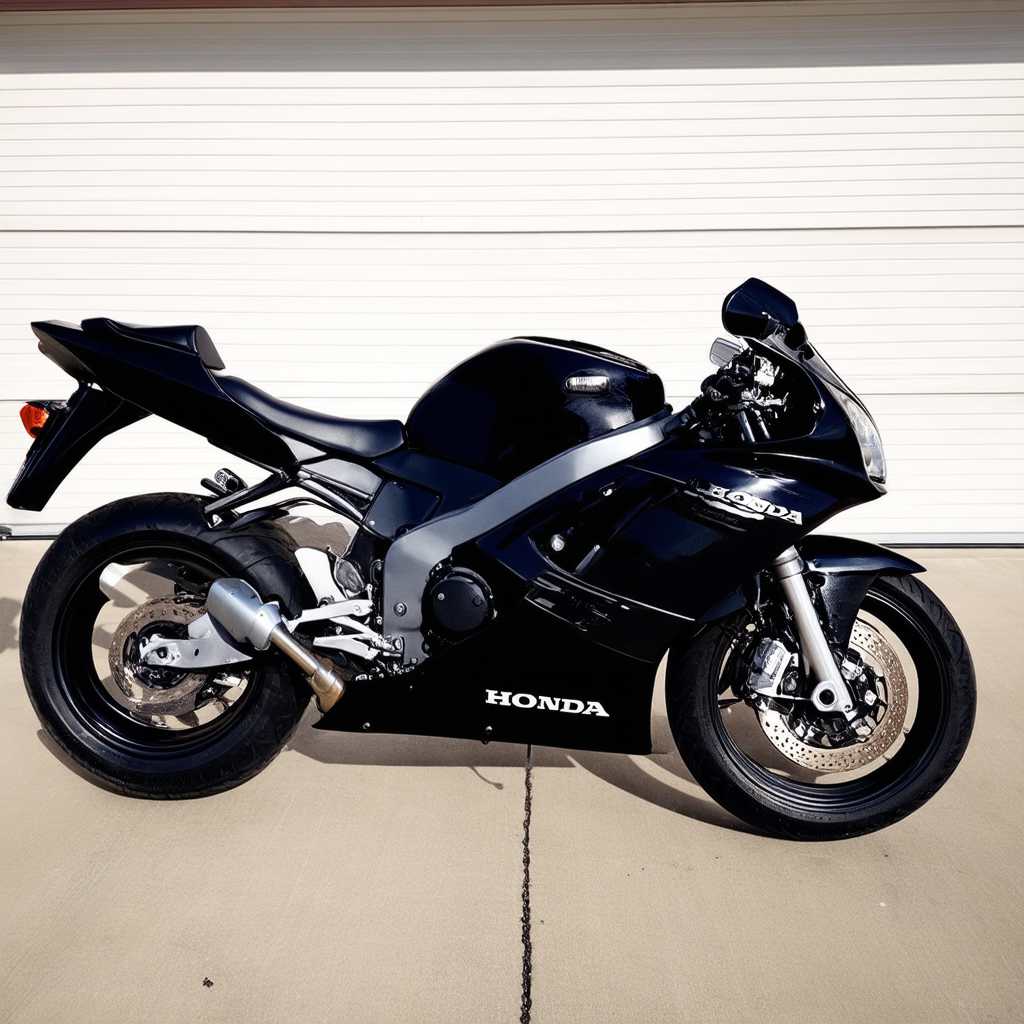}
    \end{subfigure}
    \begin{subfigure}[b]{0.095\textwidth}
        \includegraphics[width=1.0\textwidth]{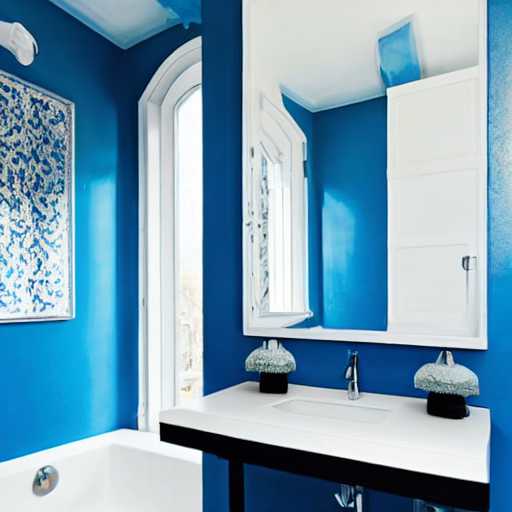}
    \end{subfigure}
     \begin{subfigure}[b]{0.095\textwidth}
        \includegraphics[width=1.0\textwidth]{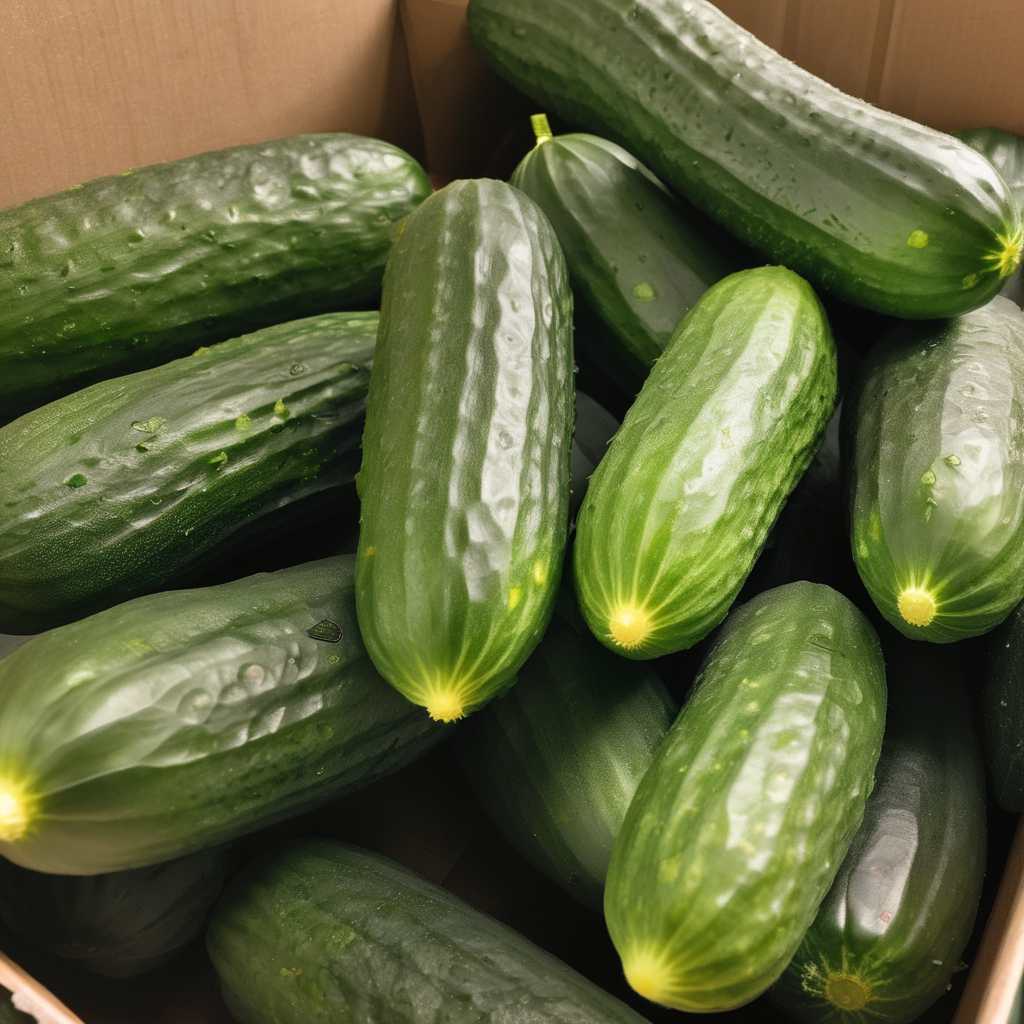}
    \end{subfigure}   
    \begin{subfigure}[b]{0.095\textwidth}
        \includegraphics[width=1.0\textwidth]{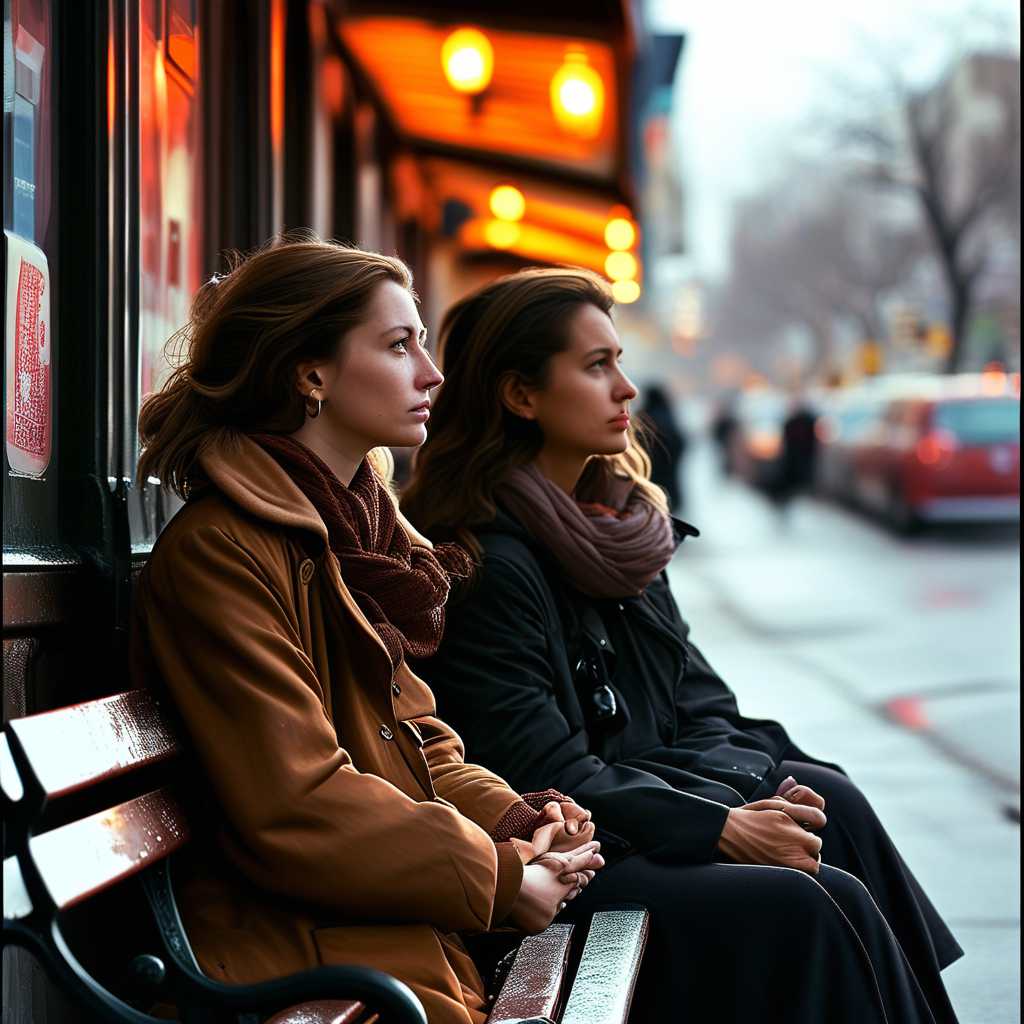}
    \end{subfigure} 
    \begin{subfigure}[b]{0.095\textwidth}
        \includegraphics[width=1.0\textwidth]{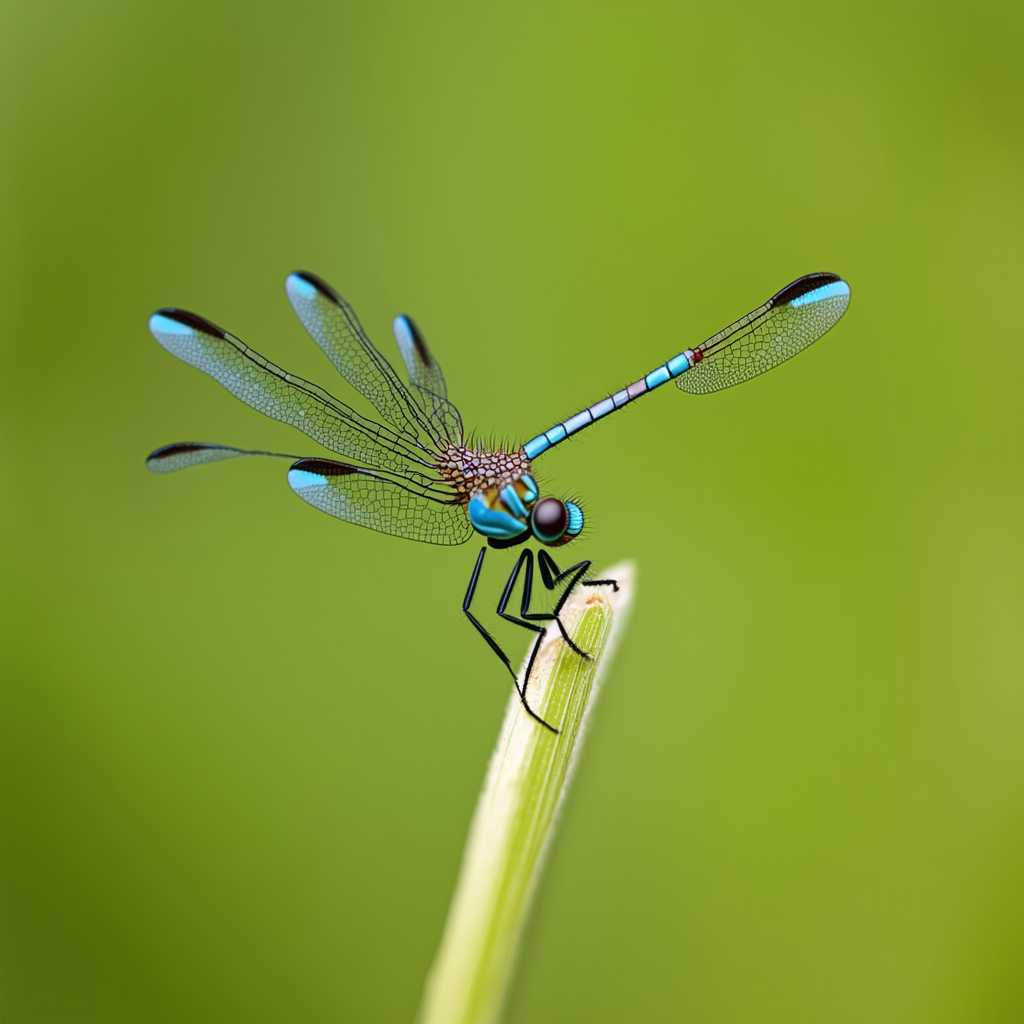}
    \end{subfigure}
    \begin{subfigure}[b]{0.095\textwidth}
        \includegraphics[width=1.0\textwidth]{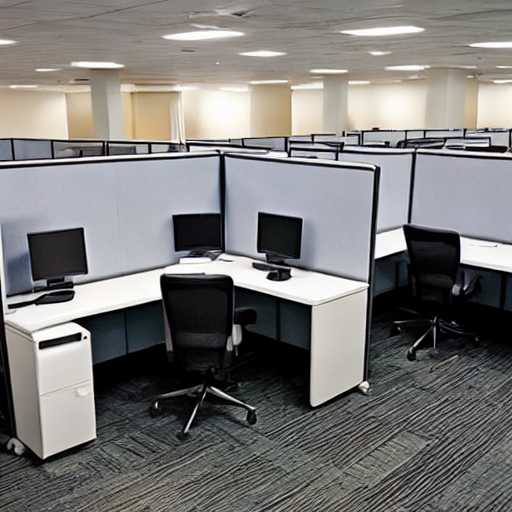}
    \end{subfigure}
    \begin{subfigure}[b]{0.095\textwidth}
        \includegraphics[width=1.0\textwidth]{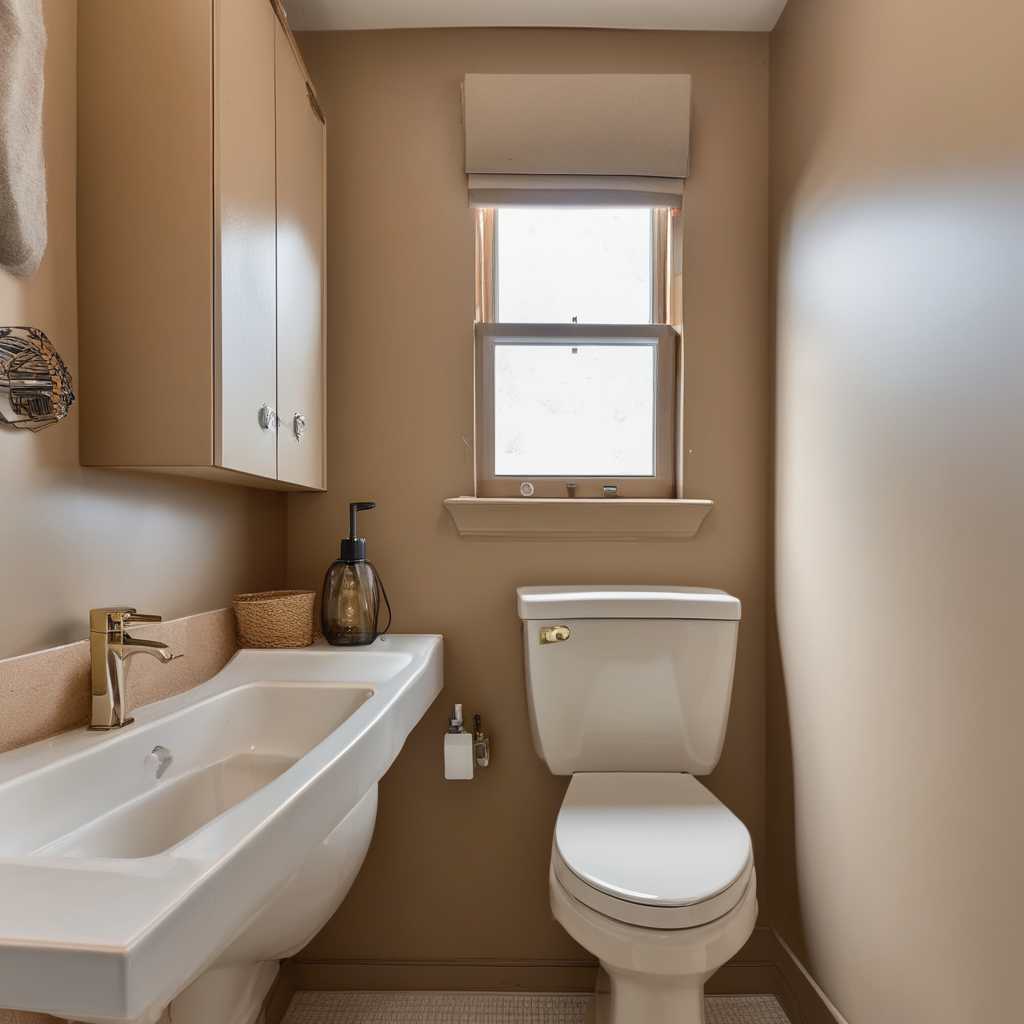}
    \end{subfigure}
    \begin{subfigure}[b]{0.095\textwidth}
        \includegraphics[width=1.0\textwidth]{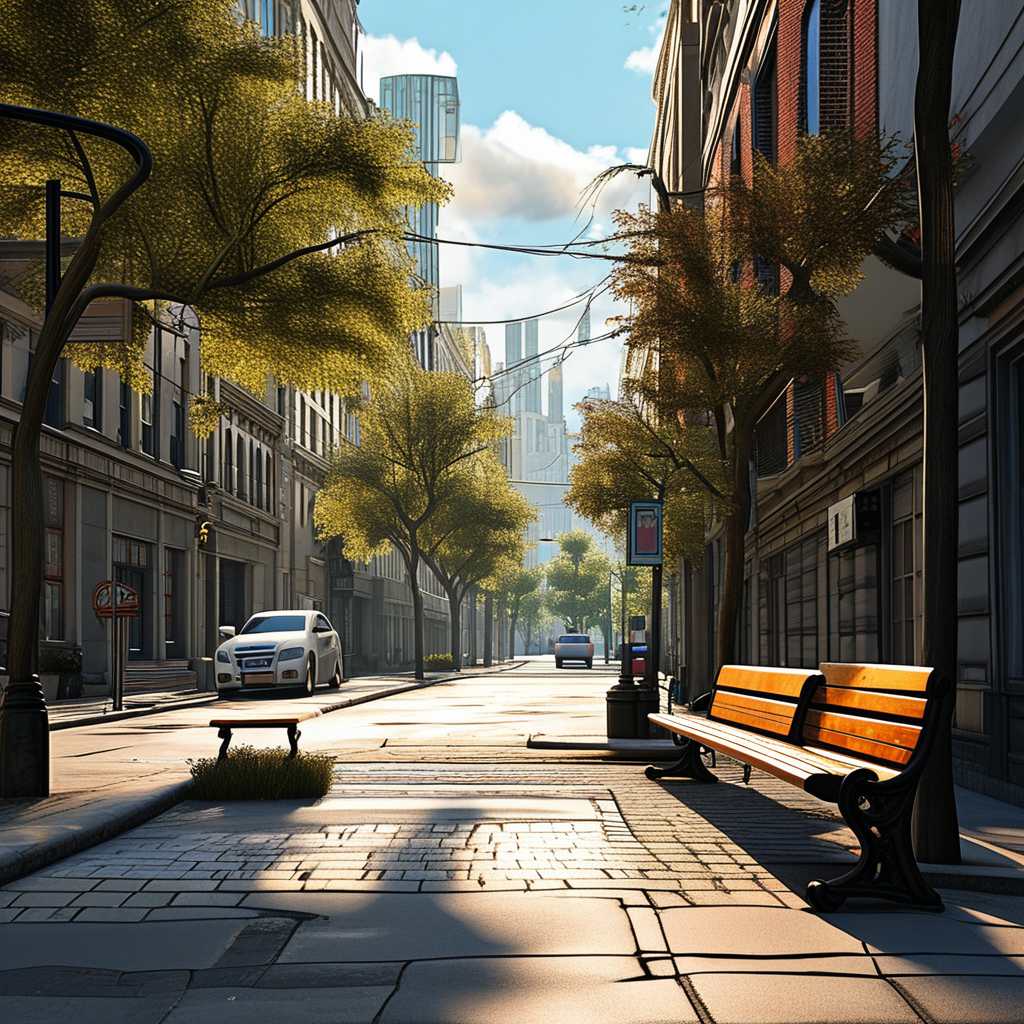}
    \end{subfigure}
    \begin{subfigure}[b]{0.095\textwidth}
        \includegraphics[width=1.0\textwidth]{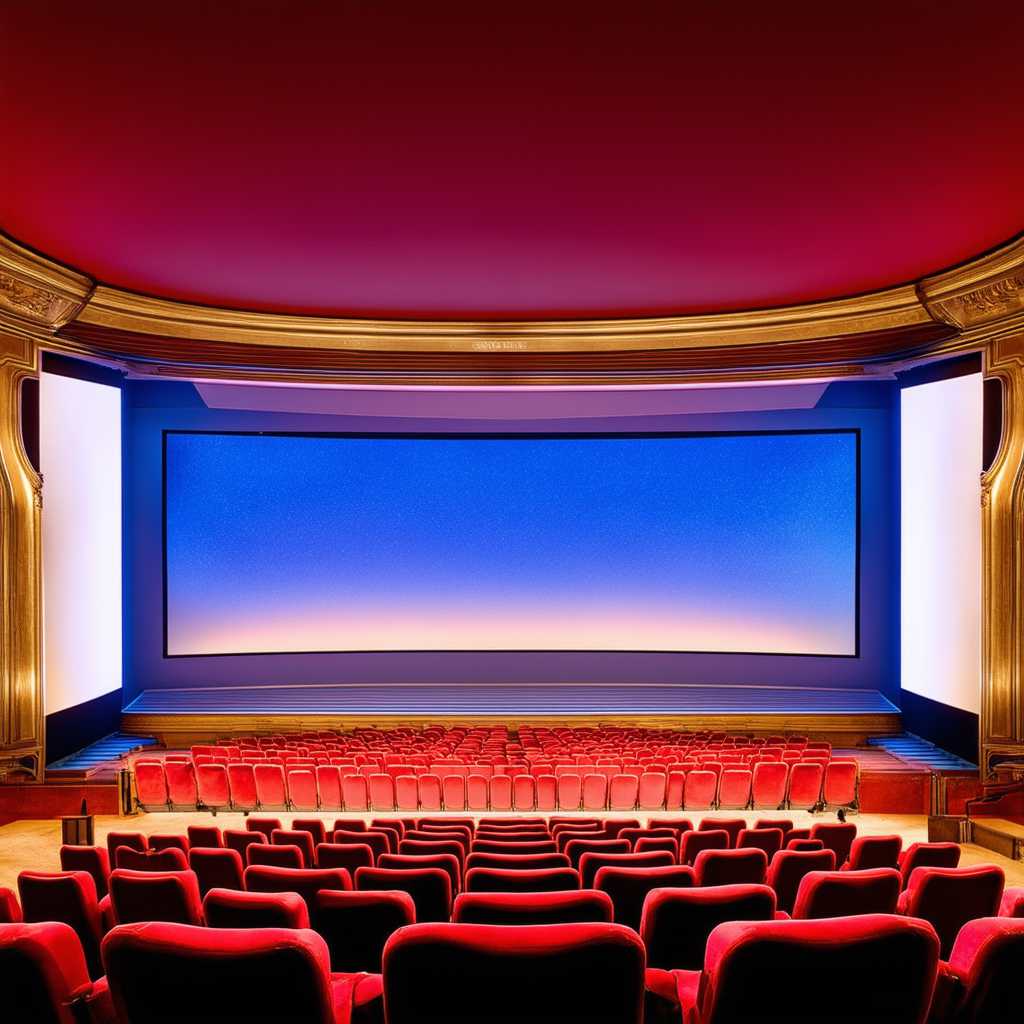}
    \end{subfigure}
    \begin{subfigure}[b]{0.095\textwidth}
        \includegraphics[width=1.0\textwidth]{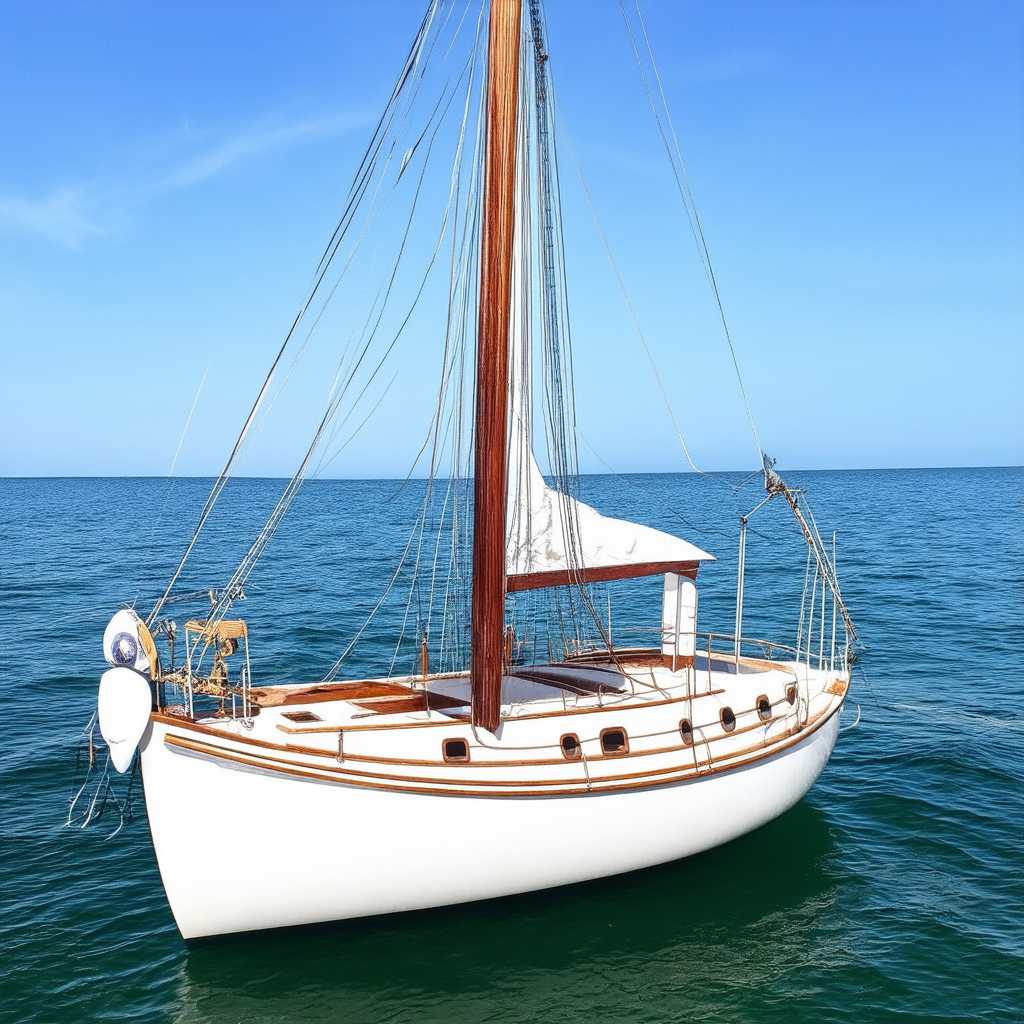}
    \end{subfigure}
    \caption{Generated examples of models with and w/o \textit{Step AG}. The first line is generated by models with \textit{Step AG} while the second line is generated by models w/o \textit{Step AG}.}
    \label{fig:teaser}
\end{teaserfigure}

\maketitle

\section{Introduction}
Generative diffusion models \cite{NEURIPS2020_DDPM, song2021DDIM} have been widely used in many generative tasks. They can be applied to image generation \cite{Rombach_2022_CVPR, NEURIPS2022_Imagen, podell2023sdxl, esser2024scaling,chen2024pixartsigma}, video generation \cite{NEURIPS2022_VideoDiff, luo2023video, ho2022imagen, wang2023modelscope, yang2024cogvideox}, audio generation \cite{kong2020diffwave, popov2021grad}, etc. Specifically, conditioning is a critical aspect of generative tasks, as it ensures that the generated content aligns with a specified condition. The most widely used conditioning method in diffusion models is classifier-free guidance \cite{ho2022classifier}, which integrates predictions from both conditional and unconditional models to achieve stable guidance.

However, classifier-free guidance has a fundamental drawback: it necessitates forwarding the model twice at each denoising step, resulting in costly inference. Moreover, as diffusion models typically require multiple denoising steps and the denoising model continues to scale up \cite{esser2024scaling, yang2024cogvideox}, this cost becomes an increasingly significant issue that harms efficiency.

Many studies have been undertaken to address this issue. Distillation has emerged as a prominent acceleration method aimed at reducing inference steps or model sizes \cite{li2024snapfusion, meng2023distillation, salimans2022progressive}. However, distillation requires additional training, which significantly restricts its applicability, as training can be both costly and challenging. And distillation is inherently model-dependent, so it must be performed each time a new model is released, further increasing the cost.

There are also training-free methods aimed at acceleration, such as \cite{ma2024deepcache}, which specifically target denoising models with a U-Net architecture \cite{ronneberger2015u}. \cite{castillo2023adaptive} first introduced the concept of adaptive guidance, revealing that classifier-free guidance in the later denoising steps is less critical and suggesting the use of similarity as a threshold. However, this approach has several shortcomings, including a necessity of threshold selection and a lack of comprehensive quantitative results. Additionally, they do not demonstrate whether their findings can be generalized across most generative diffusion models.

Inspired by \cite{castillo2023adaptive}, we aim to conduct a deeper exploration of adaptive guidance. We first demonstrate that \textbf{the adaptive guidance strategy proposed in \cite{castillo2023adaptive} fails to generalize across most models}. Next, we provide a new view that explains why classifier-free guidance is less critical in the later denoising steps. This analysis posits that, during these later steps, the signal-to-noise ratio (\textit{SNR}) is high, resulting in similar denoising outcomes across different paths.

Building on this analysis, we propose a new adaptive guidance strategy, \textit{SNR AG}, along with a simplified alternative, \textit{Step AG}. Empirical results show that limiting classifier-free guidance to the first 30\% to 50\% of denoising steps achieves similar generation performance (as shown in Figure \ref{fig:teaser}) to applying classifier-free guidance throughout the entire denoising process. We further demonstrate that this finding also applies to video generation models.

Our contributions are listed as follows:

\begin{itemize}
    \item We reveal the inapplicability of current adaptive guidance strategy and present a new view of understanding why adaptive guidance works.
    \item We propose \textit{Step AG} as a universally applicable adaptive guidance strategy. We empirically show that our proposed strategy saves 20\% to 30\% inference time with almost no cost on generation performance.
    \item We show that \textit{Step AG} provides consistent improvement across different settings such as inference steps, and various models including even video generation models, highlighting the superiority of our method. We also empirically show that our method consistently works well with the same setting, highlighting its generalization ability.
\end{itemize}

\section{Related Works}
\subsection{Diffusion Models}
\cite{NEURIPS2020_DDPM} first introduced DDPM, which serves as the foundation for subsequent diffusion models. In diffusion models, there are generally two types of conditioning algorithms: classifier guidance \cite{dhariwal2021diffusion} and classifier-free guidance \cite{ho2022classifier}. \cite{Rombach_2022_CVPR} proposes conducting denoising in latent space, a technique that has proven highly successful.

For noise scheduling in diffusion models, \cite{nichol2021improved} introduced cosine noise scheduling to enhance DDPM. \cite{song2021DDIM} proposed DDIM, which allows for significantly fewer sampling steps in diffusion models. Numerous studies aim to interpret diffusion models from various perspectives. Stochastic Differential Equations (SDE) and Ordinary Differential Equations (ODE) are essential frameworks for understanding diffusion models \cite{song2021scorebased, karras2022elucidating, kingma2024understanding}. Further research has introduced high-order solvers to enable a more precise denoising process \cite{lu2022dpm, lu2022dpm++}. Flow matching provides another compelling approach for interpreting the diffusion process \cite{albergo2022building, liu2022flow, lipman2022flow, esser2024scaling}.

Many studies also focus on practical applications of diffusion models, including \cite{podell2023sdxl, esser2024scaling, chen2024pixartsigma, ho2022imagen, peebles2023scalable} for image generation, \cite{yang2024cogvideox, wang2023modelscope, NEURIPS2022_VideoDiff, luo2023video} for video generation, and \cite{popov2021grad, kong2020diffwave} for audio generation. Diffusion models have also been successfully applied to other tasks, such as image editing \cite{brooks2023instructpix2pix}.

 \subsection{Acceleration for Diffusion Models}
Numerous efforts have been made to accelerate inference in diffusion models, with one of the most successful approaches being DDIM \cite{song2021DDIM}, along with the noise scheduling methods mentioned earlier. Additionally, distillation techniques, such as those in \cite{li2024snapfusion, meng2023distillation, salimans2022progressive}, focus on further reducing inference steps or model size. However, as noted, distillation-based methods require separate training for each specific model, which can be complex and resource-intensive.

\cite{ma2024deepcache} proposes a training-free method that caches intermediate outputs of U-Net \cite{ronneberger2015u}, a widely used denoising architecture. However, many contemporary models employ DiT \cite{peebles2023scalable} instead of U-Net, making this caching approach impractical. \cite{NEURIPS2024_f0b1515b} proposes a DiT-based caching strategy, but this approach cannot be applied to U-Net based models, revealing a severe problem of these caching-based methods that they rely on specific model architecture, lacking enough generalization ability. Some studies also apply Neural Architecture Search (NAS) \cite{liu2018darts} to optimize the denoising path \cite{li2023autodiffusion, castillo2023adaptive}, yet identifying an appropriate search objective and implementing NAS effectively can be challenging and still expensive.

Adaptive guidance, which applies classifier-free guidance to only certain steps of the denoising process, was first introduced by \cite{castillo2023adaptive}. However, as we have observed, their strategy does not generalize to most diffusion models. Similar ideas have been presented by \cite{kynkaanniemi2024applying, sadat2023cads}, but these methods focus on enhancing generation quality rather than exploring their potential for acceleration. \cite{zhang2024cross} is a similar work focusing on cross-attention instead of directly assessing CFG steps, which is beyond our discussion. Also, these approaches are all complex and require intricate hyperparameter tuning, which limits their practical applicability. 

\section{Adaptive Guidance}
\subsection{Preliminaries}
Different diffusion models formulate their forward and reverse processes in varying ways; however, a diffusion process can generally be modeled as follows: Given an original sample $x_0$ and a timestep $t$ (which may be continuous or discrete), the noisy sample $x_t$ is drawn from $p(x_t | x_0)$. Typically, $p(x_t | x_0) = \mathcal{N}(\alpha(t) x_0, \sigma^{2}(t) I)$, where $\alpha(t)$ and $\sigma(t)$ are functions defined by specific noise schedules and $I$ is an identity matrix .

This forward and reverse process can take various forms depending on the choice of noise scheduling (and training) strategies \cite{karras2022elucidating, kingma2024understanding, lu2022dpm, lipman2022flow}. Despite these differences, the model’s prediction at timestep $t$, parameterized by $\theta$, can be represented as a score $\epsilon_{\theta}(x_t, t)$.

For conditioned image generation, the input includes an additional condition $c$ that the generated image should align with. \cite{ho2022classifier} propose classifier-free guidance (CFG) to address this requirement. Specifically, classifier-free guidance trains a model capable of predicting both an unconditional score $\epsilon(x_t, t)$ and a conditional score $\epsilon(x_t, t, c)$. The final output score is obtained as :
\begin{equation}
    \tilde{\epsilon}_{\theta}(x_t, t, c) = \epsilon_{\theta}(x_t, t) + w[\epsilon_{\theta}(x_t, t, c) - \epsilon_{\theta}(x_t, t)]
\end{equation}
$w$ is a hyper-parameter, namely ``guidance scale''. 

\begin{figure*}[t]  
    \centering  
    \includegraphics[width=16cm]{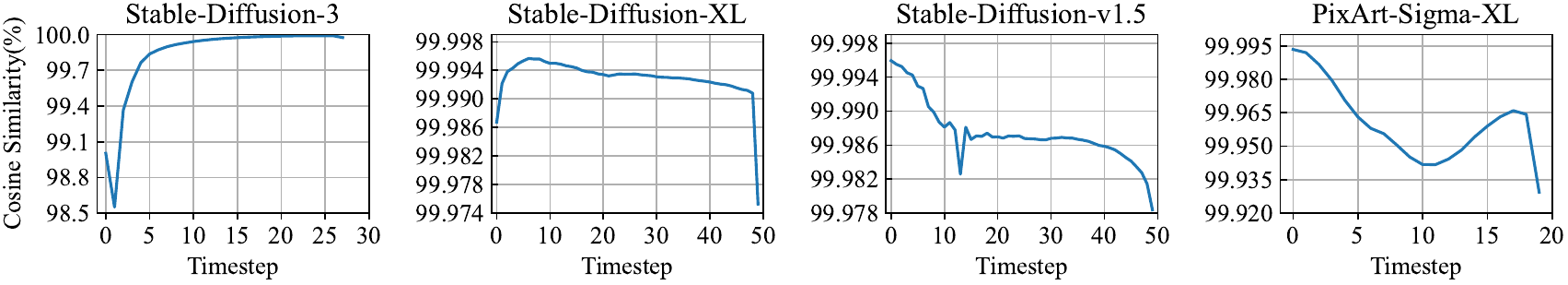}
    \caption{Cosine Similarity $\gamma_t$ of different models. We calculate cosine similarity under their default inference step settings.}
    \label{fig:cossim}
\end{figure*}  

\subsection{Existing Adaptive Guidance Strategy}
Classifier-free guidance has demonstrated exceptional performance across various generation tasks. However, it has a significant drawback: it requires two forwards in a single denoising step. Given that multiple denoising steps are needed during inference, this results in substantial computational costs. Therefore, a natural question arises: can some $\tilde{\epsilon}_{\theta}(x_t, t, c)$ be replaced by $\epsilon_{\theta}(x_t, t, c)$ or $\epsilon_{\theta}(x_t, t)$?

Following \cite{castillo2023adaptive}, this question can be formulated as follows: Given a set $\mathcal{F}_{t} = \{\tilde{\epsilon}_{\theta}(x_t, t, c), \epsilon_{\theta}(x_t, t, c), \epsilon_{\theta}(x_t, t)\}$ representing the possible choices at a particular timestep $t$, all possible denoising paths over a total of $T$ steps can then be represented as a set $S = \mathcal{F}_{T} \times \mathcal{F}_{T-1} \times ... \times \mathcal{F}_{1}$, where $\times$ represents Cartesian product. The objective of adaptive guidance (AG) is to identify a strategy $\zeta \in S$ that achieves competitive generation performance at a reduced computational cost. For example, the classic classifier-free guidance strategy can be represented as:

\begin{equation}
    \zeta_{CFG} = \{\tilde{\epsilon}_{\theta}(x_T, T, c), ..., \tilde{\epsilon}_{\theta}(x_1, 1, c)\}
\end{equation}

To find a good strategy $\zeta \in S$, \citep{castillo2023adaptive} applies neural architecture search (NAS) \cite{liu2018darts} and observes that CFG is more critical in the initial steps of denoising. Let 
\begin{equation}
     \gamma_t := \dfrac{|\epsilon_{\theta}(x_t, t, c)\cdot \epsilon_{\theta}(x_t, t)|}{||\epsilon_{\theta}(x_t, t, c)|| \cdot ||\epsilon_{\theta}(x_t, t)||}
\end{equation}

which represents the cosine similarity between conditional and unconditional score at timestep $t$,
they also find that $\gamma_t$ monotonically increases to one during the denoising process. Based on this observation, they propose a similarity-based adaptive guidance strategy, which we name \textit{Similarity AG}:

\begin{gather}
    \begin{aligned}
        \zeta_{SimAG} = \{\tilde{\epsilon}_{\theta}(x_T, T, c),...,\tilde{\epsilon}_{\theta}(x_{t_{0}}, t_{0}, c), \\
        \epsilon_{\theta}(x_{t_{0}-1}, t_{0}-1, c),...,\epsilon_{\theta}(x_{1}, 1, c)\}
    \end{aligned}\\
    \begin{aligned}
        t_{0} &= \max t, \\ s.t. \gamma_t :&= \dfrac{|\epsilon_{\theta}(x_t, t, c)\cdot \epsilon_{\theta}(x_t, t)|}{||\epsilon_{\theta}(x_t, t, c)|| \cdot ||\epsilon_{\theta}(x_t, t)||} > \gamma
\end{aligned}
\end{gather}

$\gamma$ is a pre-selected threshold.



\paragraph{Empirical Analysis}
Although \cite{castillo2023adaptive} empirically observed that $\gamma_t$ monotonically approaches one on their trained models, it remains uncertain whether this observation can be generalized to all types of diffusion models. We investigate several publicly available models: Stable-Diffusion-3 \cite{esser2024scaling}, Stable-Diffusion-XL \cite{podell2023sdxl}, Stable-Diffusion-v1.5 \cite{Rombach_2022_CVPR}, and PixArt-$\Sigma$-XL \cite{chen2024pixartsigma}. We compute the similarity $\gamma_t$ on $20$ samples (20 prompts and 1 sample per prompt) and average them to produce Figure \ref{fig:cossim}.

The results are surprising, as these models exhibit relatively high cosine similarity almost throughout, which highlights the challenge of selecting an appropriate similarity threshold. Furthermore, for many models, the similarity is not monotonically increasing at all! Thus, \textbf{using similarity as a threshold is impractical, as it would often lead to discontinuing classifier-free guidance within the very first one or two steps}, which is clearly unsuitable. 

To dive deeper, the similarity $\gamma_t$ depends on many factors, including the total denoising steps $T$, the initial noise sampled $x_T$, the model parameters $\theta$ and the noise scheduler. There is no fundamental guarantee that $\gamma_t$ would always monotonically increase. Therefore, using similarity as a threshold actually lacks solid basis, since the assumption that $\gamma_t$ is monotonically increasing cannot be validated. This observation also suggests that similarity is not the intrinsic reason why adaptive guidance is useful, since according to our observation the similarity $\gamma_t$ is always quite large. We will present more empirical analysis on \textit{Similarity AG} in Section \ref{sec:abla}.

Therefore, it is natural to ask whether there exists an adaptive guidance strategy that delivers stable performance and can generalize across all models. To address this question, we aim to provide a straightforward yet insightful analysis of why adaptive guidance is possible. Based on this analysis, we propose a simple adaptive guidance strategy that can be applied to all types of classifier-free guidance text-to-vision diffusion models.

\subsection{Our Proposed Strategy}
\label{Sec:me_the}
Still taking the most generalized formulation of the forward process $p(x_t|x_0) = \mathcal{N}(\alpha(t)x_0, \sigma^{2}(t)I)$, with a simple re-parameterization trick, we have 

\begin{equation}
    x_t = \alpha(t)x_0 + \sigma(t) \epsilon, \epsilon \sim \mathcal{N}(0, I)
\end{equation}

Let $\lambda_t = \dfrac{\alpha(t)}{\sigma(t)}$, where $\lambda_t$ is known as the \textit{Signal-To-Noise Ratio (SNR)}, indicating the noise level in the data. When $\lambda_t$ is large, the data is less noisy, as the $x_0$ part is larger. Since the diffusion process removes noise from the data, less noisy data has fewer possible denoising directions, resulting in similar denoising outcomes, indicating the potential feasibility of stop using classifier-free guidance in the later denoising steps.

This analysis highlights the significance of $\lambda_t$. Therefore, we propose using $\lambda_t$ directly as a threshold, which leads us to introduce \textit{SNR AG}.

\paragraph{SNR AG}
A strategy following \textit{SNR AG} can always be represented as:
\begin{gather}
    \begin{aligned}
        \zeta_{SNRAG} = \{\tilde{\epsilon}_{\theta}(x_T, T, c),...,\tilde{\epsilon}_{\theta}(x_{t_{0}}, t_{0}, c), \\
        \epsilon_{\theta}(x_{t_{0}-1}, t_{0} - 1, *), ..., \epsilon_{\theta}(x_{1}, 1, *)\}, 
    \end{aligned}\\
    t_0 = \max t, s.t. \lambda_t := \dfrac{\alpha(t)}{\sigma(t)} > \lambda
    \label{eq:t}
\end{gather}
where $\lambda$ is a pre-selected threshold, * represents either conditional score ($\epsilon_{\theta}(x_t, t, c)$) or unconditional score ($\epsilon_{\theta}(x_t, t)$) can be used in the later steps.

We would like to note that, though \textit{SNR AG} bears similarity with \textit{Similarity AG}, the insight is completely distinct. \textbf{\textit{SNR AG} argues that the reason why adaptive guidance works lies in the formulation of denoising process instead of certain model prediction behaviour.} Therefore, \textit{SNR AG} is naturally independent of model training or prediction behaviour, ensuring its generalization on all diffusion models, while \textit{Similarity AG} is not applicable on most models. 

\begin{figure}
    \centering
    \includegraphics[width=0.9\linewidth]{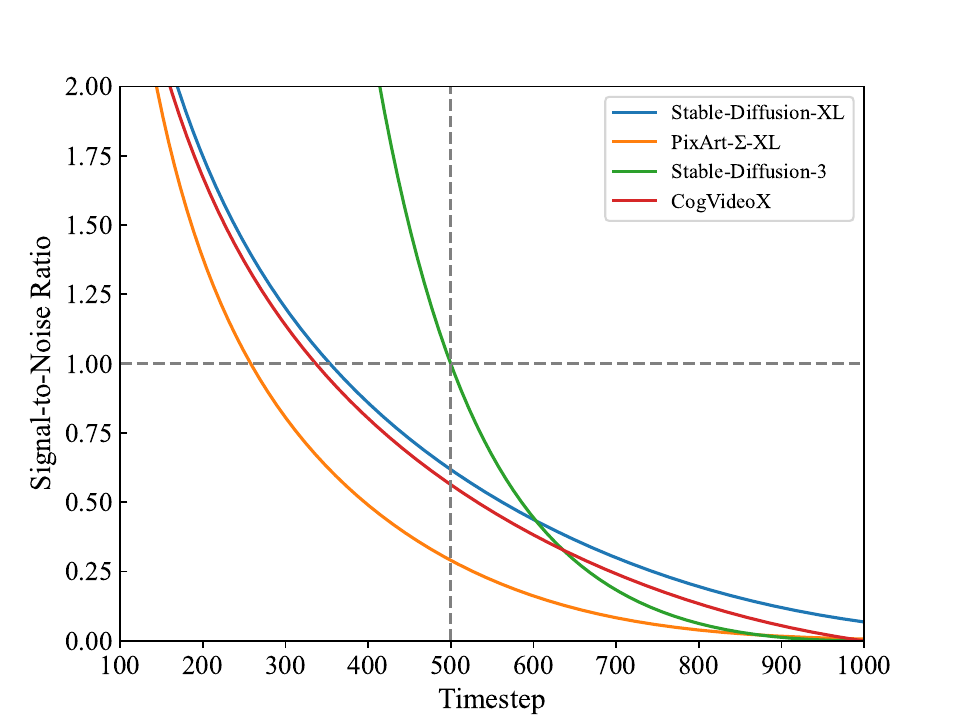}
    \caption{\textit{SNR} of different diffusion models. This is calculated under training timestep setting. Inference bears a similar trend.}
    \label{fig:snr}
\end{figure}

Further, since the reverse process in diffusion models involves removing noise from the data, \textbf{$\lambda_t$ should generally increase as $t$ decreases, making our formulation fundamentally established}. To provide a more intuitive understanding, since $\alpha(t)$ and $\sigma(t)$ only depend on certain noise scheduling, we investigate the trend of $\lambda_t$ using several open-source diffusion models (Stable-Diffusion-XL \cite{podell2023sdxl}, PixArt-$\Sigma$-XL \cite{chen2024pixartsigma}, Stable-Diffusion-3 \cite{esser2024scaling}, CogVideoX \cite{yang2024cogvideox}) to provide an empirical observation. As shown in Figure \ref{fig:snr}, the experiment results clearly demonstrate that $\lambda_t$ is increasing as $t$ decreases. 

However, \textit{SNR AG} bears a natural drawback that, it requires certain knowledge of the noise scheduler. Also, selecting $\lambda$ does not provide an intuitive understanding of how much efficiency increase the method would provide (both of which are also shortcomings of \textit{Similarity AG}). Therefore, we would like to provide a more simplified version of \textit{SNR AG} that is more user-friendly and provides guarantee of efficiency increase. 

We start by noting that once a noise scheduler and total denoising steps $T$ is fixed, the SNR $\lambda_t$ at each timestep is fixed. Therefore, for any threshold $\lambda$, there exists a corresponding threshold $t$, which is $\max t, s.t. \lambda_t>\lambda$, as indicated in Equation \ref{eq:t}. Therefore, we can directly select this $t$ as a threshold to avoid calculating $\lambda$ explicitly while maintaining the same behaviour as \textit{SNR AG}. 

Further, note that under different total denoising steps $T$, the threshold $t$ corresponding to a certain $\lambda$ is naturally different, but the SNR trend under different $T$ remains the same (which is a basis of current noise schedulers), therefore we can formulate $t = (1-p)T$, where $p \in [0, 1]$ is a hyper-parameter. This formulation means that classifier-free guidance is only applied to the first $p$(ratio) steps. Since the SNR trend remains similar under different $T$, $p$ remains similar for different $T$ under the same threshold $\lambda$, eliminating the need for specific threshold selection. This leads us to: 

\paragraph{Step AG}
A strategy following \textit{Step AG} can always be represented as:

\begin{gather}
\begin{aligned}
    \zeta_{StepAG} = \{\tilde{\epsilon}_{\theta}(x_T, T, c),...,\tilde{\epsilon}_{\theta}(x_{t_{0}}, t_{0}, c), \\
    \epsilon_{\theta}(x_{t_{0}-1}, t_{0} - 1, *), ..., \epsilon_{\theta}(x_{1}, 1, *)\},
\end{aligned}\\
    t_0 = (1-p)T
\end{gather}
where $p$ is a pre-selected threshold, namely \textbf{guidance ratio}, * bears the same meaning as before. We will further discuss which is a better option in Section \ref{sec:image_exp}.

Based on the analysis above, \textit{Step AG} has identical behaviour with \textit{SNR AG} (with corresponding thresholds selected). However, \textit{Step AG} is significantly simpler to implement and requires minimal prior knowledge of the specific noise schedulers. It also provides a predictable time-saving guarantee. Theoretically, the ratio of inference cost saved is $\dfrac{1 - p}{2}$, as only one forward pass is required in the later $t_0 = (1-p)T$ inference steps instead of two.  

To take a step further, if $p$ is needed to be selected for each model specifically, the practical application of this method would still be degraded. However, as can be observed in Figure \ref{fig:snr}, all models bear a similar \textit{SNR} trend, \textbf{indicating a potential feasibility of using the same $p$ across different models}! We present detailed experiments to verify this feasibility in Section \ref{sec:exp}.

We have to note that, using the same $p$ across different models actually results in different threshold $\lambda$ corresponding to each model. However, the aim of this design is to best simplify implementation and facilitate application of our method. \textbf{We would like to show that our method works well as long as the selected threshold is reasonable.}
 
\begin{table*}[t]
    \centering
    \setlength{\tabcolsep}{4mm}
    \begin{tabular}{c|c|c|ccc}
    \toprule[1.5pt]
        Model Name & \makecell[c]{Guidance \\Ratio($p$)} & Score Type & FID($\downarrow$) & CLIP Score($\uparrow$) & SPI(s)($\downarrow$)\\
    \toprule[1.5pt]
        \multirow{5}{*}{Stable-Diffusion-3} & 1.0 & - & 58.52 & \textbf{26.13} & 5.25\\
        \cline{2-6}
        ~ &  \multirow{2}{*}{0.5}  & Unconditional & 58.24 \textcolor{blue}{$_{\downarrow 0.4\%}$} & 25.87 \textcolor{SeaGreen}{$_{\downarrow 1.0\%}$} & 4.08 \textcolor{blue}{$_{\downarrow 22.3\%}$} \\
        ~ & ~  & Conditional & 58.53 \textcolor{SeaGreen}{$_{\uparrow 0.0\%}$} & 25.98 \textcolor{SeaGreen}{$_{\downarrow 0.6\%}$} & 4.08 \textcolor{blue}{$_{\downarrow 22.3\%}$} \\
        \cline{2-6}
        ~ &  \multirow{2}{*}{0.3}  & Unconditional & \textbf{58.07}\textcolor{blue}{$_{\downarrow 0.7\%}$} & 24.95 \textcolor{SeaGreen}{$_{\downarrow 4.5\%}$} & \textbf{3.59} \textcolor{blue}{$_{\downarrow 31.6\%}$}\\
        ~ & ~ & Conditional & 58.33\textcolor{blue}{$_{\downarrow 0.3\%}$} & 25.57 \textcolor{SeaGreen}{$_{\downarrow 2.1\%}$} & \textbf{3.59} \textcolor{blue}{$_{\downarrow 31.6\%}$} \\
        \midrule[1.0pt]
        \multirow{5}{*}{Stable-Diffusion-XL} & 1.0 & - & \textbf{53.83} & \textbf{26.02} & 13.44 \\
        \cline{2-6}
        ~ &  \multirow{2}{*}{0.5} &  Unconditional & 53.96 \textcolor{SeaGreen}{$_{\uparrow 0.2\%}$} & 25.62 \textcolor{SeaGreen}{$_{\downarrow 1.5\%}$} & 10.46 \textcolor{blue}{$_{\downarrow 22.2\%}$} \\
        ~ &~ & Conditional & 53.97 \textcolor{SeaGreen}{$_{\uparrow 0.2\%}$} & 25.71 \textcolor{SeaGreen}{$_{\downarrow 1.2\%}$} & 10.47 \textcolor{blue}{$_{\downarrow 22.1\%}$} \\
        \cline{2-6}
        ~ &  \multirow{2}{*}{0.3} &  Unconditional & 54.28 \textcolor{SeaGreen}{$_{\uparrow 0.8\%}$} & 24.82 \textcolor{SeaGreen}{$_{\downarrow 4.6\%}$} & \textbf{9.28} \textcolor{blue}{$_{\downarrow 31.0\%}$} \\
        ~ & ~ & Conditional & 54.15 \textcolor{SeaGreen}{$_{\uparrow 0.6\%}$}  & 25.24 \textcolor{SeaGreen}{$_{\downarrow 3.0\%}$} & 9.31 \textcolor{blue}{$_{\downarrow 30.7\%}$} \\
        \midrule[1.0pt]
        \multirow{5}{*}{Stable-Diffusion-1.5} & 1.0 &  - & 52.97 & \textbf{25.52} & 2.39\\
        \cline{2-6}
        ~ &  \multirow{2}{*}{0.5} &  Unconditional & 53.06 \textcolor{SeaGreen}{$_{\uparrow 0.2\%}$} & 24.88 \textcolor{SeaGreen}{$_{\downarrow 2.5\%}$} & 1.94 \textcolor{blue}{$_{\downarrow 18.8\%}$}\\
        ~ &  ~ & Conditional & 53.06 \textcolor{SeaGreen}{$_{\uparrow 0.2\%}$} & 25.05 \textcolor{SeaGreen}{$_{\downarrow 1.8\%}$} & 1.95 \textcolor{blue}{$_{\downarrow 18.4\%}$}\\
        \cline{2-6}
        ~ &  \multirow{2}{*}{0.3}  & Unconditional & 53.20 \textcolor{SeaGreen}{$_{\uparrow 0.4\%}$} & 23.41 \textcolor{SeaGreen}{$_{\downarrow 8.3\%}$} & 1.79 \textcolor{blue}{$_{\downarrow 25.1\%}$}\\
        ~ &  ~ & Conditional & \textbf{52.61} \textcolor{blue}{$_{\downarrow 0.7\%}$} & 24.32 \textcolor{SeaGreen}{$_{\downarrow 4.7\%}$} & \textbf{1.78} \textcolor{blue}{$_{\downarrow 25.5\%}$} \\
        \midrule[1.0pt]
        \multirow{5}{*}{PixArt-$\Sigma$-XL} &  1.0 & - & 63.90 & \textbf{25.80} & 5.96\\
        \cline{2-6}
        ~ &  \multirow{2}{*}{0.5}  & Unconditional & 63.04 \textcolor{blue}{$_{\downarrow 1.3\%}$} & 25.07 \textcolor{SeaGreen}{$_{\downarrow 2.8\%}$} & 4.63 \textcolor{blue}{$_{\downarrow 22.3\%}$}\\
        ~ &  ~ & Conditional & \textbf{62.60} \textcolor{blue}{$_{\downarrow 2.0\%}$} & 25.47 \textcolor{SeaGreen}{$_{\downarrow 1.3\%}$} & 4.64 \textcolor{blue}{$_{\downarrow 22.1\%}$}\\
        \cline{2-6}
        ~ & \multirow{2}{*}{0.3} &  Unconditional & 85.10 \textcolor{SeaGreen}{$_{\uparrow 33.2\%}$} & 19.81 \textcolor{SeaGreen}{$_{\downarrow 23.2\%}$} & 4.11 \textcolor{blue}{$_{\downarrow 31.0\%}$} \\
        ~ &  ~ & Conditional & 64.02 \textcolor{SeaGreen}{$_{\uparrow 0.1\%}$} & 24.43 \textcolor{SeaGreen}{$_{\downarrow 5.3\%}$} & \textbf{4.10} \textcolor{blue}{$_{\downarrow 31.2\%}$}  \\
        \bottomrule[1.5pt]
    \end{tabular}
    
    \caption{Evaluation results of \textit{Step AG} under different models and settings. Note that there is no difference between using conditional and unconditional scores under $p=1.0$. We label improvements against $p=1.0$ in \textcolor{blue}{blue} and declines against $p=1.0$ in \textcolor{SeaGreen}{SeaGreen}.}
    \label{tab:main_exp}
\end{table*}

\section{Experiment Setup}
\label{sec:exp_setup}

\subsection{Models, Datasets and Metrics}
We use only open-source models for our experiments, as we lack access to the inference procedures of closed-source models. For image generation models, we primarily examine four models: Stable-Diffusion-3 \cite{esser2024scaling}, Stable-Diffusion-XL \cite{podell2023sdxl}, Stable-Diffusion-1.5 \cite{Rombach_2022_CVPR}, and PixArt-$\Sigma$ \cite{chen2024pixartsigma}. These models vary in size, architecture, and noise scheduling strategies, and collectively represent typical diffusion models for image generation. Following \cite{esser2024scaling}, we use the MSCOCO 2014 validation set \cite{lin2014microsoft} for evaluation. We also conduct an extra evaluation on ImageNet dataset \cite{deng2009imagenet}, whose details can be found in Appendix B. Each metric is calculated using 4000 samples (1000 prompts and 4 samples per prompt). We select FID \cite{heusel2017gans} \footnote{FID is calculated with pytorch\_fid \citep{Seitzer2020FID}.} and CLIP Score \cite{hessel2021clipscore, radford2021learning} as our evaluation metrics following previous practice.

For video generation models, we discuss two models: CogVideoX \cite{yang2024cogvideox} and ModelScope \cite{wang2023modelscope, luo2023video}. We choose VBench \cite{huang2023vbench} for evaluation. Specifically, we select eight metrics from VBench following \cite{yang2024cogvideox} (motion smoothness, human action, subject consistency, appearance style, multiple objects, temporal flickering, scene and dynamic degree). 

For acceleration, we use seconds per image (SPI) or seconds per video (SPV) as the metric. It is calculated with an inference batch size of one.

\subsection{Implementation Details}
In our experiments, we primarily focus on \textit{Step AG}, as it is equivalent to \textit{SNR AG} and considerably easier to implement. 

Referring to Figure \ref{fig:snr}, we observe that all models exhibit a relatively low \textit{SNR} at $p=0.5$. For more aggressive acceleration, we find that most models reach $\lambda_t = 1$ around $p \approx 0.3$, so we also evaluate $p=0.3$ in the following experiments. 

The guidance scale $w$ is another important hyper-parameter we aim to investigate. We set $w = 7$ for image generation experiments and $w = 6$ for video generation experiments generally \textbf{following default settings}. Additionally, we examine $w = 15$ to explore the impact of different values of $w$, as detailed in Section \ref{sec:guidance_scale}.

The number of inference steps $T$ is also a crucial factor influencing model behavior, which we will discuss in detail in Section \ref{sec:abla}. For other experiments, we simply \textbf{use the default value of $T$}.

Lastly, as we have demonstrated, in later few steps where classifier-free guidance is not applied, either the conditional score $\epsilon_{\theta}(x_t, t, c)$ or the unconditional score $\epsilon_{\theta}(x_t, t)$ can be used. We also aim to explore the effect of using different scores in this phase, a setting we refer to as Score Type.

The diffusion prior (i.e., the initial noise sampled) remains consistent across different $p$, $w$, and Score Types for the same model to ensure a fair comparison among various settings. 

More details about our experiment setup can be found in Appendix A. 
\section{Experiment Results and Analysis}
\label{sec:exp}

\subsection{Image Generation Results}
\label{sec:image_exp}

First of all, we would like to show that \textit{Step AG} is universally applicable in text-to-image generation. The results are shown in Table \ref{tab:main_exp}.

As shown in the results, FID and CLIP Score metrics remain competitive for both $p=0.5$ and $p=0.3$ across all models. These findings demonstrate the effectiveness of \textit{Step AG} for all text-to-image diffusion models.

The inference time saved generally aligns with our previous analysis, where \textit{Step AG} theoretically saves $\dfrac{1 - p}{2}$ of the inference time. Setting $p = 0.5$ results in an average time saving of 20\%, while setting $p = 0.3$ achieves an average saving of 30\%. The discrepancy between the empirical results and the theoretical upper bound ($\dfrac{1 - p}{2}$) is due to constant costs, such as text encoding and image latent decoding. This improvement is consistent across all tested models, underscoring the universal applicability of \textit{Step AG}.

Regarding image quality and image-text alignment, FID sometimes decreases when setting $p<1.0$, indicating \textit{Step AG} sometimes partially improves image quality. This aligns with observations from \cite{kynkaanniemi2024applying, sadat2023cads}. Our finding is even more encouraging, as \textit{Step AG} offers this benefit with significantly less complexity. However, we notice that CLIP Score also shows a slight decrease as the Guidance Ratio $p$ decreases, indicating that this benefit comes with the cost of harming image-text alignment, which is not discussed by previous works. Also, under some cases, image quality decreases when \textit{Step AG} is applied, so there is no guarantee of quality improvement using \textit{Step AG}.  

\paragraph{Influence of Different Score Types}
As discussed earlier, both the conditional score $\epsilon_{\theta}(x_t, t, c)$ and the unconditional score $\epsilon_{\theta}(x_t, t)$ can be used in the later denoising steps. An interesting question is whether there is a significant difference in outcomes between using these two scores. Our results show that, in some cases, using the unconditional score yields even better FID compared to using the conditional score. However, this improvement often comes at the cost of further reduced image-text alignment. Generally using either conditional or unconditional score is acceptable. 

We also would like to note that, while the decrease in CLIP Score is generally modest, certain models, such as Stable-Diffusion-1.5 and PixArt-$\Sigma$-XL, exhibit a marked drop in CLIP Score when using the unconditional score under $p=0.3$. Therefore, \textbf{we recommend using the conditional score in \textit{Step AG} to prevent unexpected, severe performance declines}.

\subsection{Further Analysis}
\label{sec:abla}

\paragraph{Analysis on Inference Steps $T$}
We first investigate whether \textit{Step AG} performs effectively with different selections for inference steps $T$. As indicated previously, using the unconditional score $\epsilon_{\theta}(x_t, t)$ in the later steps can degrade image-text alignment, so we focus exclusively on using the conditional score in the later denoising steps here. We use Stable-Diffusion-3 as the base model and select $T = 28$, $21$, and $19$ for evaluation. The results are shown in Table \ref{tab:ablation}. 

\begin{table}[htbp]
    \centering
    \begin{tabular}{c|ccc}
        \toprule
        $T$ & Guidance Ratio($p$) & FID ($\downarrow$) & CLIP Score ($\uparrow$) \\
        \midrule
        \multirow{3}{*}{28} & 1.0 & 58.52 & \textbf{26.13} \\
        ~ & 0.5 & 58.53 & 25.98 \\
        ~ & 0.3 & \textbf{58.33} & 25.57  \\
        \hline   
        \multirow{3}{*}{21} & 1.0 & 58.73 & \textbf{26.13} \\
        ~ & 0.5 & 58.97 & 25.95 \\
        ~ & 0.3 & \textbf{58.59} & 25.48  \\
        \hline   
        \multirow{3}{*}{19} &  1.0 & 58.57 & \textbf{26.10} \\
        ~ & 0.5 & \textbf{58.55} & 25.89 \\
        ~ & 0.3 & 58.67 & 25.33 \\
        \bottomrule   
    \end{tabular}
    \caption{Performance of \textit{Step AG} under different inference steps $T$ settings in Stable-Diffusion-3.}
    \label{tab:ablation}
\end{table}

As shown in the results, \textit{Step AG} achieves competitive performance across all $T$ settings. Our observation from Section \ref{sec:image_exp} holds: reducing the number of classifier-free guidance steps compromises image-text alignment (CLIP Score). Another interesting observation is that fewer total inference steps sometimes do not result in a significant performance drop. For instance, using $T=21$ yields a similar CLIP Score and FID compared to $T=28$.

A natural question arises: why using \textit{Step AG} instead of simply reducing the total number of inference steps? This result demonstrates that both approaches can, in fact, be combined to further accelerate inference. As long as the total number of inference steps remains within a reasonable range, \textit{Step AG} can be applied as a cost-free acceleration method. Additionally, it is widely recognized that excessively reducing inference steps may significantly degrade model performance, whereas, as we have shown, \textit{Step AG} generally preserves model performance. A qualitative example will be presented in Section \ref{sec:case}. 

\paragraph{Comparison with \textit{Similarity AG}}
To verify our analysis about \textit{Similarity AG}, we replicate it on Stable-Diffusion-1.5 with different thresholds $\gamma$. Other settings remain the same as previous experiments. The results are shown in Table \ref{tab:sim-ag-sd15}.

\begin{table}[ht]
    \centering
    \setlength{\tabcolsep}{1.5mm}
    \begin{tabular}{c|ccc}
        \toprule
        \makecell[c]{Similarity \\ Threshold $\gamma$} & FID ($\downarrow$) & CLIP Score ($\uparrow$) & SPI(s) ($\downarrow$) \\
        \midrule
        1.0 & \textbf{52.97} & \textbf{25.52} & 2.39 \\
        Ours, $p=0.5$ & 53.06 & 25.05 & 1.95 \\
        0.9 & 74.00 & 21.11 & \textbf{1.59} \\
        0.95 & 74.00 & 21.11 & \textbf{1.59} \\
        0.99 & 74.00 & 21.11 & \textbf{1.59} \\
        0.999 & 74.00 & 21.11 & \textbf{1.59} \\
        0.9999 & 73.87 & 21.12 & \textbf{1.59} \\
        \bottomrule
    \end{tabular}
    \caption{Performance of \textit{Similarity AG} on Stable-Diffusion-1.5 under different $\gamma$ selections. $\gamma = 1.0$ refers to the baseline where classifier-free guidance is applied in the whole denoising process.}
    \label{tab:sim-ag-sd15}
\end{table}

As can be seen, the performance of \textit{Similarity AG} remain terrible as similarity threshold $\gamma$ changes, with FID increasing almost 40\% and CLIP Score decreasing 17\%. This observation matches our analysis that \textit{Similarity AG} does not perform well on Stable-Diffusion-1.5 whose corresponding similarity is not increasing during denoising process, and mostly just stops classifier-free guidance in the very first step. This result clearly demonstrates the inapplicability of \textit{Similarity AG} and highlights the importance of \textit{Step AG}. In fact, compared with \textit{Similarity AG}, our method achieves a notable speedup with little performance drop, while \textit{Similarity AG} behaves almost same under different threshold selections, with a slightly better speedup but much worse generation performance. 

\paragraph{Influence of Different Guidance Scales}
\label{sec:guidance_scale}
We would like to explore the influence of different guidance scales, so we select $w=7$ and $w=15$ in the following experiments. The results are shown in Table \ref{tab:abla_w}. 

\begin{table}[htbp]
    \centering
    \setlength{\tabcolsep}{2.5mm}
    \begin{tabular}{c|c|c|cc}
    \toprule[1.5pt]
        Model Name & $p$ & $w$  & FID($\downarrow$) & CLIP Score($\uparrow$) \\
    \toprule[1.5pt]
        \multirow{6}{*}{\makecell[c]{Stable-\\Diffusion-\\3}} &   \multirow{2}{*}{1.0} & 7  & 58.52 & \textbf{26.13}\\
       ~ & ~ & 15 & 65.19 & 25.64 \\
        \cline{2-5}
        ~ &  \multirow{2}{*}{0.5} & 7  & 58.53 & 25.98\\
        ~ & ~ & 15  & 70.96 & 25.52\\
        \cline{2-5}
        ~ & \multirow{2}{*}{0.3} & 7  & \textbf{58.33} & 25.57 \\
        ~ & ~ & 15  & 72.35 & 25.05  \\
        \midrule[1pt]
        \multirow{6}{*}{\makecell[c]{Stable-\\Diffusion-\\XL}} &  \multirow{2}{*}{1.0} & 7 & \textbf{53.83} & 26.02  \\
        ~ & ~ & 15 & 55.48 & \textbf{26.54} \\
        \cline{2-5}
        ~ &  \multirow{2}{*}{0.5} & 7 & 53.97 & 25.71  \\
        ~ & ~ & 15 & 55.73 & 26.35  \\
        \cline{2-5}
        ~ &  \multirow{2}{*}{0.3} & 7 & 54.15 & 25.24\\
        ~ & ~ & 15 & 55.52 & 26.02  \\
        \midrule[1pt]
        \multirow{6}{*}{\makecell[c]{Stable-\\Diffusion-\\1.5}} & \multirow{2}{*}{1.0} & 7 & 52.97 & 25.52  \\
        ~ & ~ & 15 & 56.76 & \textbf{25.87}\\
        \cline{2-5}
        ~ &  \multirow{2}{*}{0.5} & 7 & 53.06 & 25.05 \\
        ~ & ~ & 15 & 57.75 & 25.48  \\
        \cline{2-5}
        ~ &  \multirow{2}{*}{0.3} & 7 & \textbf{52.61} & 24.32  \\
        ~ & ~ & 15 & 57.87 & 24.83  \\
        \midrule[1pt]
        \multirow{6}{*}{\makecell[c]{PixArt-\\$\Sigma$-XL}} &  \multirow{2}{*}{1.0} & 7 & 63.90 & \textbf{25.80} \\
        ~ & ~ & 15  & 77.55 & 25.06\\
        \cline{2-5}
        ~ &  \multirow{2}{*}{0.5} & 7 & \textbf{62.60} & 25.47 \\
        ~ & ~ & 15 & 82.95 & 24.67 \\
        \cline{2-5}
        ~ &  \multirow{2}{*}{0.3} & 7 & 64.02 & 24.43 \\
        ~ & ~ & 15 & 83.25 & 22.82 \\
        \bottomrule[1.5pt]
    \end{tabular}
    \caption{Performance of \textit{Step AG} under different $w$ settings of different models.}
    \label{tab:abla_w}
\end{table}

A traditional idea of Guidance Scale is that, larger Guidance Scale results in better image-text alignment but worse image quality \cite{ho2022classifier}. But we are surprised to find that sometimes larger Guidance Scales actually decrease both image quality and image-text alignment!(Stable-Diffusion-3, PixArt-$\Sigma$-XL)
    
Further analysis about why guidance scale has such influence is beyond our discussion. We just would like to point out that generally increasing guidance scale tends to heavily harm image quality, so we suggest using guidance scale close to the model's default setting. Using a larger guidance scale, even when \textit{Step AG} is applied, does not help increase generation performance.

\subsection{Video Generation Results}
\label{sec:vid_exp}
\begin{figure}[t]
    \centering
    \begin{subfigure}[b]{0.45\textwidth}
        \includegraphics[width=0.95\textwidth]{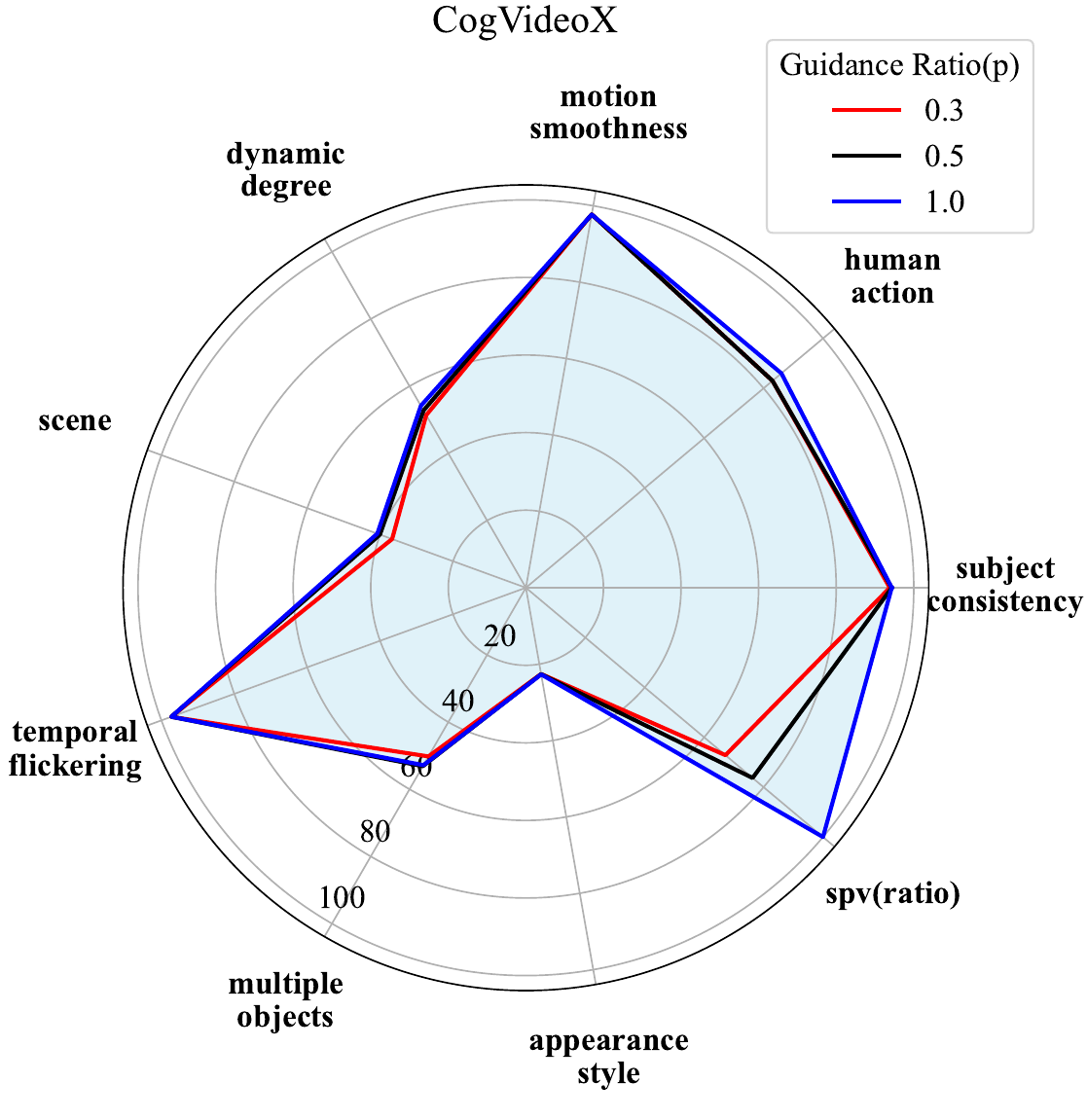}
    \end{subfigure}

    \begin{subfigure}[b]{0.45\textwidth}
        \includegraphics[width=0.95\textwidth]{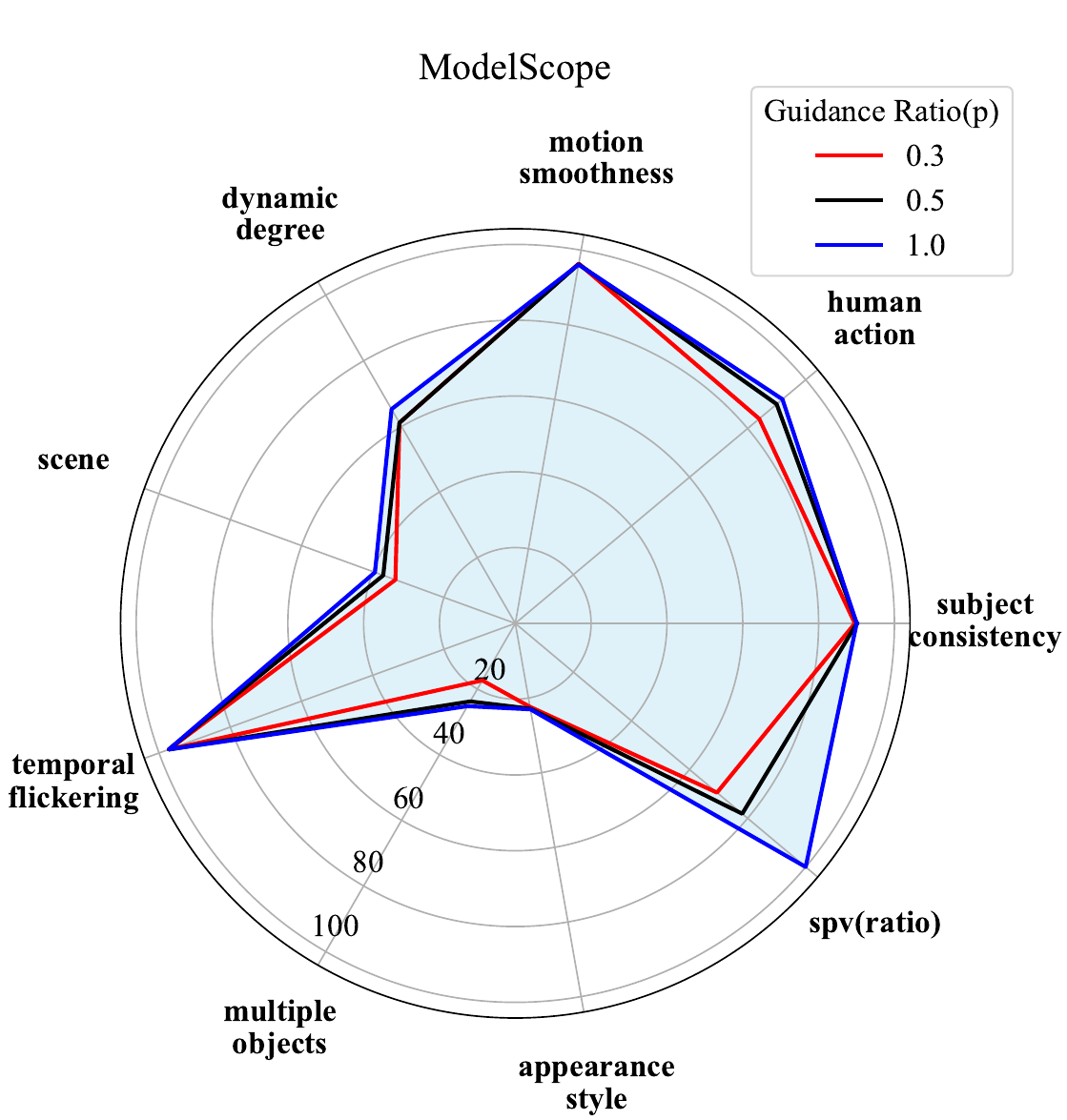}
    \end{subfigure}
    
    \caption{Performance of \textit{Step AG} of different video generation models. SPV(ratio) is compared with spv under $p=1.0$. }
    \label{fig:video}
\end{figure}

VBench includes multiple evaluation perspectives, so we present the complete visualized results in Figure \ref{fig:video}. Similarly, we focus solely on using the conditional score in the later denoising steps. The detailed results can be found in Appendix B.

As can be seen from the results, both $p=0.5$ and $p=0.3$ achieves competitive generation performance, as their curves generally overlap with that of $p=1.0$. A notable difference lies in SPV (ratio), which shows that setting $p=0.5$ only requires 80\% inference time compared with $p=1.0$ while setting $p=0.3$ only requires 70\%, indicating a notable speedup of 20\% to 30\%. This result aligns with the observation in previous image generation experiments and further demonstrates the universal applicability of \textit{Step AG}. Since video generation is generally much slower, the generalization of \textit{Step AG} on video generation is very exciting.

\begin{figure*}
    \centering
    \begin{subfigure}[b]{0.24\textwidth}
        \includegraphics[width=1.0\textwidth]{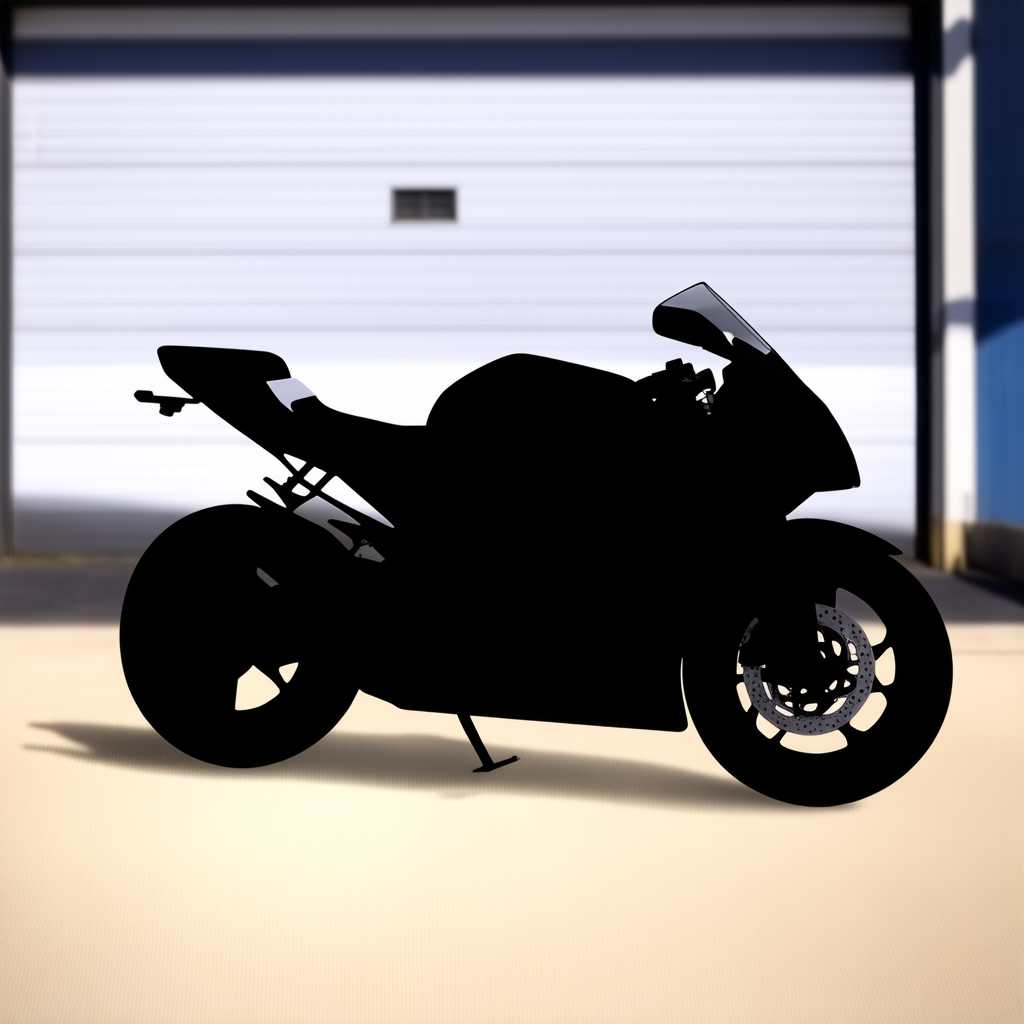}
        \caption*{$T = 12, p=1.0$}
    \end{subfigure}
    \begin{subfigure}[b]{0.24\textwidth}
        \includegraphics[width=1.0\textwidth]{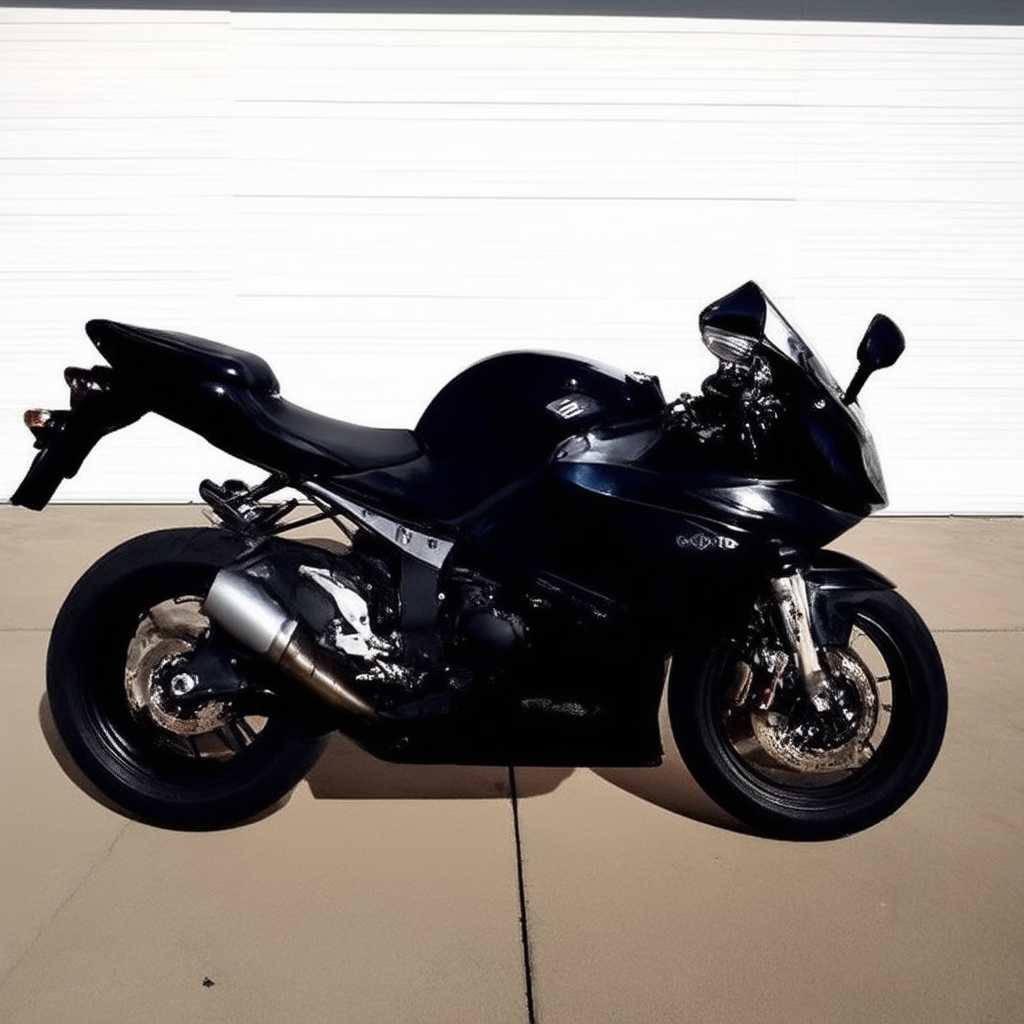}
         \caption*{$T = 19, p=0.3$}
    \end{subfigure}
    \begin{subfigure}[b]{0.24\textwidth}
        \includegraphics[width=1.0\textwidth]{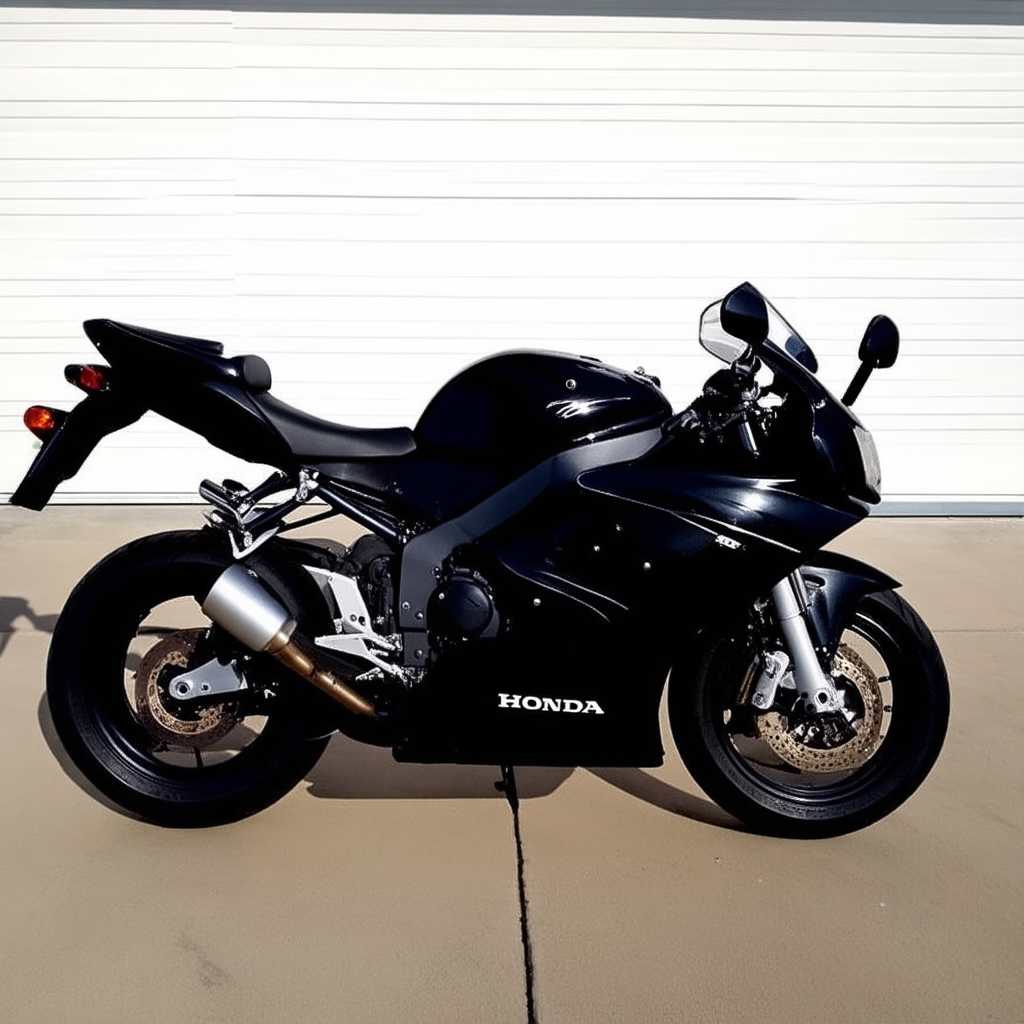}
         \caption*{$T = 19, p=1.0$}
    \end{subfigure}
    \begin{subfigure}[b]{0.24\textwidth}
        \includegraphics[width=1.0\textwidth]{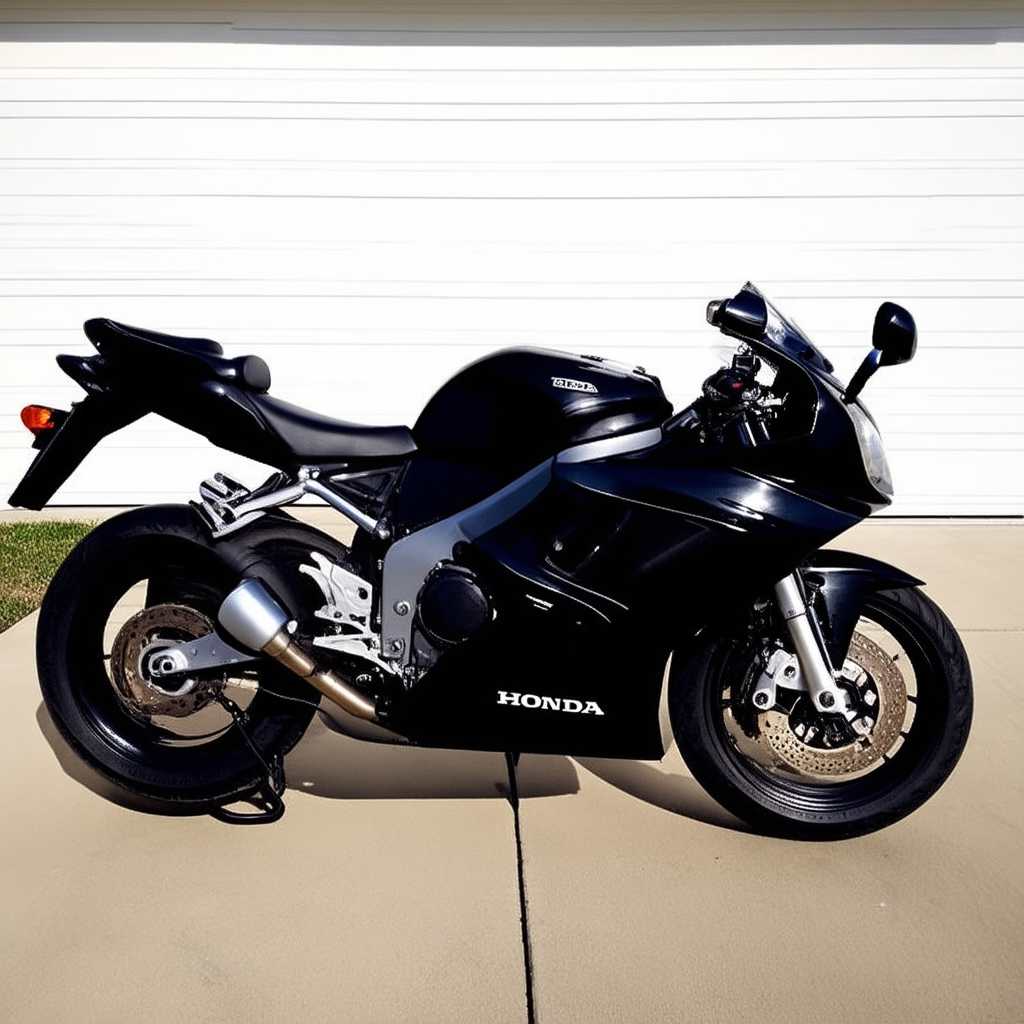}
        \caption*{$T=21, p=1.0$}
    \end{subfigure}
    
    \begin{subfigure}[b]{0.24\textwidth}
        \includegraphics[width=1.0\textwidth]{images/i5/28_0.5.jpg}
        \caption*{$T=28, p=0.5$}
    \end{subfigure}
    \begin{subfigure}[b]{0.24\textwidth}
        \includegraphics[width=1.0\textwidth]{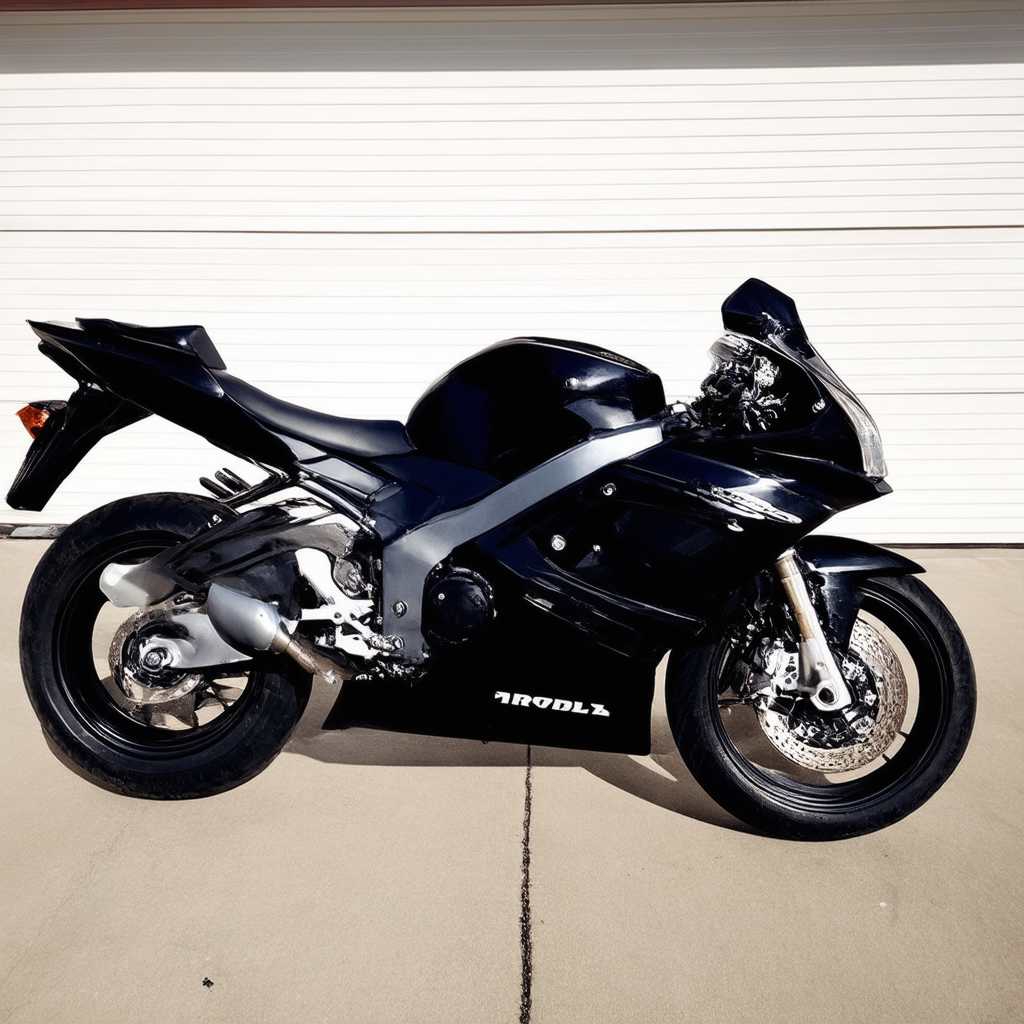}
        \caption*{$T=28, p=0.5$, Uncond}
    \end{subfigure}
    \begin{subfigure}[b]{0.24\textwidth}
        \includegraphics[width=1.0\textwidth]{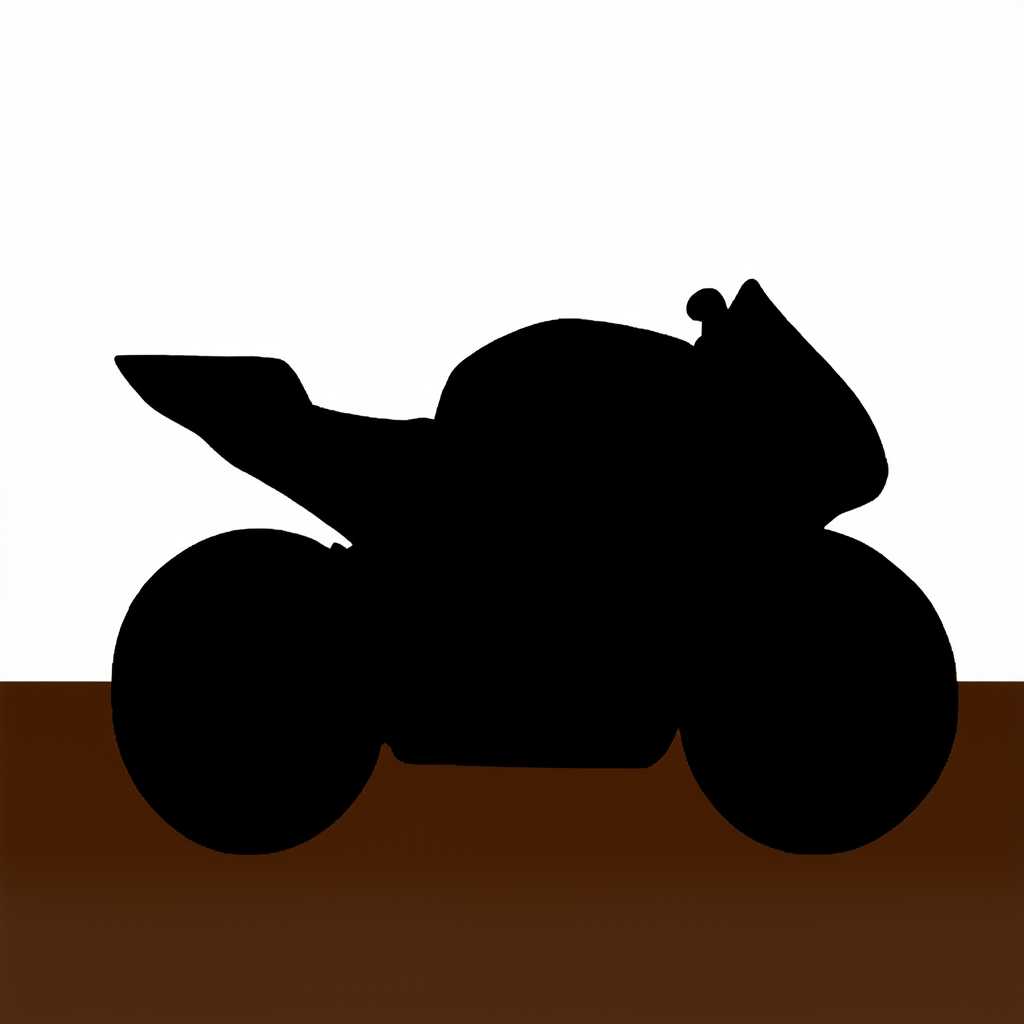}
        \caption*{$T=28, p=0.5, w=15$}
    \end{subfigure}
    \begin{subfigure}[b]{0.24\textwidth}
        \includegraphics[width=1.0\textwidth]{images/i5/28_1.0.jpg}
        \caption*{$T=28, p=1.0$}
    \end{subfigure}
    \caption{`` A black Honda motorcycle parked in front of a garage. " generated by Stable-Diffusion-3 under different settings. Our default setting is $w=7$ and using conditional score in the later denoising steps. Images with different settings are specified.}
    \label{fig:example}
\end{figure*}

\subsection{Qualitative Analysis}
\label{sec:case}

We would like to present some qualitative examples in Figure \ref{fig:example}. More examples can be found in Appendix C.

First, we analyze the behavior of \textit{Step AG}. As observed, applying \textit{Step AG} tends to preserve the model’s behavior, yielding visually similar generation results across different $p$ values when $T$ is kept constant. This outcome illustrates the effectiveness of \textit{Step AG} and aligns with our insight: when \textit{SNR} $\lambda_t$ is large, the final output resides within a limited range, resulting in similar sampling outcomes. However, we also observe that reducing the total inference steps does not exhibit this property—decreasing the number of inference steps notably alters the final sampling results, impacting elements such as shapes, content, and style. For instance, a patch of grass appears in the generated image at $T=21$, but is absent at $T=28$ and $T=19$. This observation underscores the stability of \textit{Step AG}.

Secondly, as is widely acknowledged, reducing the total number of inference steps too aggressively can significantly degrade performance. Although the total number of forward passes under $T=12, p=1.0$ and $T=19, p=0.3$ are the same, their performance differs greatly—setting $T=12$ loses almost all detail, rendering only the basic shape of a motorcycle, whereas setting $T=19, p=0.3$ achieves similar quality to $T=19, p=1.0$. This result further demonstrates the feasibility of \textit{Step AG}. It can not only be combined with a reduced number of inference steps, but can also be applied in cases where reducing steps is not an option.

Thirdly, by comparing the generated samples produced using conditional and unconditional scores, we observe that these samples are also visually similar. This result further supports our earlier analysis that using either conditional or unconditional score in the later denoising steps is acceptable. However, as we have discussed earlier, since we also observe that using the unconditional score may sometimes degrade image-text alignment, we still recommend using the conditional score.

Lastly, applying an excessively large $w$ does not enhance generation performance, even when classifier-free guidance is omitted in the later steps of the denoising process. It is natural to wonder whether using a larger guidance scale would improve generation performance since classifier-free guidance is applied in less steps. However, as observed, since \textit{Step AG} tends to preserve model behavior, if a large $w$ yields poor results without \textit{Step AG}, it will not improve performance after \textit{Step AG} is applied. So simply using the same optimal $w$ as \textit{Step AG} is not applied is enough.
 
\section{Conclusion}
In this work, we highlight the limitations of previous adaptive guidance strategy and propose 
a new adaptive guidance strategy \textit{Step AG} with more solid basis, indicating only the first 30\% to 50\% steps in classifier-free guidance is essential. Our empirical results demonstrate that \textit{Step AG} is a universally applicable adaptive guidance strategy that reduces inference costs by 20\% to 30\% with minimal generation performance decrease. We show that \textit{Step AG} performs effectively across various settings. Furthermore, we validate the effectiveness of \textit{Step AG} in video generation tasks, underscoring its applicability. We believe that \textit{Step AG} can be directly implemented in many text-to-vision diffusion systems to achieve significant acceleration, potentially advancing the use of text-to-vision diffusion models in efficiency-focused applications.

\bibliographystyle{ACM-Reference-Format}
\bibliography{main}

\appendix
\section{Detailed Experiment Setup}
\label{sec:app-exp-setup}
We list detailed experiment settings including pretrained-weights, number of inference steps and calculation resource as in Table \ref{tab:model_weights}, \ref{tab:exp_setup}. Since our target is not comparing between different models, we do not intentionally maintain the same setting across different models to further demonstrate \textit{Step AG} works properly under various settings.

\begin{table}[htbp]
    \centering
    \begin{tabular}{c|c}
        \hline
        Model Name & Pretrained Weights \\
        \hline
        Stable-Diffusion-3 & \makecell[c]{Stable-Diffusion-3-\\Medium-Diffusers} \footnotemark[1] \\
        \hline
        Stable-Diffusion-XL & Stable-Diffusion-XL-base-1.0 \footnotemark[2] \\
        \hline
        Stable-Diffusion-1.5 & Stable-Diffusion-v1-5 \footnotemark[3] \\
        \hline
        PixArt-$\Sigma$-XL & PixArt-Sigma-XL-2-1024-MS \footnotemark[4] \\
        \hline
        CogVideoX & CogVideoX-5B \footnotemark[5] \\
        \hline
        ModelScope & text-to-video-ms-1.7b \footnotemark[6] \\
        \hline
    \end{tabular}
    \caption{Details of our model weights.}
    \label{tab:model_weights}
\end{table}
\footnotetext[1]{\url{https://huggingface.co/stabilityai/stable-diffusion-3-medium-diffusers}}
\footnotetext[2]{\url{https://huggingface.co/stabilityai/stable-diffusion-xl-base-1.0}}
\footnotetext[3]{\url{https://huggingface.co/stable-diffusion-v1-5/stable-diffusion-v1-5}}
\footnotetext[4]{\url{https://huggingface.co/PixArt-alpha/PixArt-Sigma-XL-2-1024-MS}}
\footnotetext[5]{\url{https://huggingface.co/THUDM/CogVideoX-5b}}
\footnotetext[6]{\url{https://huggingface.co/ali-vilab/text-to-video-ms-1.7b}}
\begin{table}[htbp]
    \centering
    \begin{tabular}{c|cc}
        \hline
        Model Name  & Default $T$ & Calculation Resource \\
        \hline
        Stable-Diffusion-3 & 28 & Nvidia L40 \\
        Stable-Diffusion-XL & 50 & Nvidia RTX 3090 \\
        Stable-Diffusion-1.5 & 50 & Nvidia RTX 3090 \\
        PixArt-$\Sigma$-XL & 20 & Nvidia RTX 3090 \\
        CogVideoX &  50 & Nvidia RTX 3090 \\
        ModelScope &  50 & Nvidia RTX 3090 \\
        \hline
    \end{tabular}
    \caption{Details of our experiment setup.}
    \label{tab:exp_setup}
\end{table}

\section{More Experiment Results and Analysis}
\label{sec:app-results}

\begin{table*}[htbp]
    \centering
    \setlength{\tabcolsep}{4.0mm}
    \begin{tabular}{c|c|c|ccc}
    \toprule[1.5pt]
        Model Name & \makecell[c]{Guidance \\Ratio($p$)} & Score Type & FID($\downarrow$) & CLIP Score($\uparrow$) & SPI(s)($\downarrow$)\\
    \toprule[1.5pt]
        \multirow{5}{*}{Stable-Diffusion-3} & 1.0 & - & 31.29 & \textbf{25.24} & 5.55 \\
        \cline{2-6}
        ~ &  \multirow{2}{*}{0.5}  & Unconditional & \textbf{31.02} & 24.87 & 4.31  \\
        ~ & ~  & Conditional & 31.08 & 25.00 & 4.31  \\
        \cline{2-6}
        ~ & \multirow{2}{*}{0.3}  & Unconditional & 31.75 & 24.11 & 3.79 \\
        ~ & ~ & Conditional & 31.16 & 24.67 & \textbf{3.78}  \\
        \midrule[1.0pt]
        \multirow{5}{*}{Stable-Diffusion-XL} & 1.0 & - & \textbf{53.83} & \textbf{26.02} & 13.54 \\
        \cline{2-6}
        ~ &  \multirow{2}{*}{0.5} &  Unconditional & 53.96 & 25.62 & 10.55  \\
        ~ &~ & Conditional & 53.97 & 25.71 & 10.56 \\
        \cline{2-6}
        ~ & \multirow{2}{*}{0.3} &  Unconditional & 54.28 & 24.82 & \textbf{9.36} \\
        ~ & ~ & Conditional & 54.15 & 25.24 & 9.37  \\
        \midrule[1.0pt]
        \multirow{5}{*}{Stable-Diffusion-1.5} &  1.0 &  - & \textbf{27.25} & \textbf{25.61} & 2.44\\
        \cline{2-6}
        ~ &  \multirow{2}{*}{0.5} &  Unconditional & 28.07 & 24.93 & 1.99 \\
        ~ &  ~ & Conditional & 27.75 & 25.12 & 1.99\\
        \cline{2-6}
        ~ & \multirow{2}{*}{0.3}  & Unconditional & 31.67 & 23.45 & \textbf{1.82} \\
        ~ &  ~ & Conditional & 28.46 & 24.40 & \textbf{1.82}  \\
        \midrule[1.0pt]
        \multirow{5}{*}{PixArt-$\Sigma$-XL} & 1.0 & - & 42.21 & \textbf{24.13} & 6.01 \\
        \cline{2-6}
        ~ &  \multirow{2}{*}{0.5}  & Unconditional & 40.92 & 23.71 & 4.67 \\
        ~ &  ~ & Conditional & \textbf{39.63} & 23.91 & 4.67 \\
        \cline{2-6}
        ~ & \multirow{2}{*}{0.3} &  Unconditional & 52.83 & 21.17 & 4.15  \\
        ~ &  ~ & Conditional & 40.00 & 23.16 & \textbf{4.14}   \\
        \bottomrule[1.5pt]
    \end{tabular}
    
    \caption{Evaluation results of \textit{Step AG} under different models and settings. Note that there is no difference between using conditional and unconditional scores under $p=1.0$.}
    \label{tab:imagenet_exp}
\end{table*}

\paragraph{Results on ImageNet}
\label{sec:app-imagenet}
To show the effectiveness of \textit{Step AG} more comprehensively, we also conduct experiments on ImageNet \cite{deng2009imagenet} dataset. The settings remain the same as in Section 4. Since ImageNet only contains image category instead of description, we use ``\texttt{A photo of \{category\}}" as the corresponding prompt. The results are shown in Table \ref{tab:imagenet_exp}. The results remain consistent with previous experiments that \textit{Step AG} offers notable speedup with little harm on generation performance, which shows the superiority of our method.

\paragraph{Video Generation Results}
We list detailed video generation results in Table \ref{tab:vid-cog} and \ref{tab:vid-ms}.

\begin{table*}
\setlength{\tabcolsep}{1.8mm}
    \begin{tabular}{c|ccccccccc}
        \hline
        \makecell[c]{Guidance \\ Ratio ($p$)} &  \makecell[c]{Subject \\ Consistency} & \makecell[c]{Human \\Action} & \makecell[c]{Motion \\ Smoothness} & \makecell[c]{Dynamic \\ Degree} & \makecell[c]{Scene} & \makecell[c]{Temporal \\ Flickering} & \makecell[c]{Multiple \\ Objects} & \makecell[c]{Appearance \\ Style} & SPV(s)($\downarrow$) \\
        \hline
        1.0 & \textbf{94.33} & \textbf{86.00} & 97.71 & \textbf{54.17} & 36.70 & 97.26 & 50.22 & \textbf{22.69} & 601.70 \\
        0.5 & 94.32 & 83.00 & 97.73 & 52.78 & 40.04 & 97.27 & \textbf{53.12} & 22.64 & 459.13 \\
        0.3 & 93.81 & 83.00 & \textbf{97.74} & 51.39 & \textbf{40.77} & \textbf{97.29} & 52.82 & 22.59 & \textbf{404.16} \\
        \hline
    \end{tabular}
    \caption{Video generation results of CogVideoX.}
    \label{tab:vid-cog}
\end{table*}

\begin{table*}
\setlength{\tabcolsep}{1.8mm}
    \begin{tabular}{c|ccccccccc}
        \hline
        \makecell[c]{Guidance \\ Ratio ($p$)} &  \makecell[c]{Subject \\ Consistency} & \makecell[c]{Human \\Action} & \makecell[c]{Motion \\ Smoothness} & \makecell[c]{Dynamic \\ Degree} & \makecell[c]{Scene} & \makecell[c]{Temporal \\ Flickering} & \makecell[c]{Multiple \\ Objects} & \makecell[c]{Appearance \\ Style} & SPV(s)($\downarrow$) \\
        \hline
        1.0 & 90.05 & \textbf{92.00} & 96.13 & \textbf{65.27} & \textbf{39.39} & 97.26 & \textbf{25.22} & \textbf{23.02} & 12.77 \\
        0.5 & \textbf{90.06} & 90.00 & 96.25 & 61.11 & 37.06 & 97.25 & 23.78 & 22.86 & 9.98 \\
        0.3 & 89.56 & 84.00 & \textbf{96.40} & 61.11 & 33.65 & \textbf{97.29} & 17.38 & 22.36 & \textbf{8.87} \\
        \hline
    \end{tabular}
    \caption{Video generation results of ModelScope.}
    \label{tab:vid-ms}
\end{table*}

\section{More Qualitative Results}
\label{sec:app-case}
\begin{figure*}[htbp]
    \centering
    \begin{subfigure}[b]{0.24\textwidth}
        \includegraphics[width=1.0\textwidth]{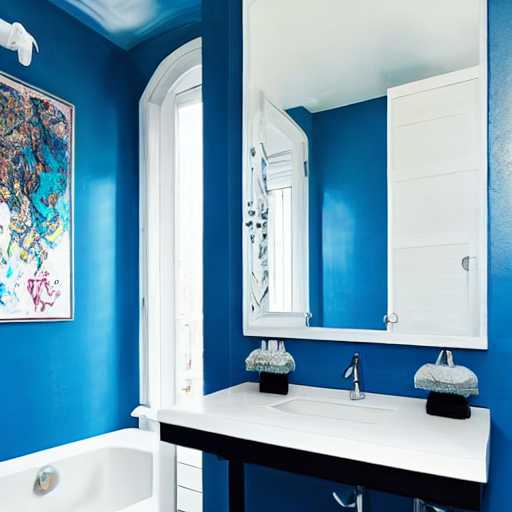}
        \caption{$p=0.3, w=7$, Uncond}
    \end{subfigure}
    \begin{subfigure}[b]{0.24\textwidth}
        \includegraphics[width=1.0\textwidth]{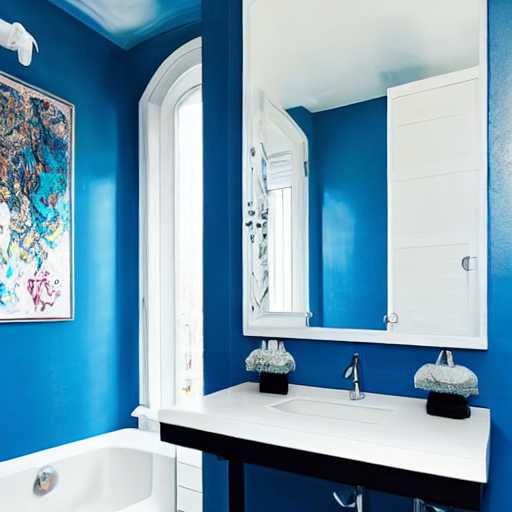}
        \caption{$p=0.3, w=7$, Cond}
    \end{subfigure}
    \begin{subfigure}[b]{0.24\textwidth}
        \includegraphics[width=1.0\textwidth]{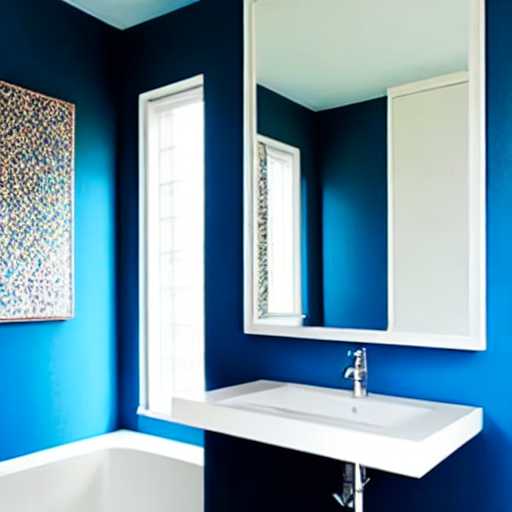}
        \caption{$p=0.3, w=15$, Cond}
    \end{subfigure}
    \begin{subfigure}[b]{0.24\textwidth}
        \includegraphics[width=1.0\textwidth]{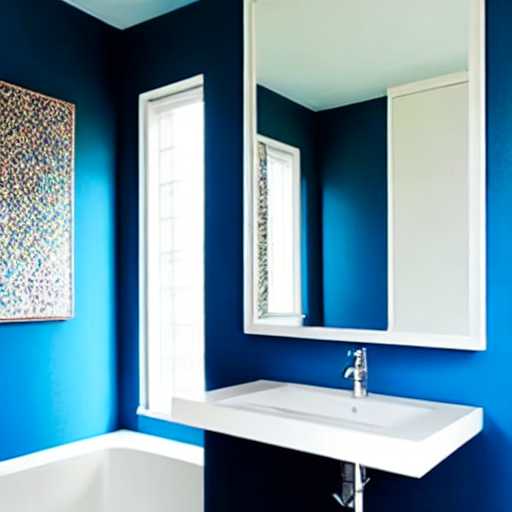}
        \caption{$p=0.3, w=15$, Uncond}
    \end{subfigure}
    \begin{subfigure}[b]{0.24\textwidth}
        \includegraphics[width=1.0\textwidth]{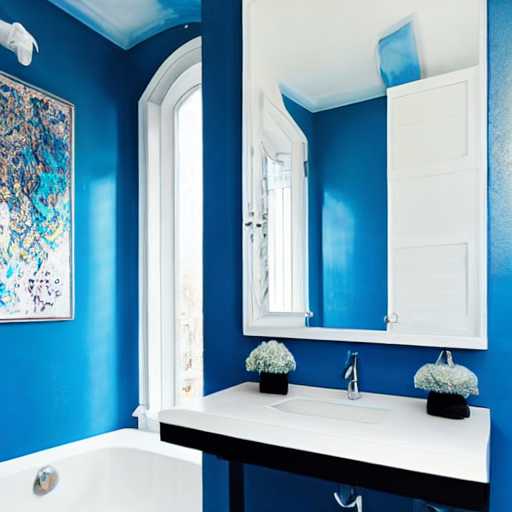}
        \caption{$p=0.5, w=7$, Uncond}
    \end{subfigure}
    \begin{subfigure}[b]{0.24\textwidth}
        \includegraphics[width=1.0\textwidth]{images/i6/0.5_cond_7.jpg}
        \caption{$p=0.5, w=7$, Cond}
    \end{subfigure}
    \begin{subfigure}[b]{0.24\textwidth}
        \includegraphics[width=1.0\textwidth]{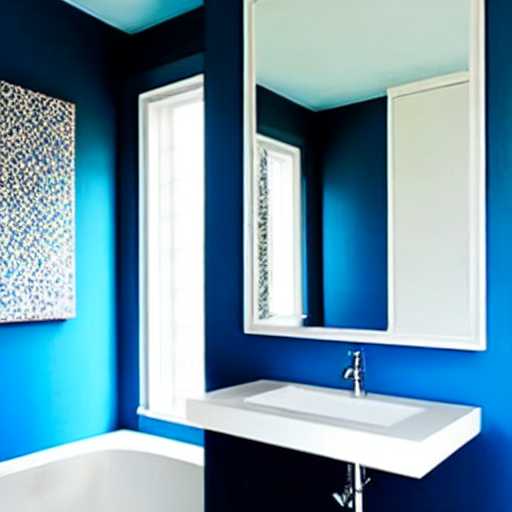}
        \caption{$p=0.5, w=15$, Cond}
    \end{subfigure}
    \begin{subfigure}[b]{0.24\textwidth}
        \includegraphics[width=1.0\textwidth]{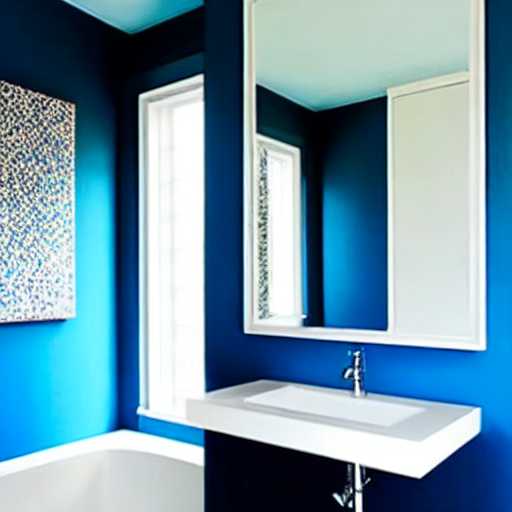}
        \caption{$p=0.5, w=15$, Uncond}
    \end{subfigure}
    \begin{subfigure}[b]{0.24\textwidth}
        \includegraphics[width=1.0\textwidth]{images/i6/1.0_cond_7.jpg}
        \caption{$p=1.0, w=7$}
    \end{subfigure}
    \begin{subfigure}[b]{0.24\textwidth}
        \includegraphics[width=1.0\textwidth]{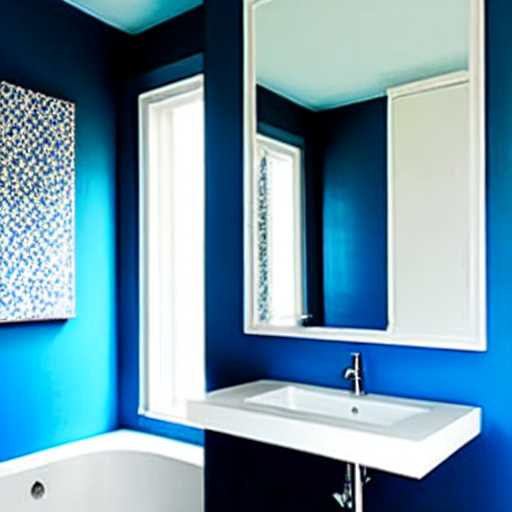}
        \caption{$p=1.0, w=15$}
    \end{subfigure}
    \caption{``A room with blue walls and a white sink and door" generated by Stable-Diffusion-1.5 under different settings. }
    \label{fig:sd15-case}
\end{figure*}

\begin{figure*}[htbp]
    \centering
    \begin{subfigure}[b]{0.24\textwidth}
        \includegraphics[width=1.0\textwidth]{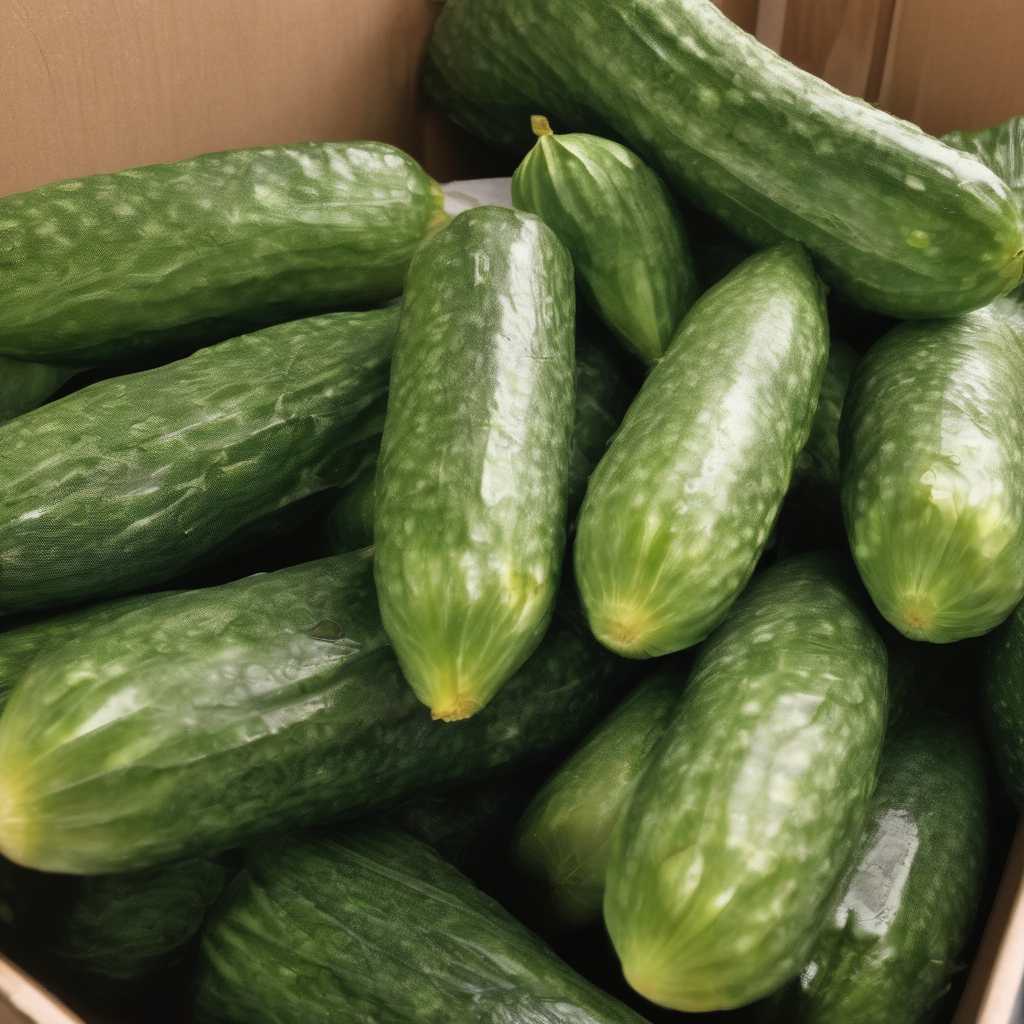}
        \caption{$p=0.3, w=7$, Uncond}
    \end{subfigure}
    \begin{subfigure}[b]{0.24\textwidth}
        \includegraphics[width=1.0\textwidth]{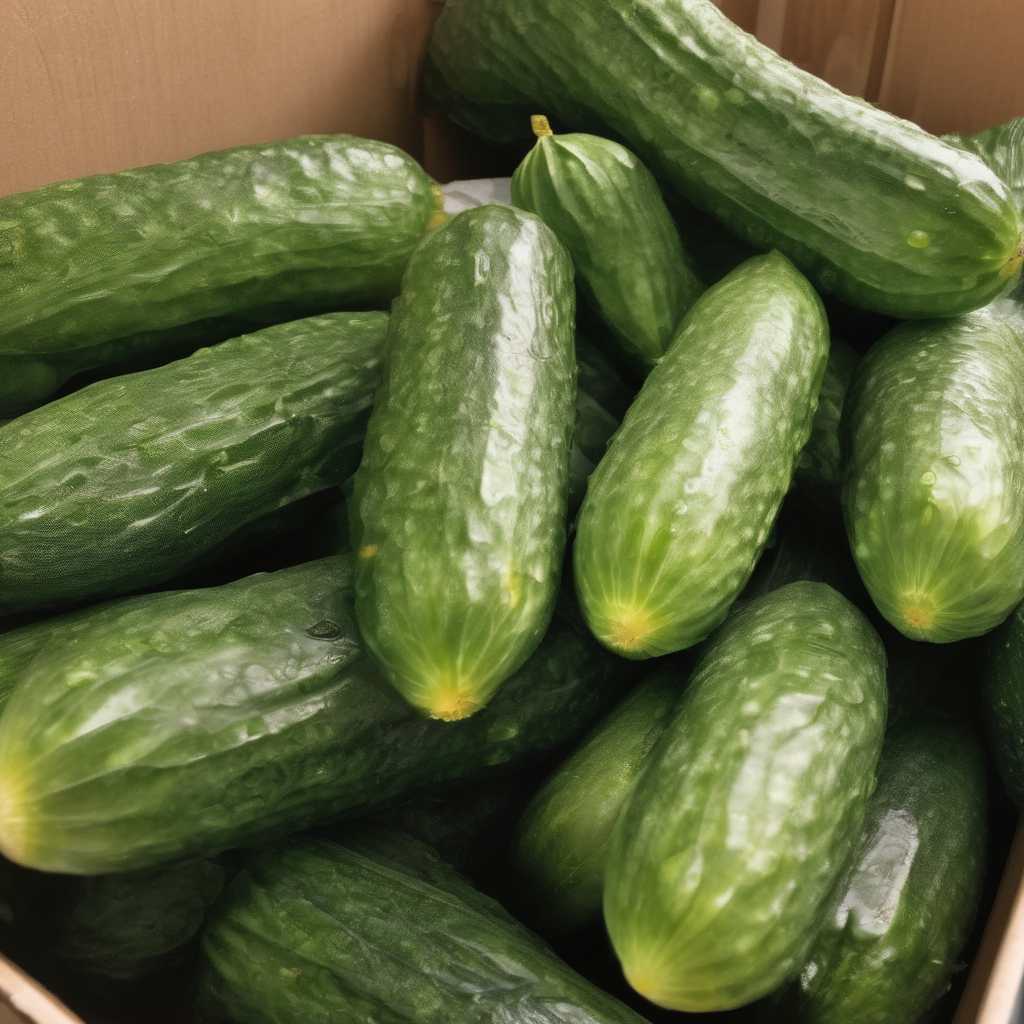}
        \caption{$p=0.3, w=7$, Cond}
    \end{subfigure}
    \begin{subfigure}[b]{0.24\textwidth}
        \includegraphics[width=1.0\textwidth]{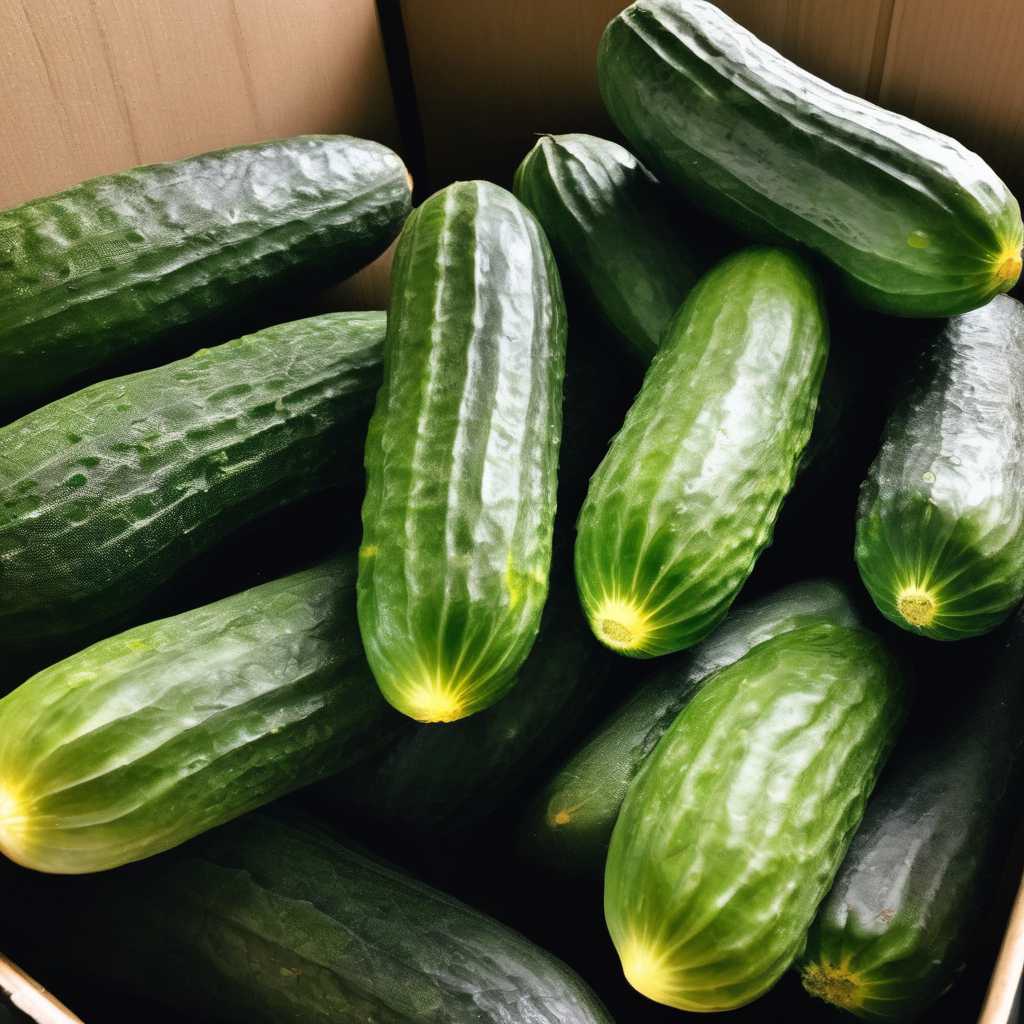}
        \caption{$p=0.3, w=15$, Cond}
    \end{subfigure}
    \begin{subfigure}[b]{0.24\textwidth}
        \includegraphics[width=1.0\textwidth]{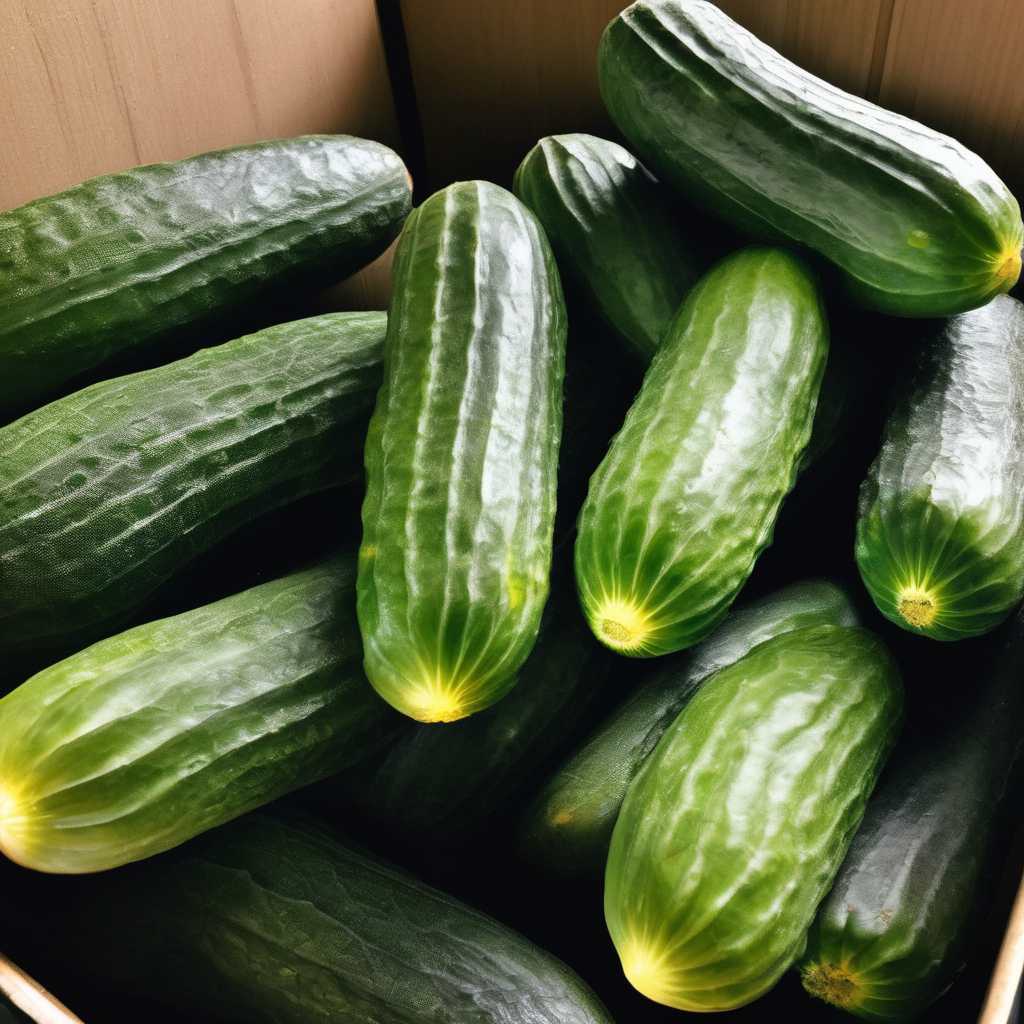}
        \caption{$p=0.3, w=15$, Uncond}
    \end{subfigure}
    \begin{subfigure}[b]{0.24\textwidth}
        \includegraphics[width=1.0\textwidth]{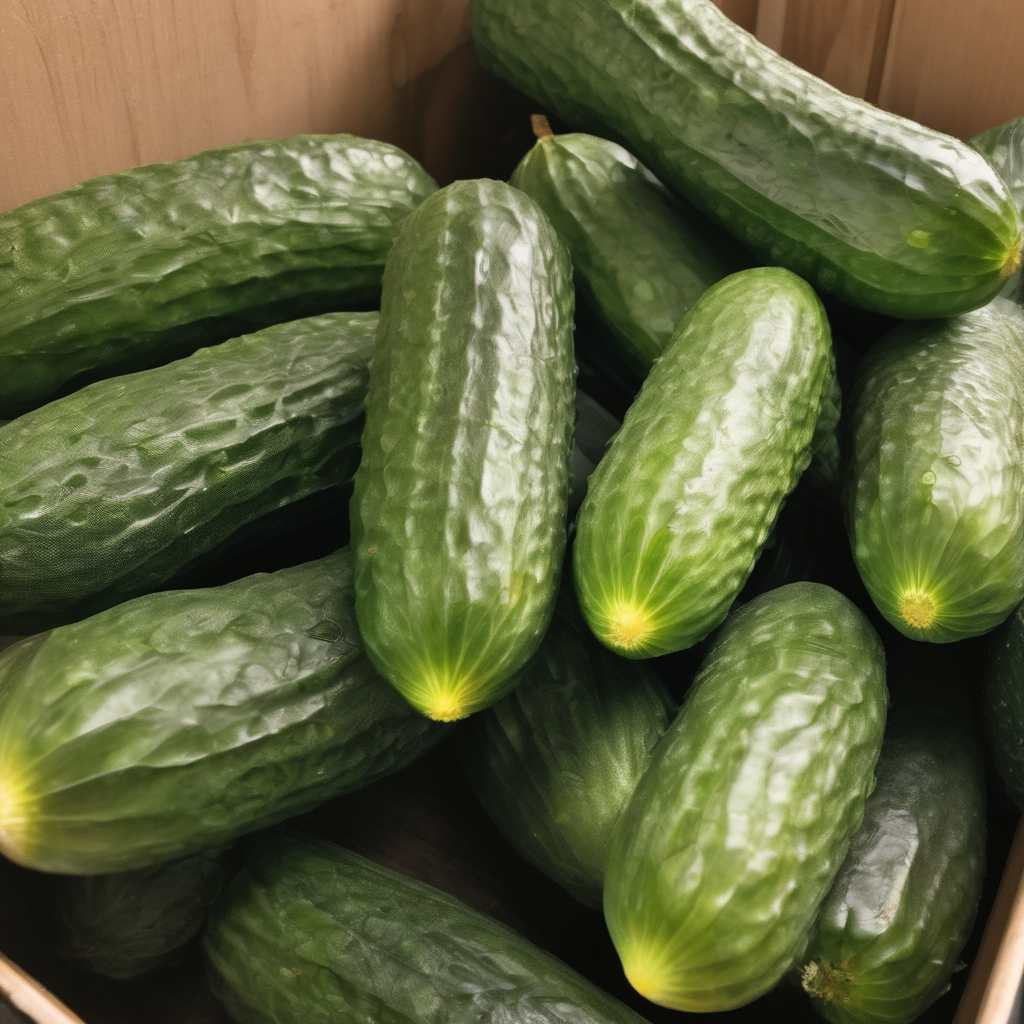}
        \caption{$p=0.5, w=7$, Uncond}
    \end{subfigure}
    \begin{subfigure}[b]{0.24\textwidth}
        \includegraphics[width=1.0\textwidth]{images/i7/0.5_cond_7.jpg}
        \caption{$p=0.5, w=7$, Cond}
    \end{subfigure}
    \begin{subfigure}[b]{0.24\textwidth}
        \includegraphics[width=1.0\textwidth]{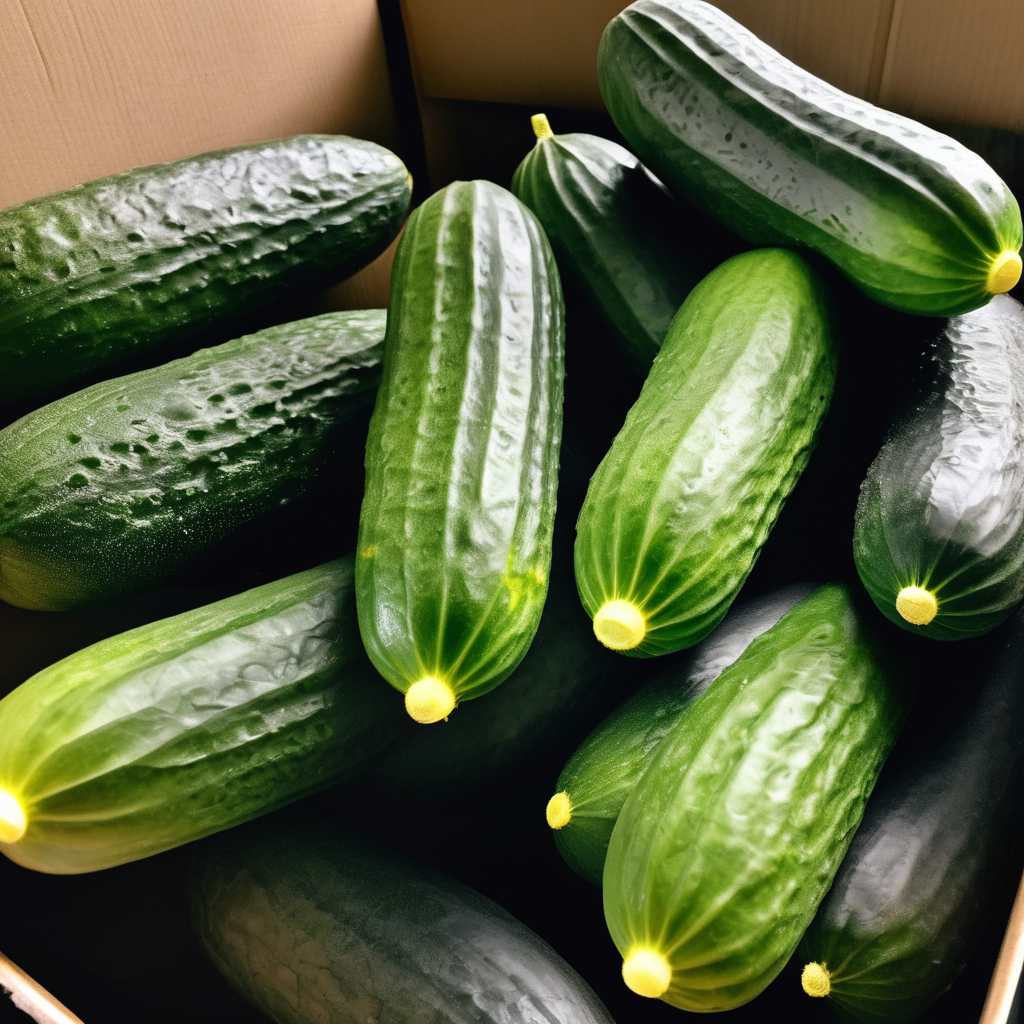}
        \caption{$p=0.5, w=15$, Cond}
    \end{subfigure}
    \begin{subfigure}[b]{0.24\textwidth}
        \includegraphics[width=1.0\textwidth]{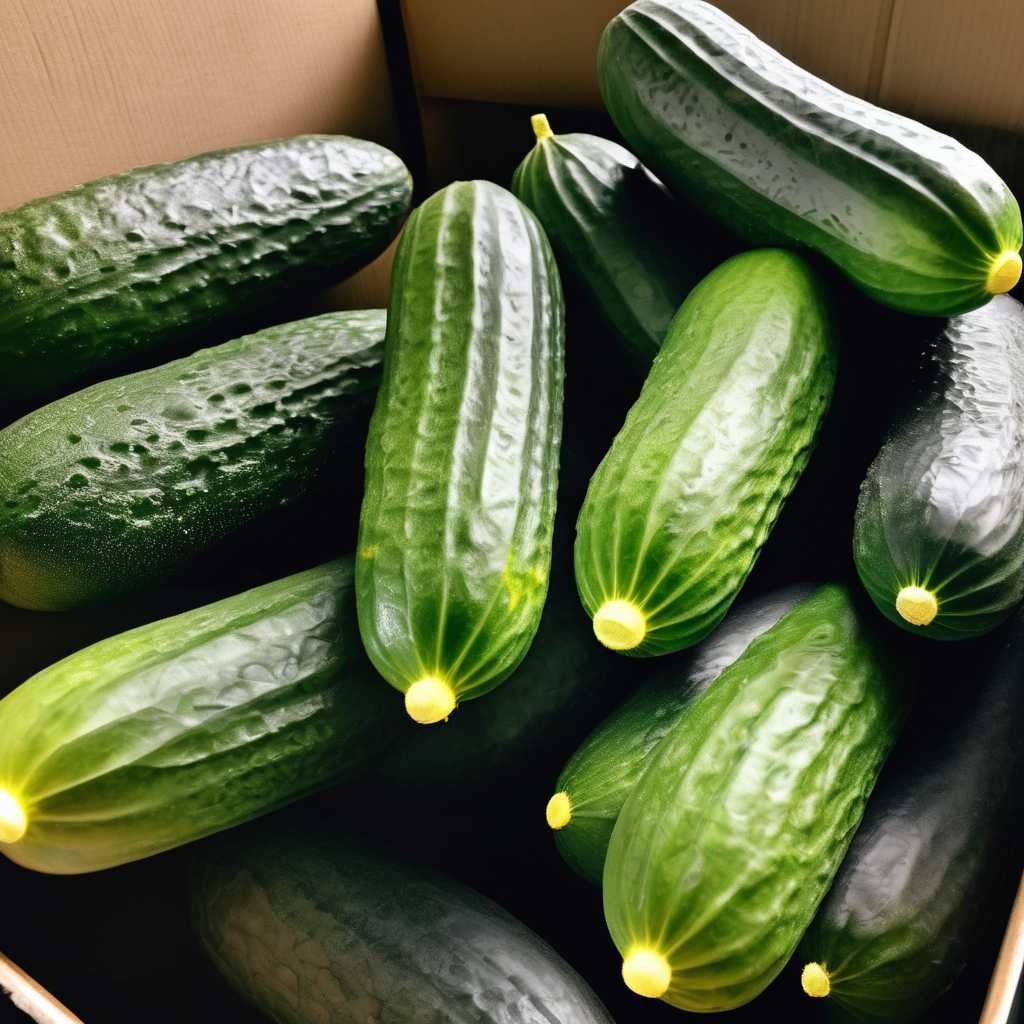}
        \caption{$p=0.5, w=15$, Uncond}
    \end{subfigure}
    \begin{subfigure}[b]{0.24\textwidth}
        \includegraphics[width=1.0\textwidth]{images/i7/1.0_7.jpg}
        \caption{$p=1.0, w=7$}
    \end{subfigure}
    \begin{subfigure}[b]{0.24\textwidth}
        \includegraphics[width=1.0\textwidth]{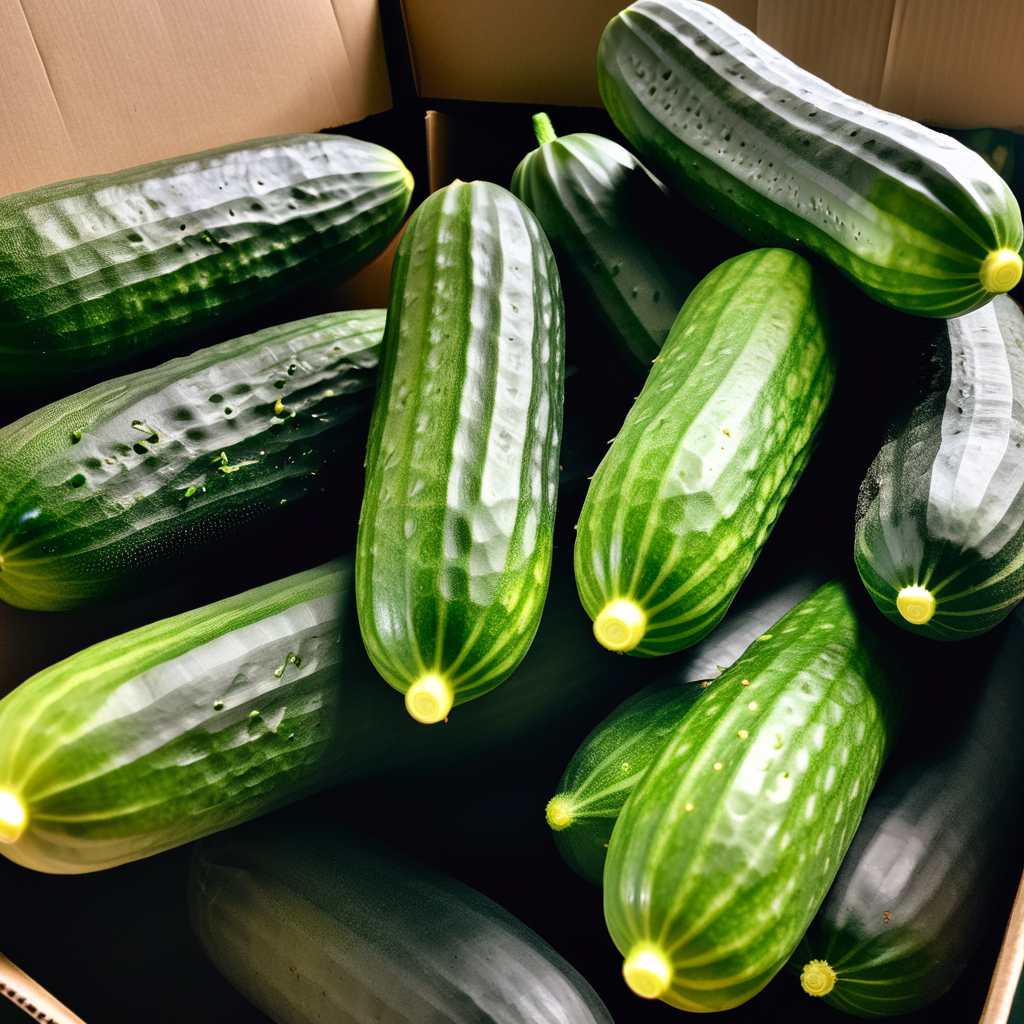}
        \caption{$p=1.0, w=15$}
    \end{subfigure}
    \caption{``This is an open box containing four cucumbers" generated by Stable-Diffusion-XL under different settings. }
    \label{fig:sdxl-case}
\end{figure*}

\begin{figure*}[htbp]
    \centering
    \begin{subfigure}[b]{0.24\textwidth}
        \includegraphics[width=1.0\textwidth]{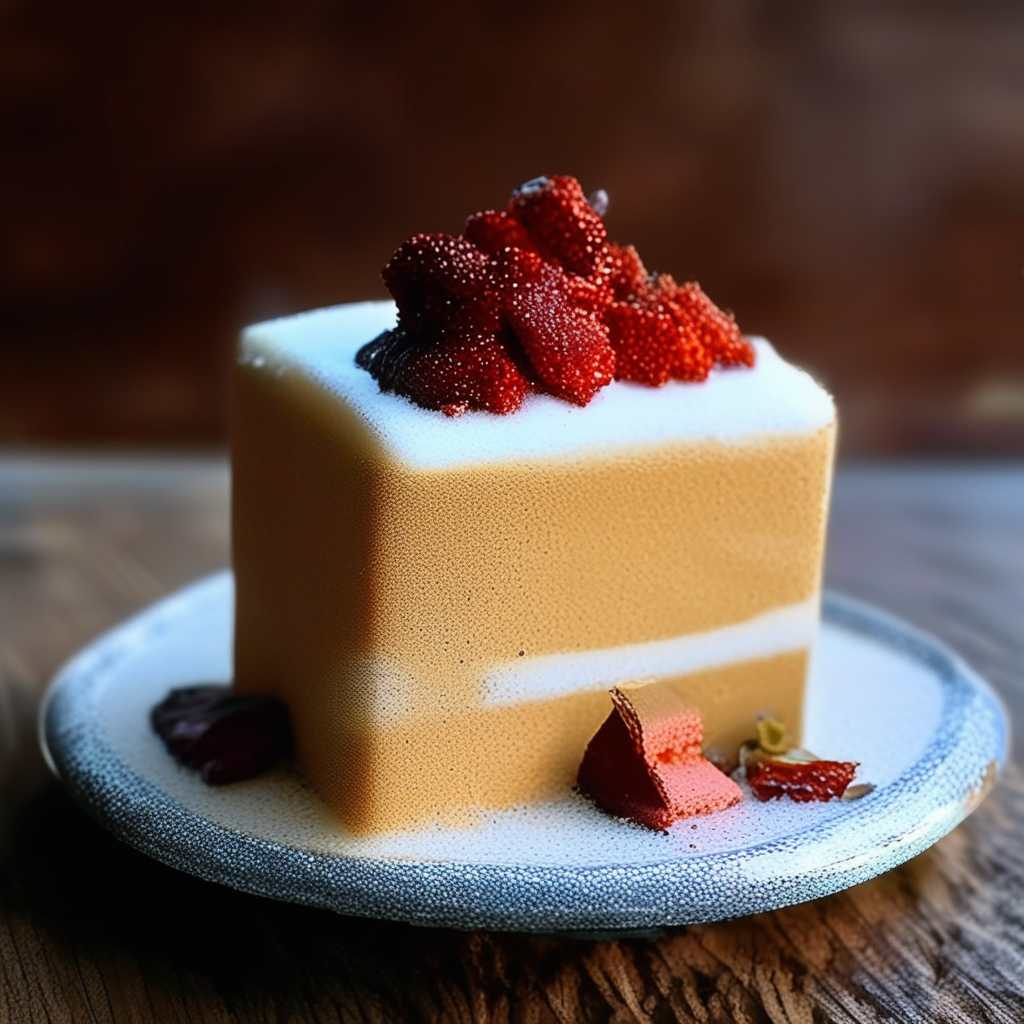}
        \caption{$p=0.3, w=7$, Uncond}
    \end{subfigure}
    \begin{subfigure}[b]{0.24\textwidth}
        \includegraphics[width=1.0\textwidth]{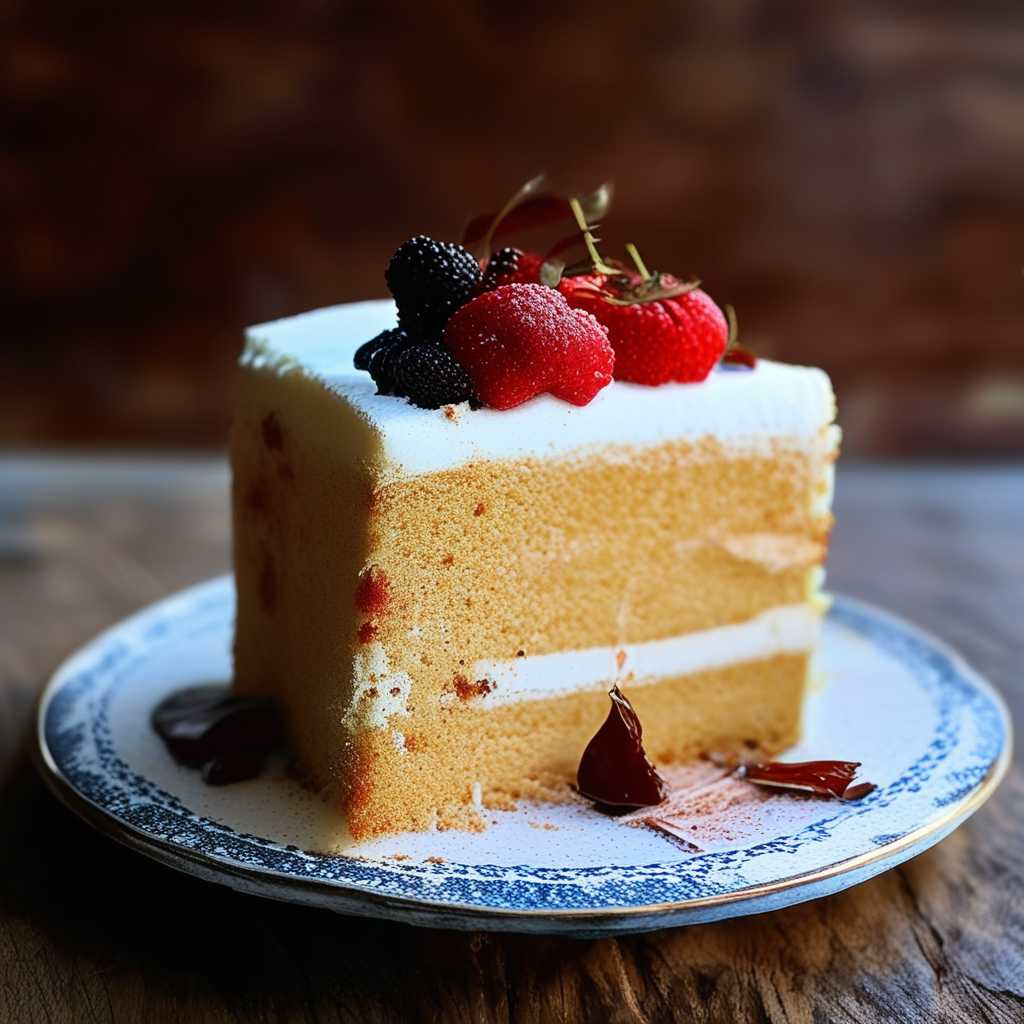}
        \caption{$p=0.3, w=7$, Cond}
    \end{subfigure}
    \begin{subfigure}[b]{0.24\textwidth}
        \includegraphics[width=1.0\textwidth]{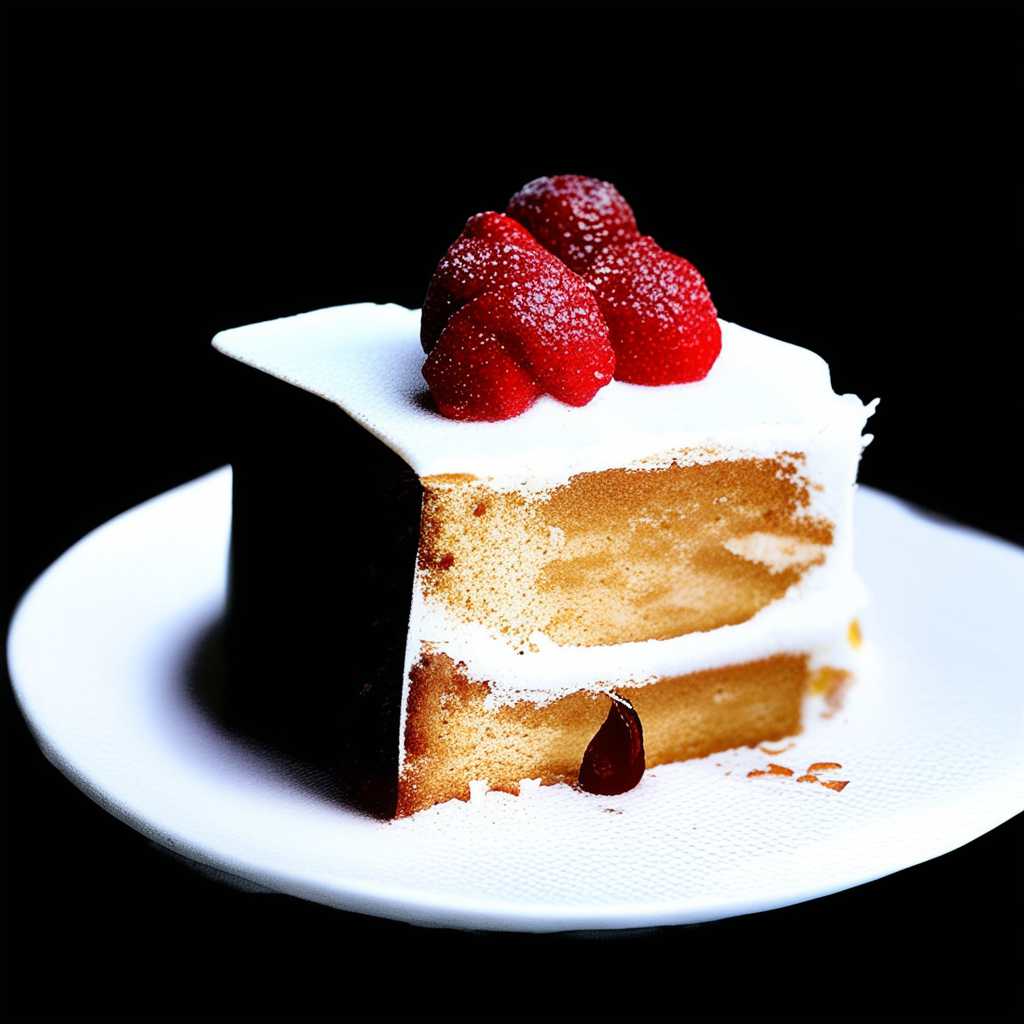}
        \caption{$p=0.3, w=15$, Cond}
    \end{subfigure}
    \begin{subfigure}[b]{0.24\textwidth}
        \includegraphics[width=1.0\textwidth]{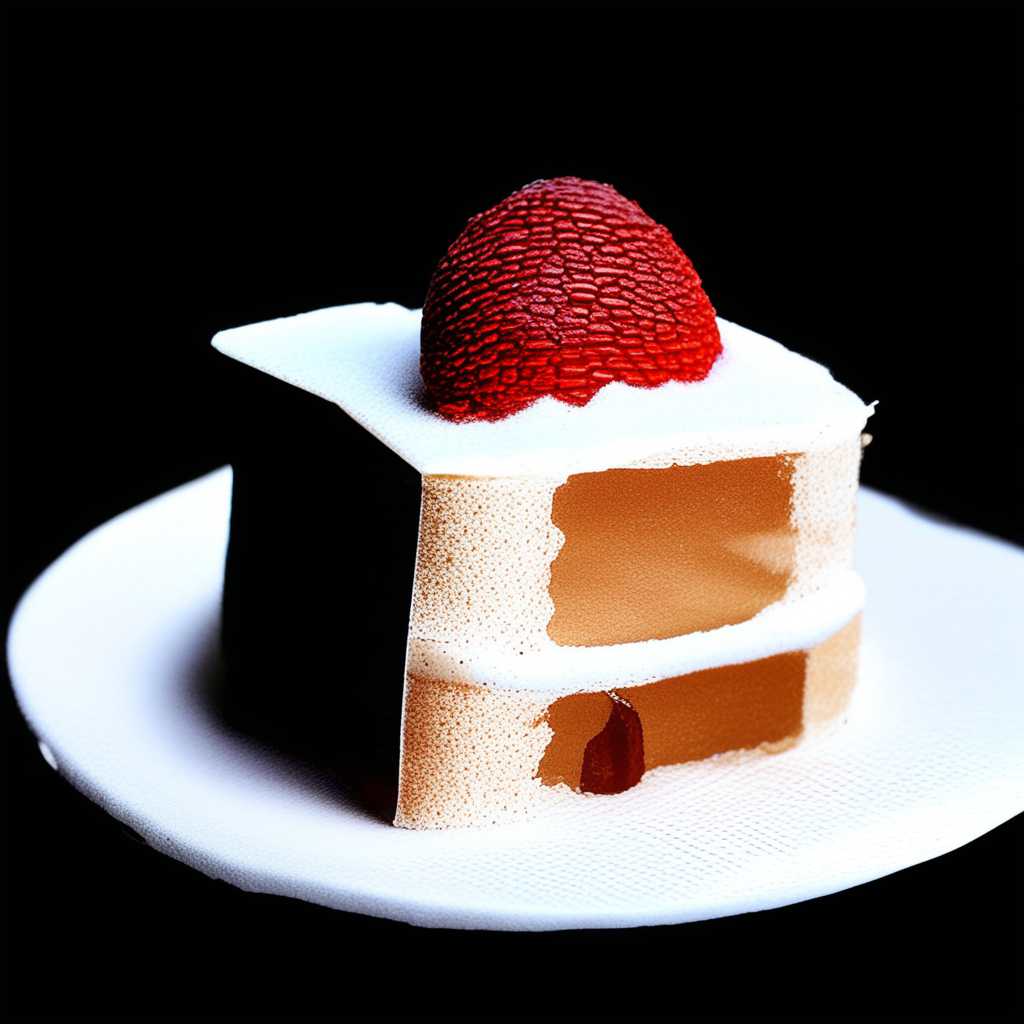}
        \caption{$p=0.3, w=15$, Uncond}
    \end{subfigure}
    \begin{subfigure}[b]{0.24\textwidth}
        \includegraphics[width=1.0\textwidth]{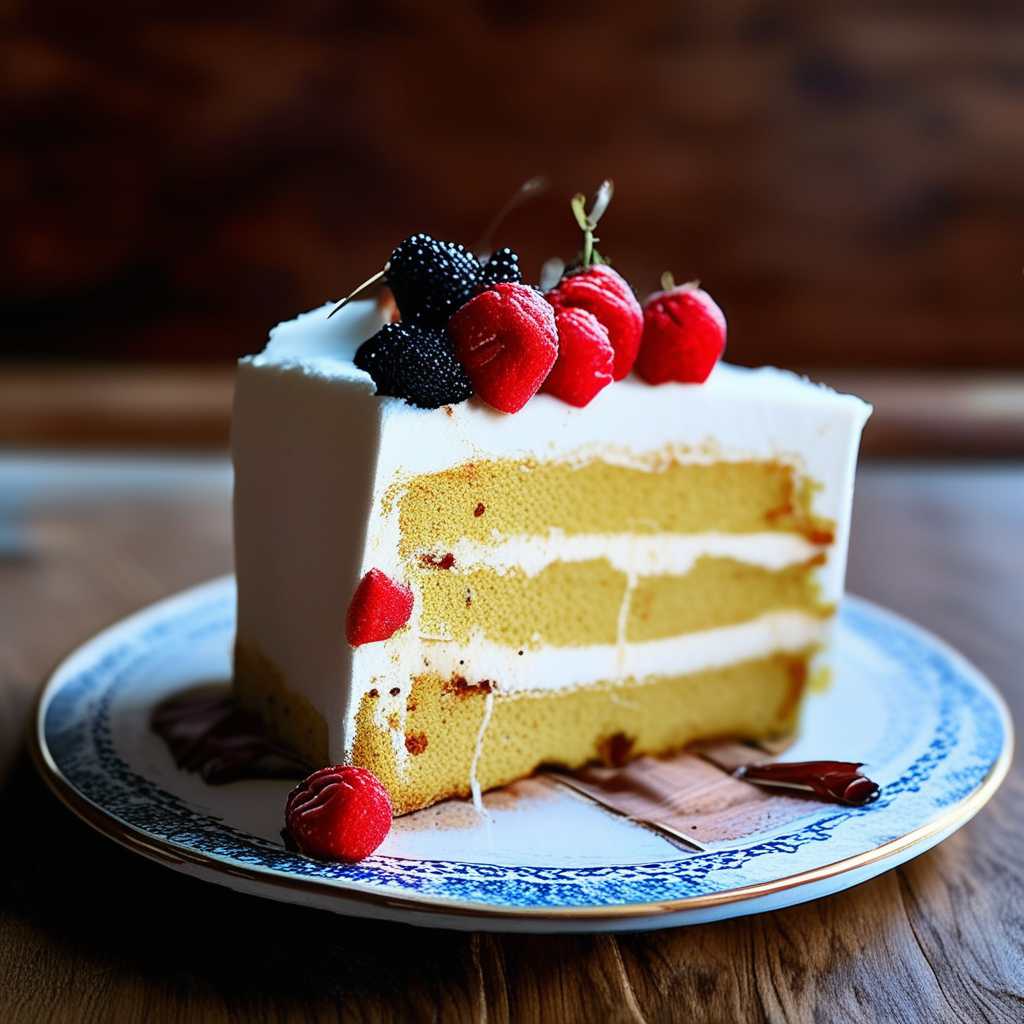}
        \caption{$p=0.5, w=7$, Uncond}
    \end{subfigure}
    \begin{subfigure}[b]{0.24\textwidth}
        \includegraphics[width=1.0\textwidth]{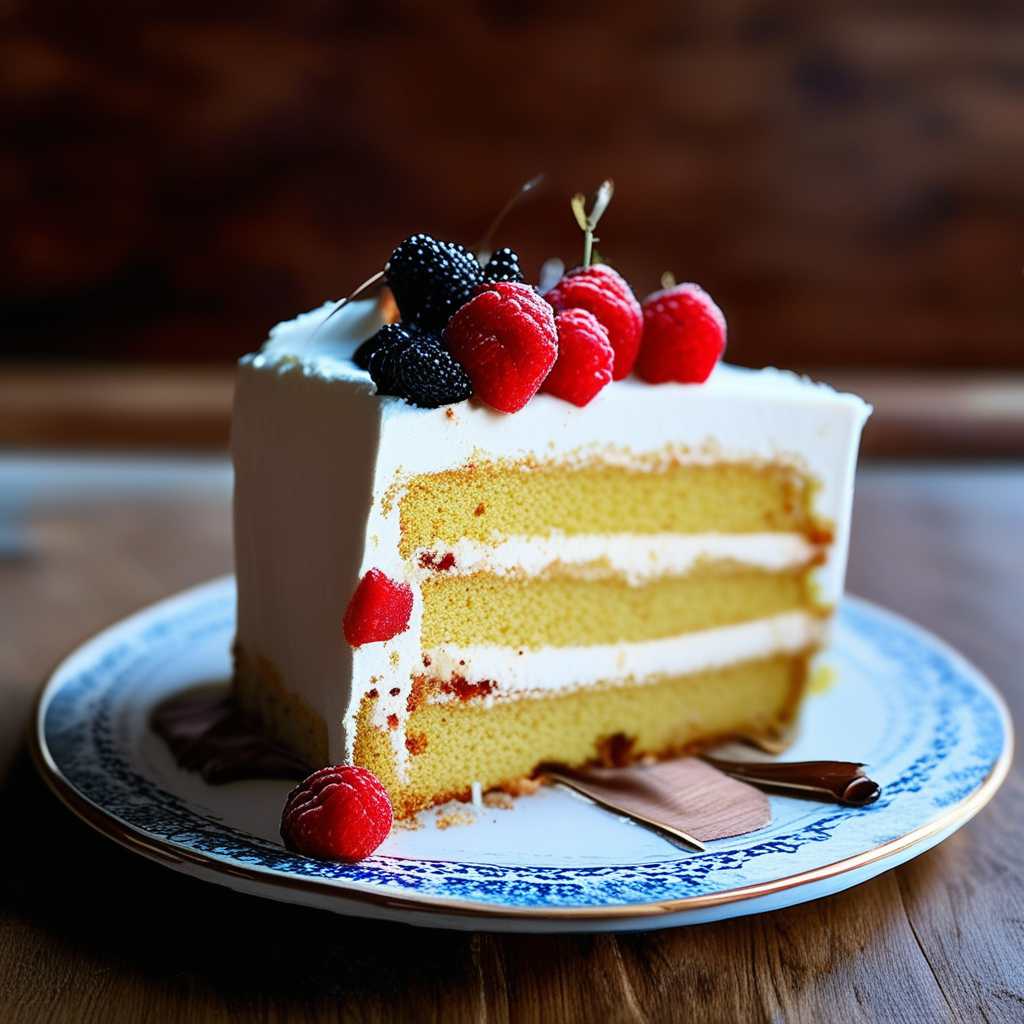}
        \caption{$p=0.5, w=7$, Cond}
    \end{subfigure}
    \begin{subfigure}[b]{0.24\textwidth}
        \includegraphics[width=1.0\textwidth]{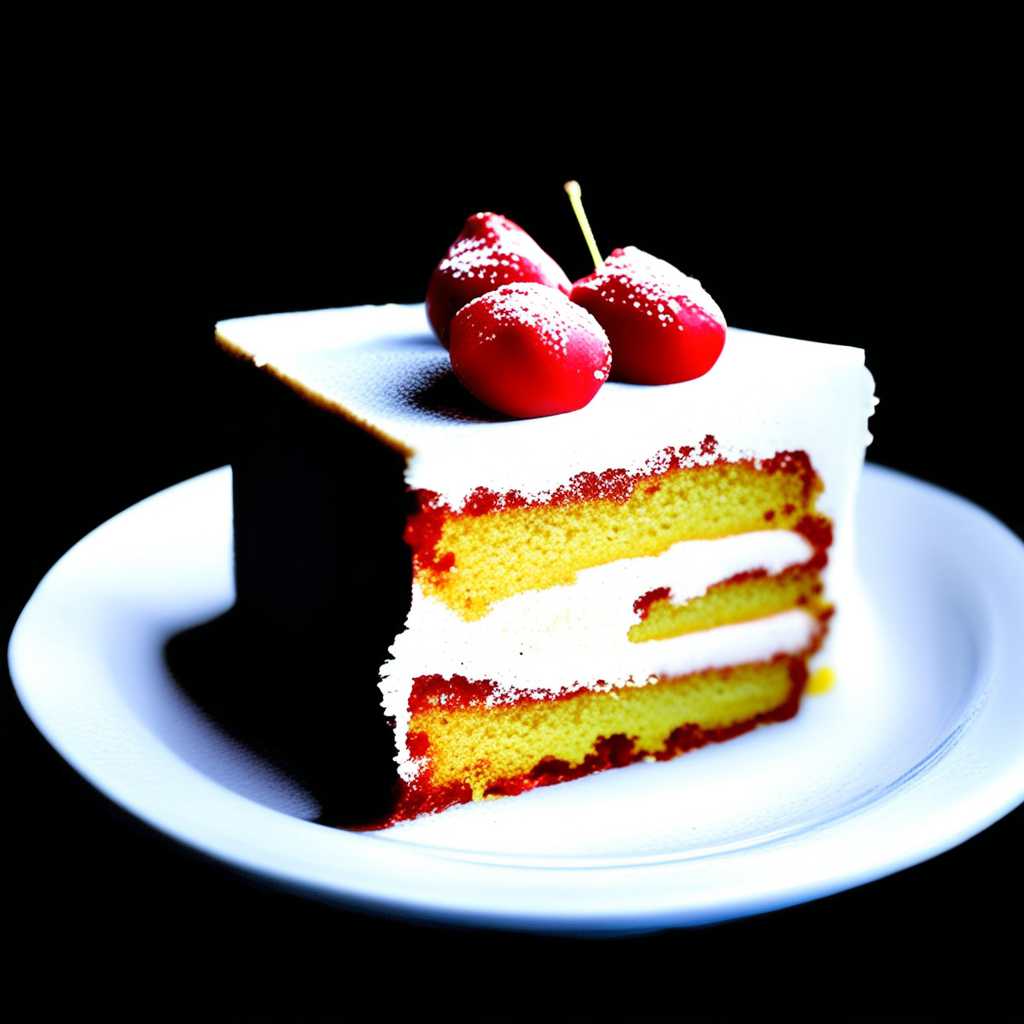}
        \caption{$p=0.5, w=15$, Cond}
    \end{subfigure}
    \begin{subfigure}[b]{0.24\textwidth}
        \includegraphics[width=1.0\textwidth]{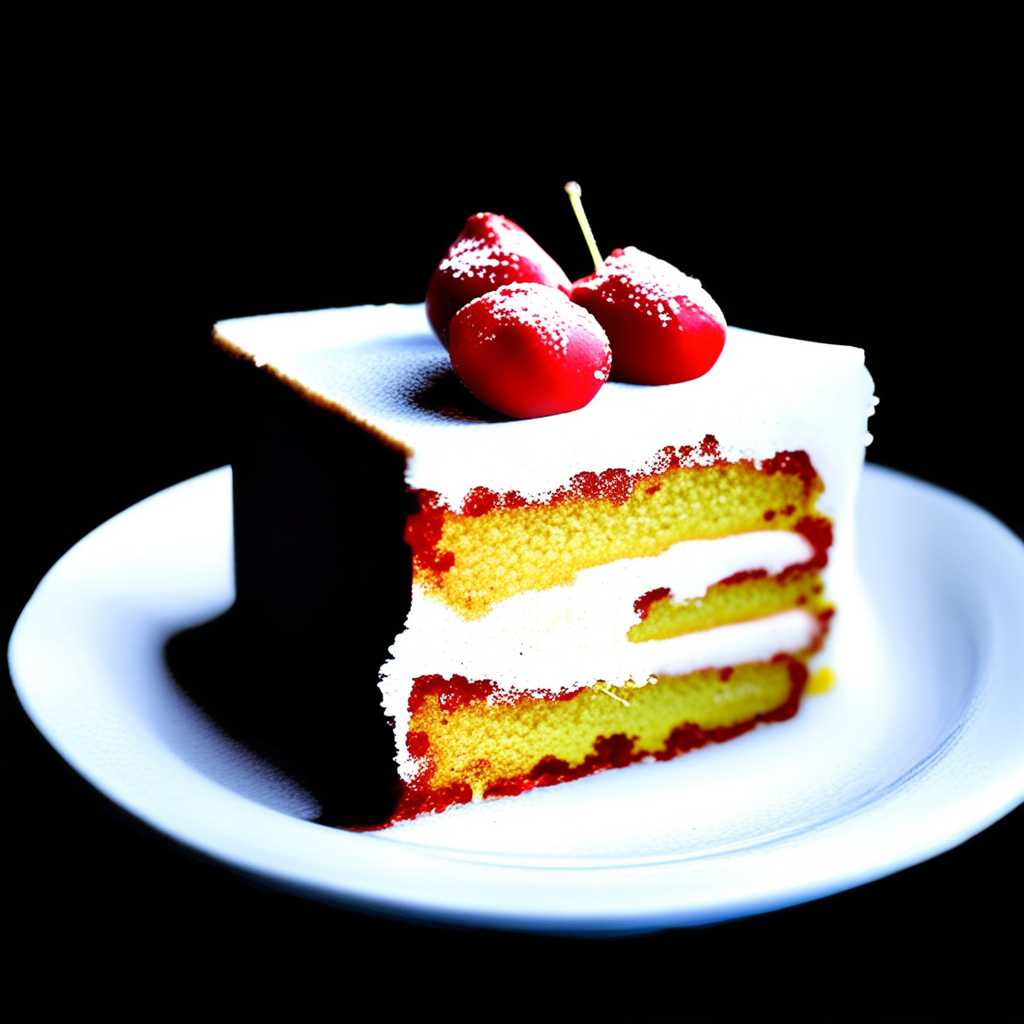}
        \caption{$p=0.5, w=15$, Uncond}
    \end{subfigure}
    \begin{subfigure}[b]{0.24\textwidth}
        \includegraphics[width=1.0\textwidth]{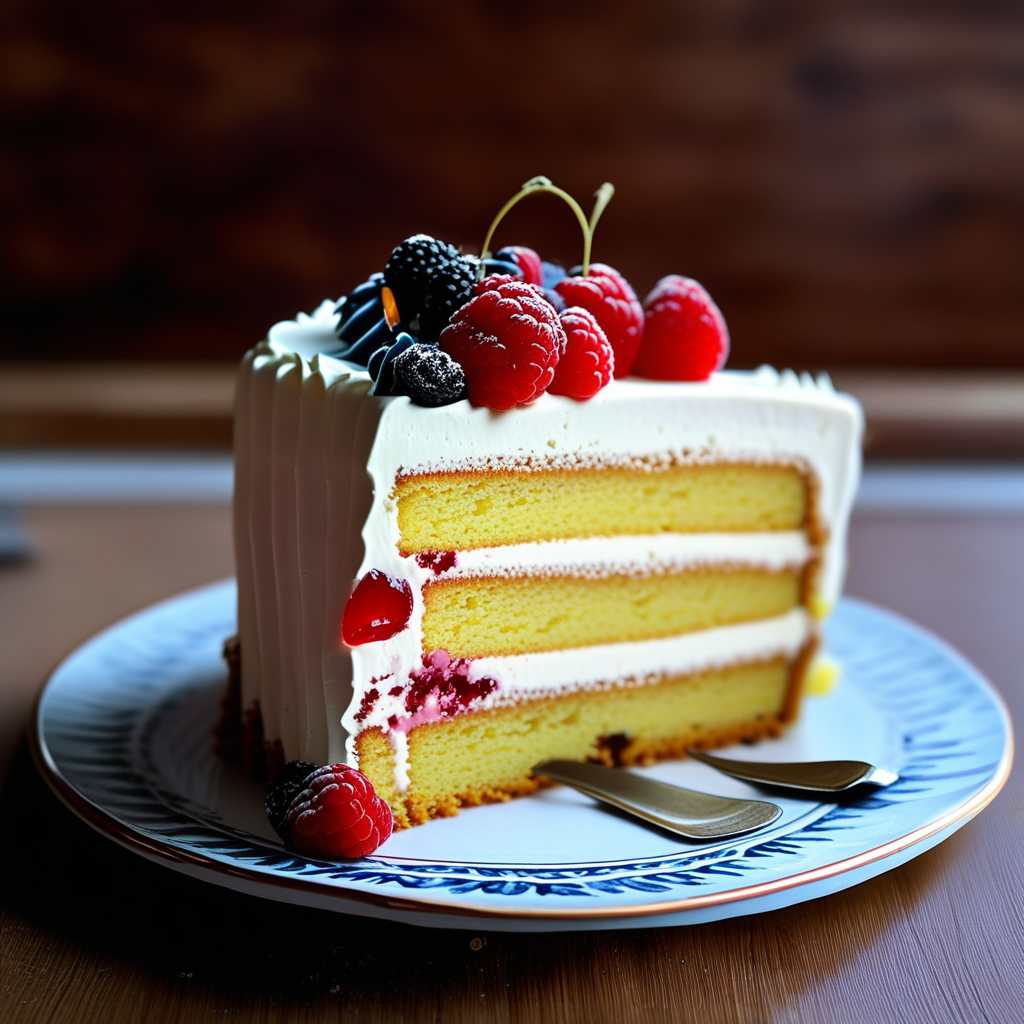}
        \caption{$p=1.0, w=7$}
    \end{subfigure}
    \begin{subfigure}[b]{0.24\textwidth}
        \includegraphics[width=1.0\textwidth]{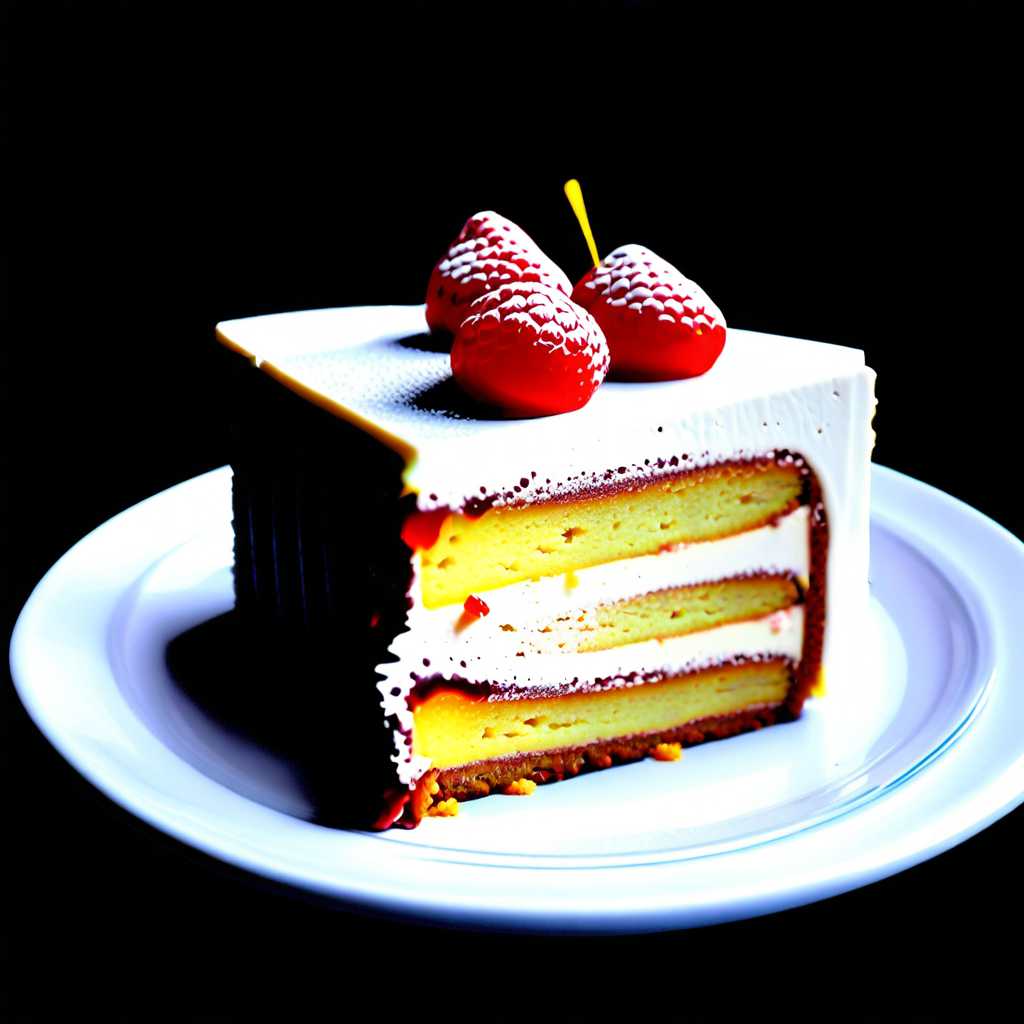}
        \caption{$p=1.0, w=15$}
    \end{subfigure}
    \caption{``A half eaten dessert cake sitting on a cake plate." generated by PixArt-$\Sigma$-XL under different settings. }
    \label{fig:pixart-case}
\end{figure*}

\begin{figure*}[htbp]
    \centering
    \begin{subfigure}[b]{1.0\textwidth}
        \includegraphics[width=0.135\textwidth]{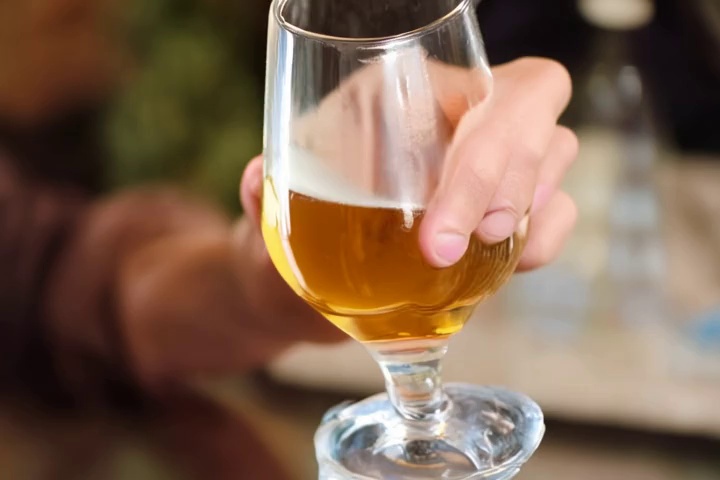}
        \includegraphics[width=0.135\textwidth]{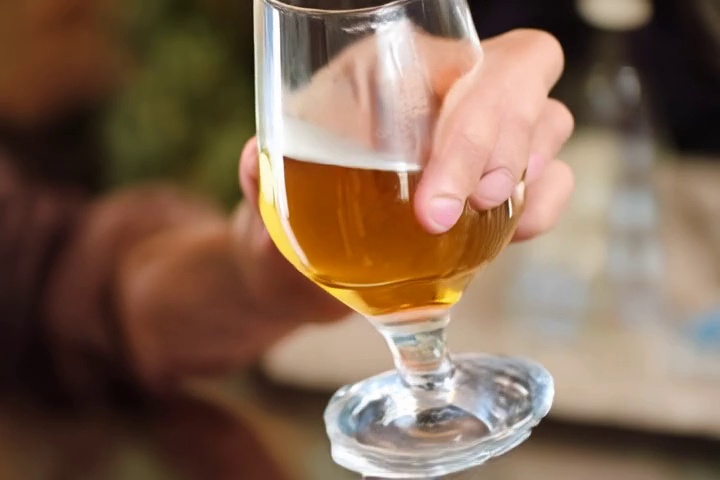}
        \includegraphics[width=0.135\textwidth]{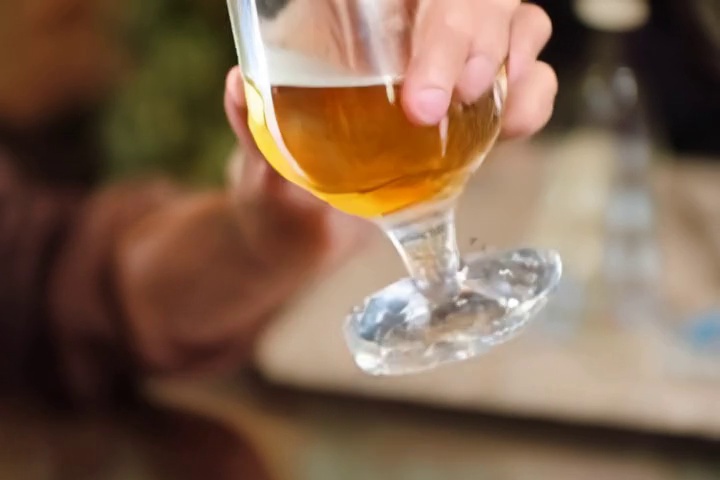}
        \includegraphics[width=0.135\textwidth]{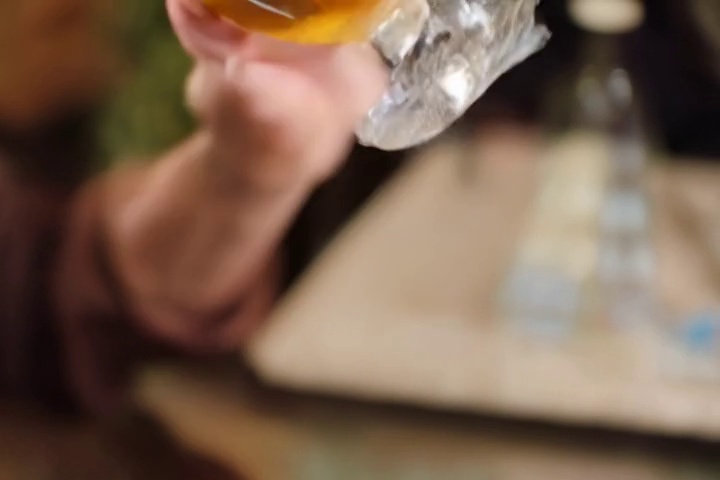}
        \includegraphics[width=0.135\textwidth]{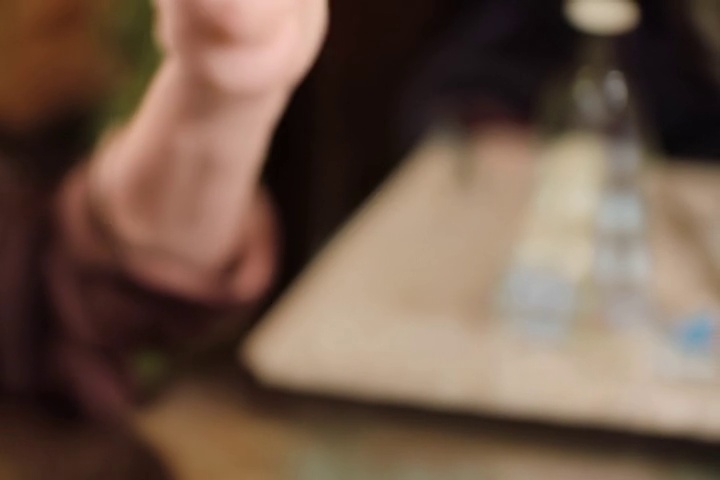}
        \includegraphics[width=0.135\textwidth]{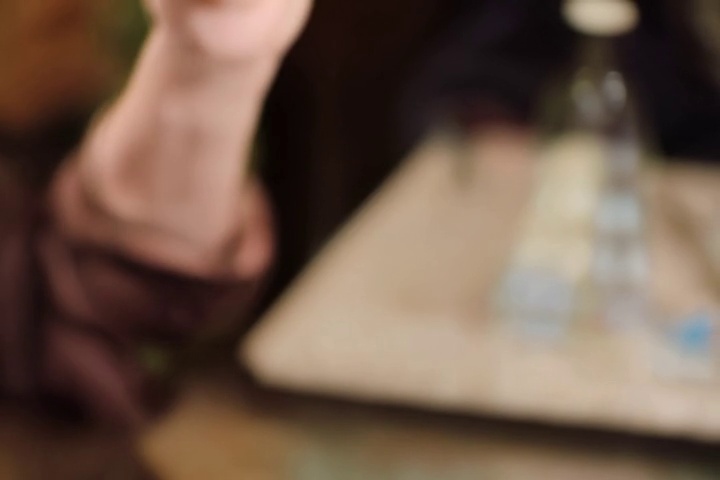}
        \includegraphics[width=0.135\textwidth]{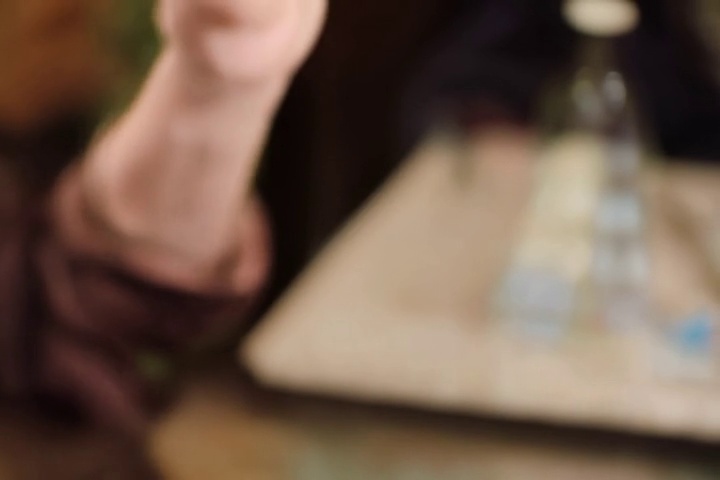}
        \caption{$p=0.3$, Cond}
    \end{subfigure}
    \begin{subfigure}[b]{1.0\textwidth}
        \includegraphics[width=0.135\textwidth]{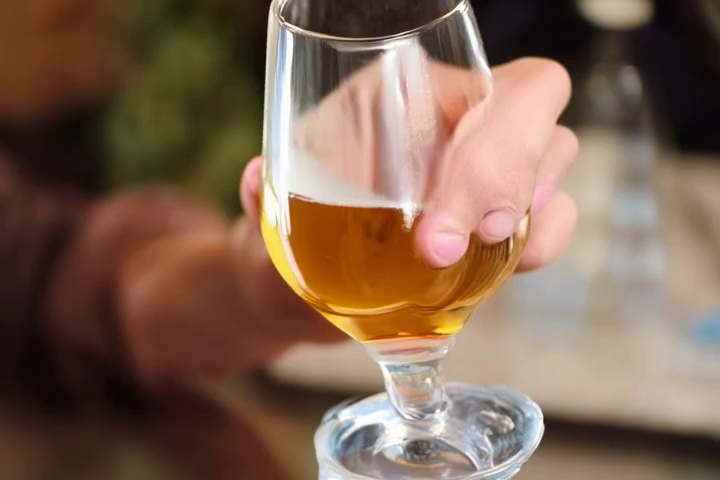}
        \includegraphics[width=0.135\textwidth]{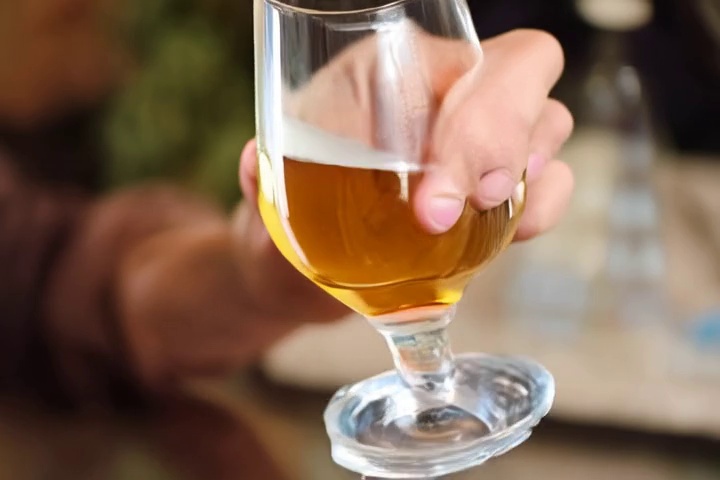}
        \includegraphics[width=0.135\textwidth]{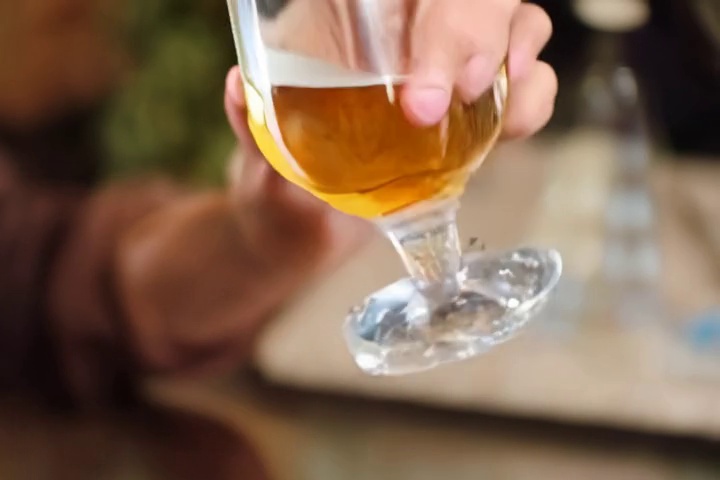}
        \includegraphics[width=0.135\textwidth]{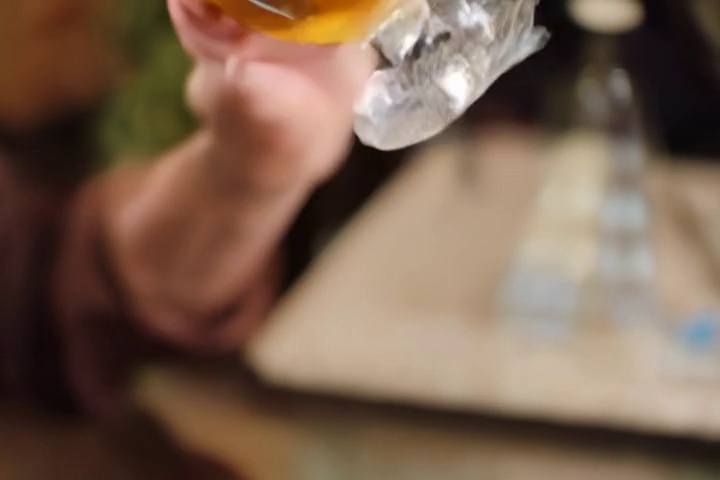}
        \includegraphics[width=0.135\textwidth]{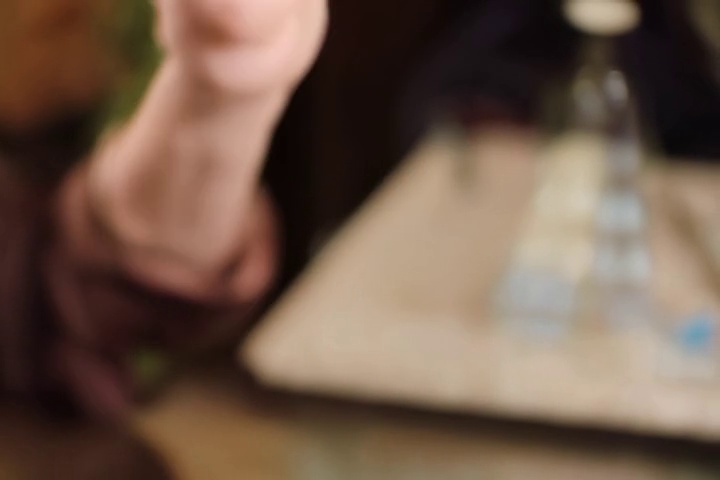}
        \includegraphics[width=0.135\textwidth]{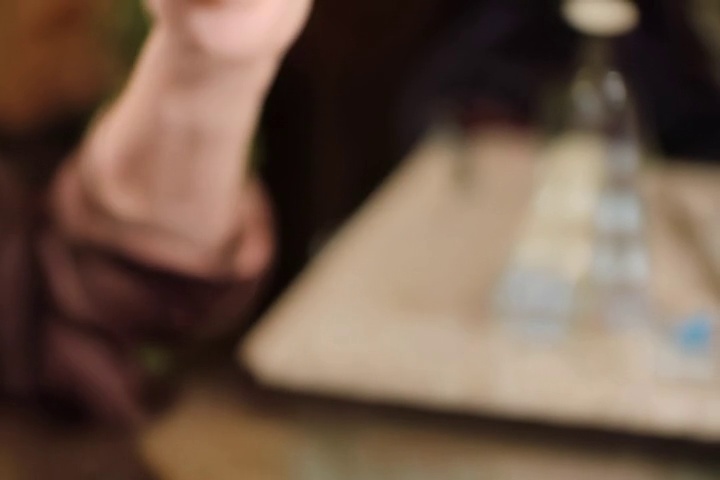}
        \includegraphics[width=0.135\textwidth]{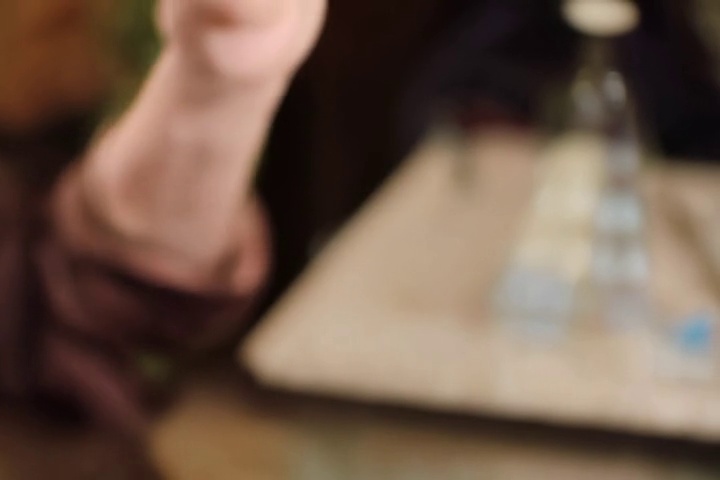}
        \caption{$p=0.3$, Uncond}
    \end{subfigure}
    \begin{subfigure}[b]{1.0\textwidth}
        \includegraphics[width=0.135\textwidth]{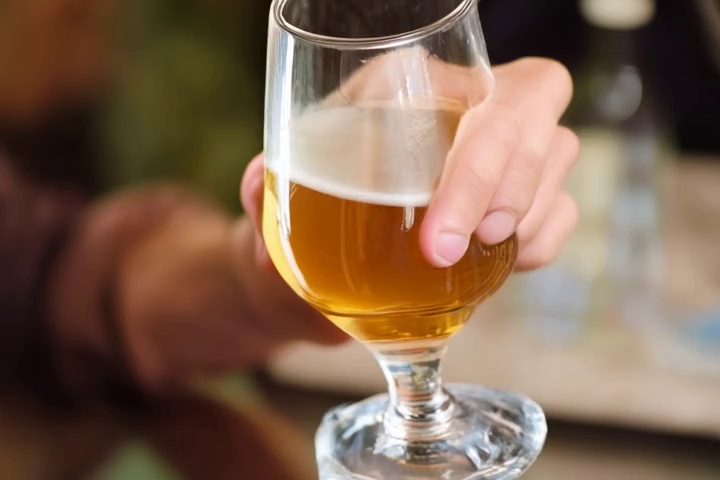}
        \includegraphics[width=0.135\textwidth]{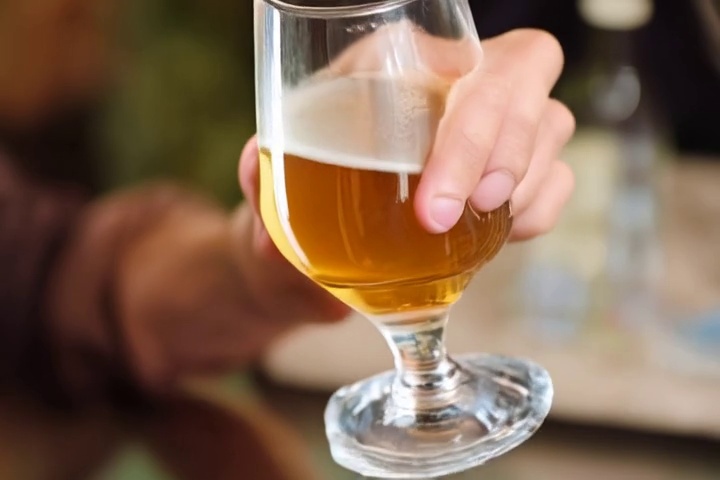}
        \includegraphics[width=0.135\textwidth]{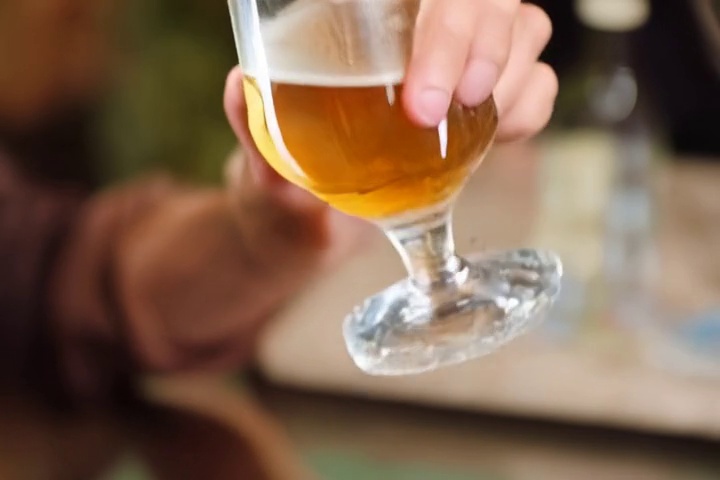}
        \includegraphics[width=0.135\textwidth]{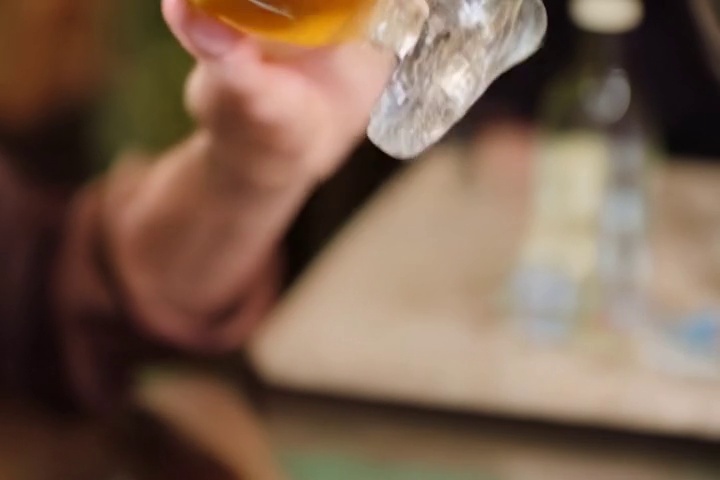}
        \includegraphics[width=0.135\textwidth]{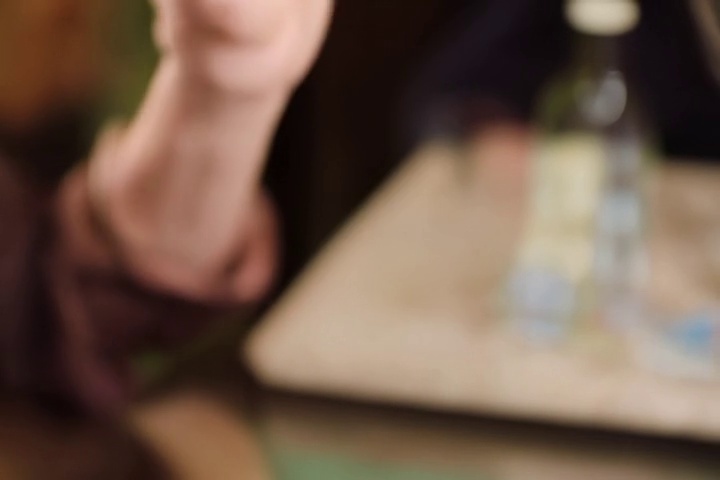}
        \includegraphics[width=0.135\textwidth]{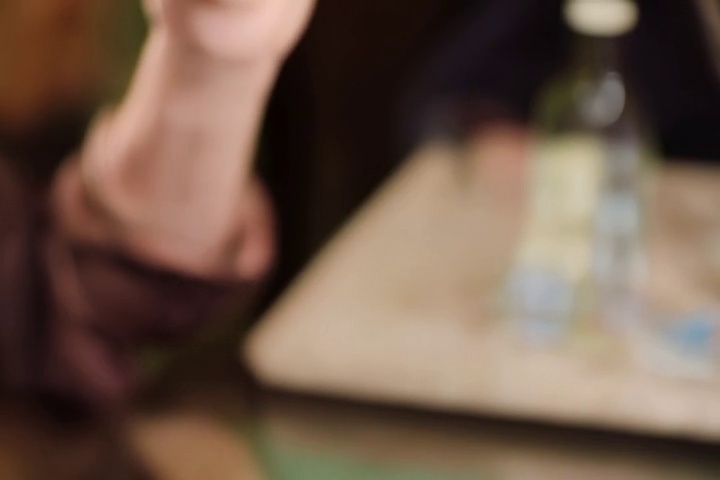}
        \includegraphics[width=0.135\textwidth]{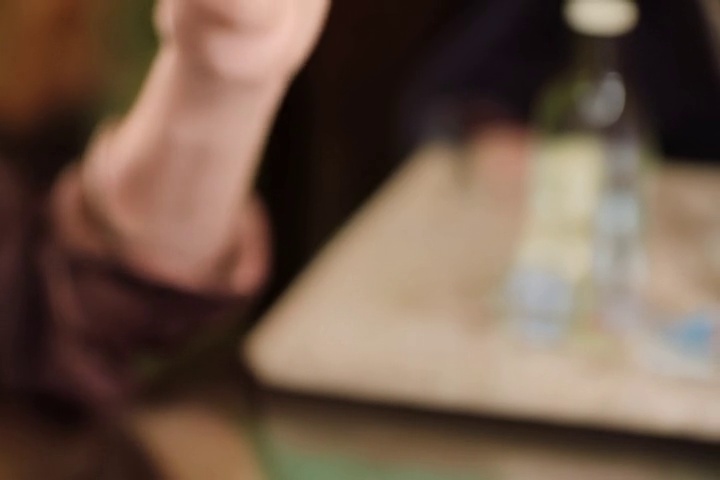}
        \caption{$p=0.5$, Cond}
    \end{subfigure}
    \begin{subfigure}[b]{1.0\textwidth}
        \includegraphics[width=0.135\textwidth]{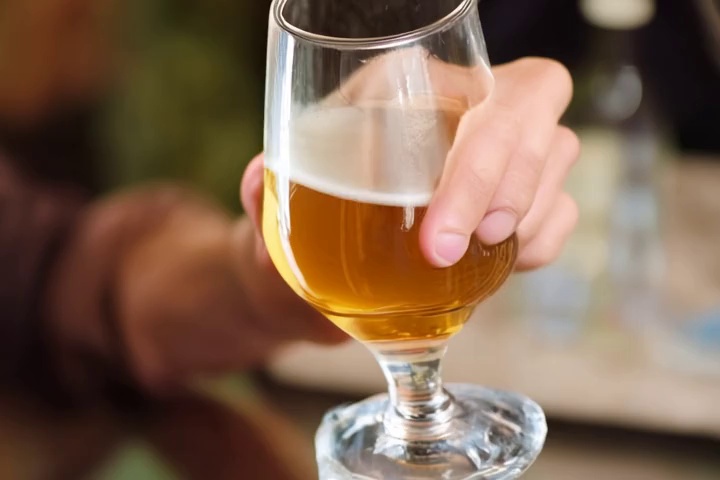}
        \includegraphics[width=0.135\textwidth]{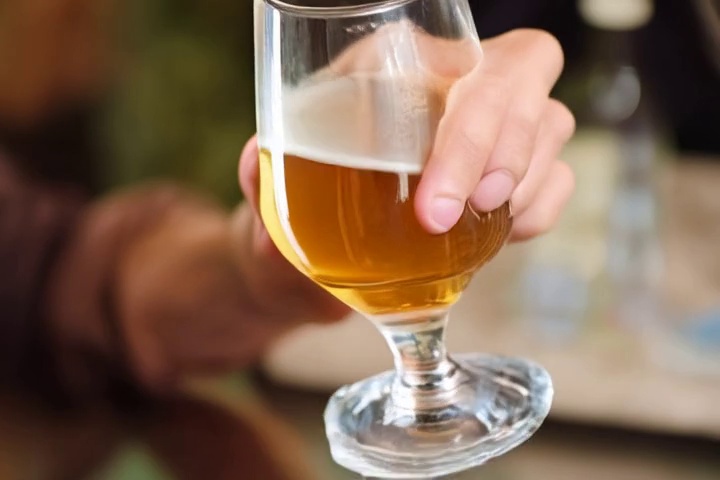}
        \includegraphics[width=0.135\textwidth]{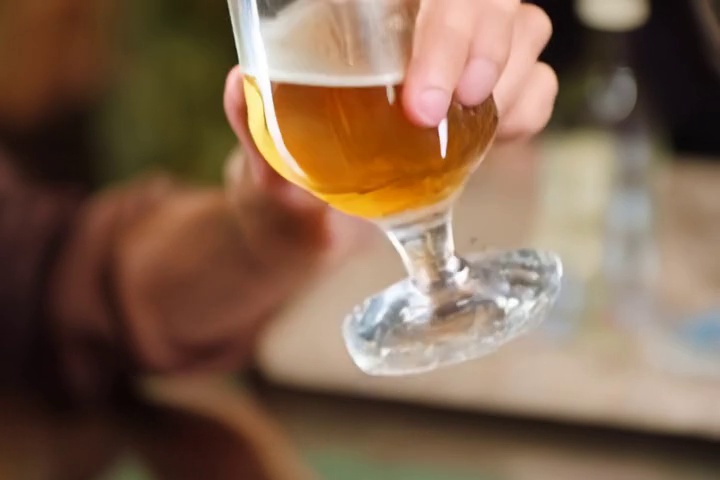}
        \includegraphics[width=0.135\textwidth]{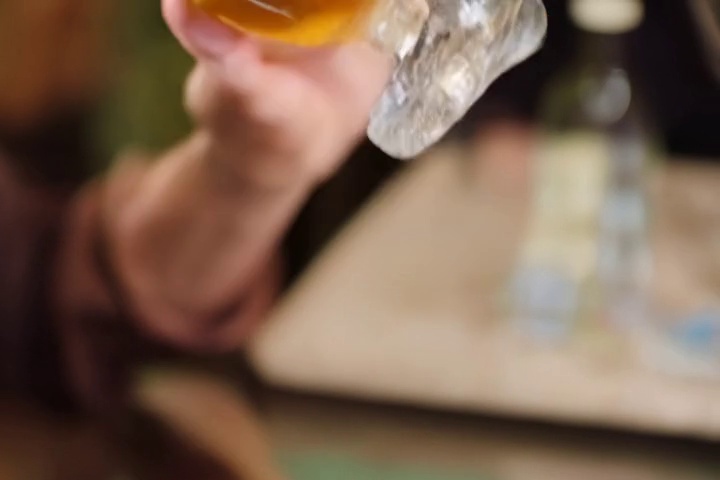}
        \includegraphics[width=0.135\textwidth]{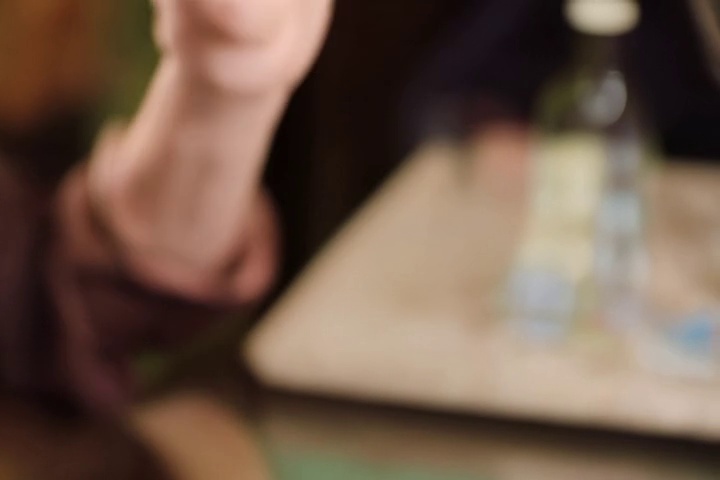}
        \includegraphics[width=0.135\textwidth]{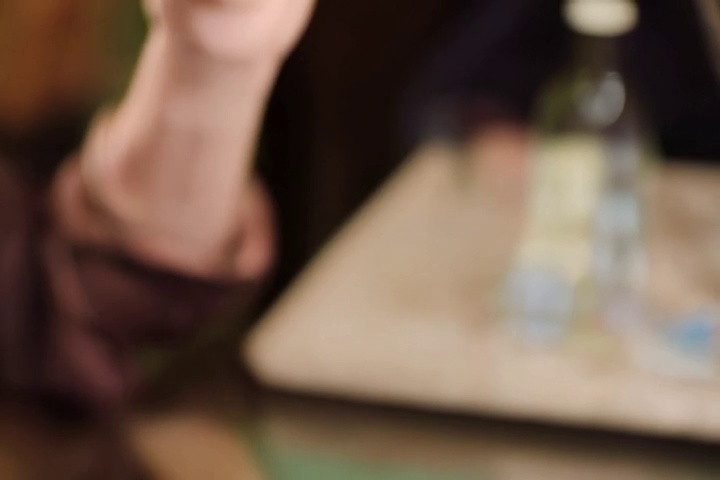}
        \includegraphics[width=0.135\textwidth]{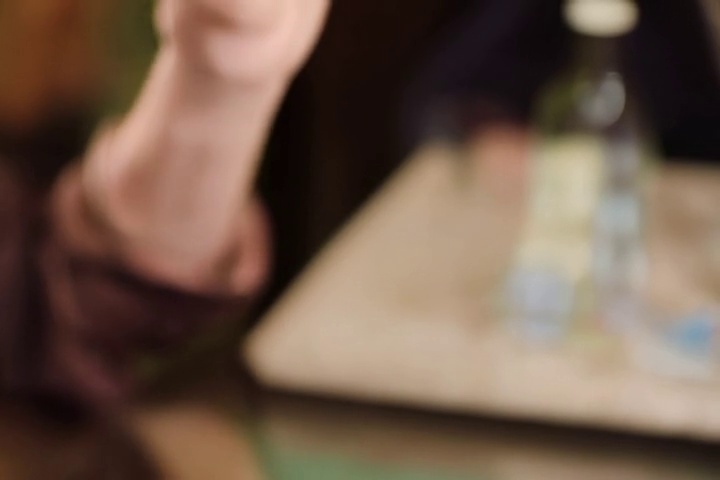}
        \caption{$p=0.5$, Uncond}
    \end{subfigure}
    \begin{subfigure}[b]{1.0\textwidth}
        \includegraphics[width=0.135\textwidth]{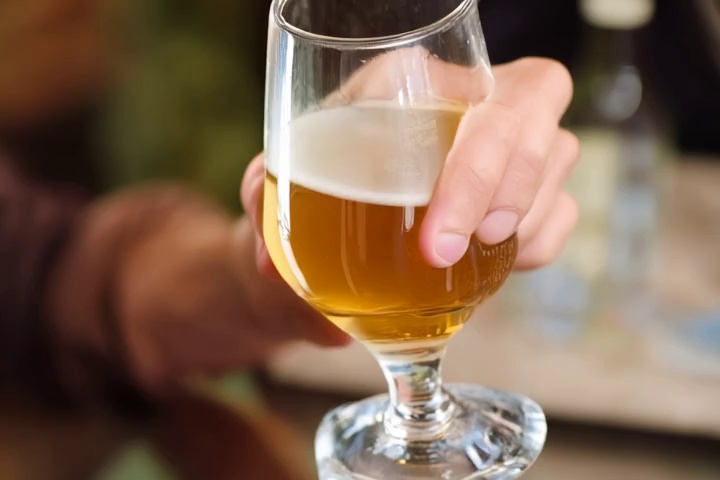}
        \includegraphics[width=0.135\textwidth]{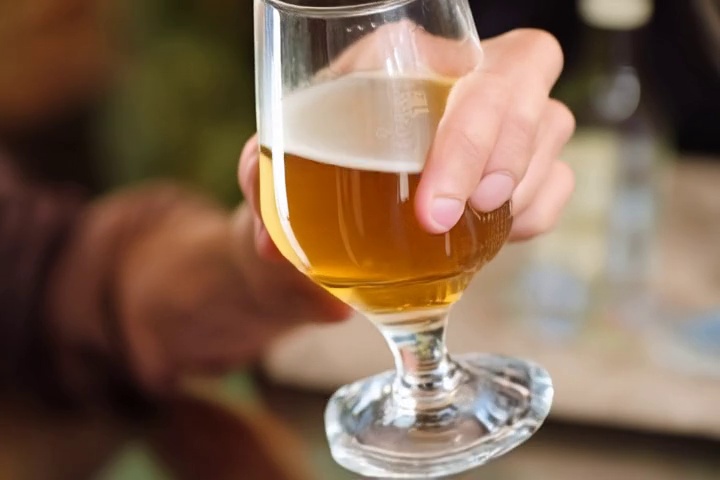}
        \includegraphics[width=0.135\textwidth]{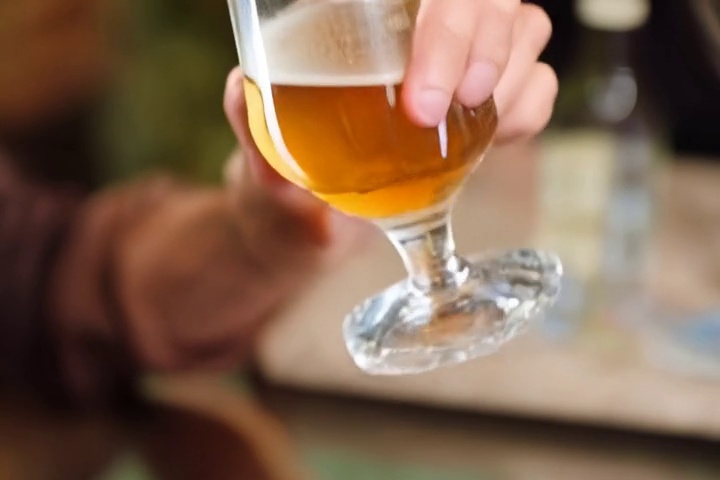}
        \includegraphics[width=0.135\textwidth]{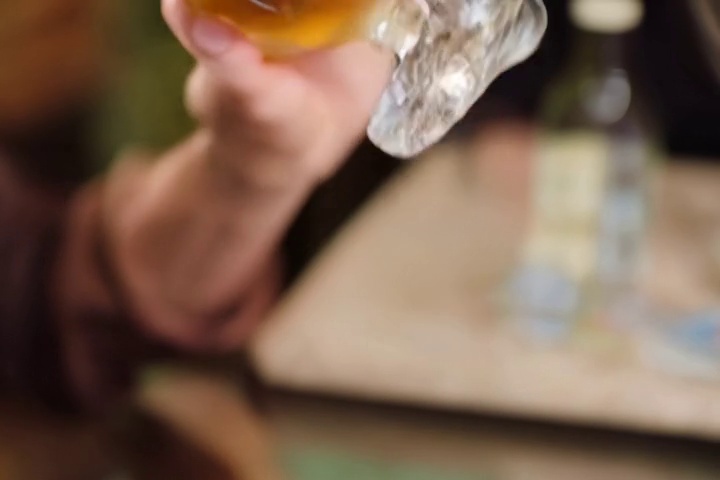}
        \includegraphics[width=0.135\textwidth]{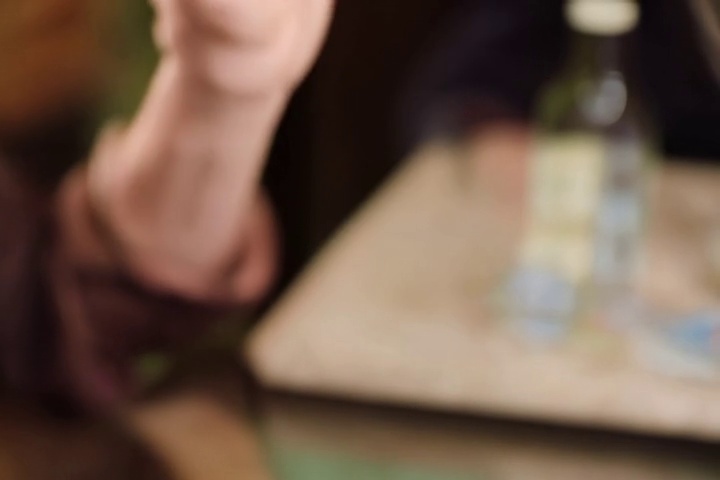}
        \includegraphics[width=0.135\textwidth]{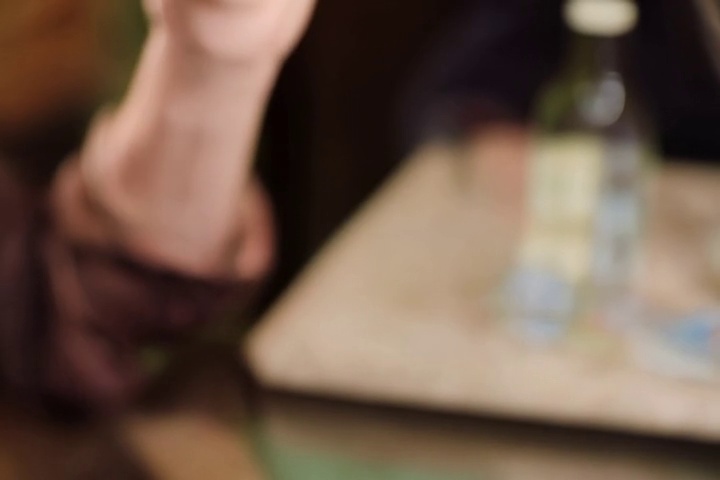}
        \includegraphics[width=0.135\textwidth]{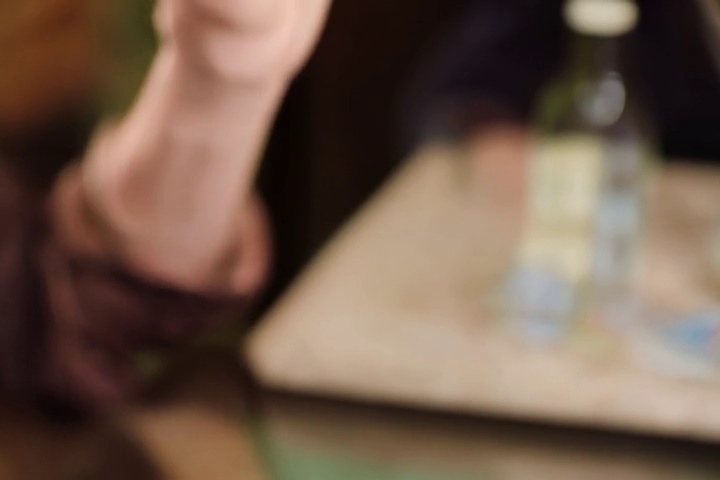}
        \caption{$p=1.0$}
    \end{subfigure}
    \caption{``A person is tasting beer" generated by CogVideoX under different settings. }
    \label{fig:cog-case}
\end{figure*}

\begin{figure*}[htbp]
    \centering
    \begin{subfigure}[b]{1.0\textwidth}
        \includegraphics[width=0.12\textwidth]{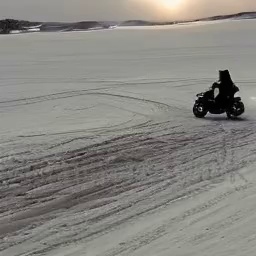}
        \includegraphics[width=0.12\textwidth]{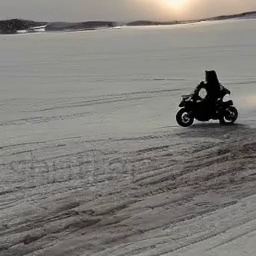}
        \includegraphics[width=0.12\textwidth]{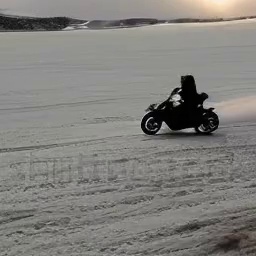}
        \includegraphics[width=0.12\textwidth]{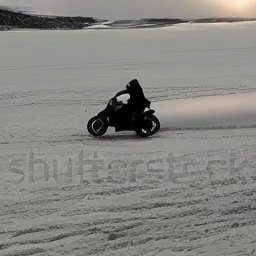}
        \includegraphics[width=0.12\textwidth]{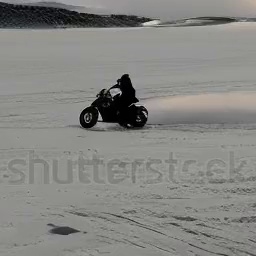}
        \includegraphics[width=0.12\textwidth]{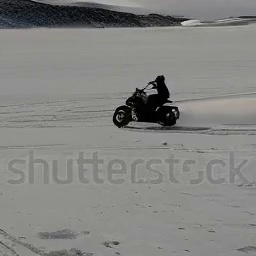}
        \includegraphics[width=0.12\textwidth]{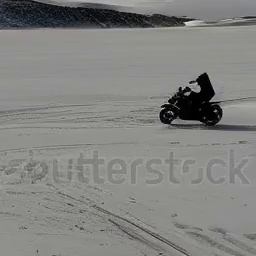}
        \includegraphics[width=0.12\textwidth]{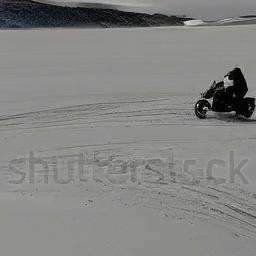}
        \caption{$p=0.3, w=6$, Cond}
    \end{subfigure}
    \begin{subfigure}[b]{1.0\textwidth}
        \includegraphics[width=0.12\textwidth]{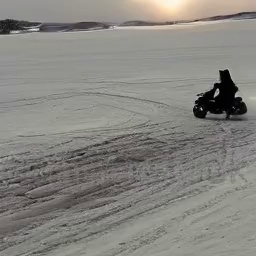}
        \includegraphics[width=0.12\textwidth]{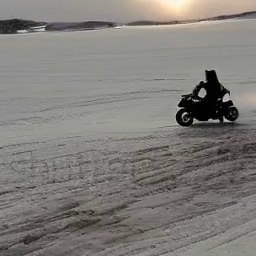}
        \includegraphics[width=0.12\textwidth]{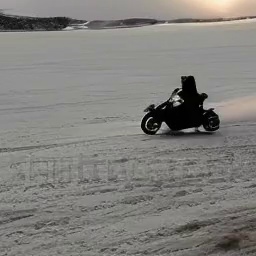}
        \includegraphics[width=0.12\textwidth]{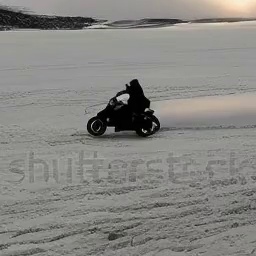}
        \includegraphics[width=0.12\textwidth]{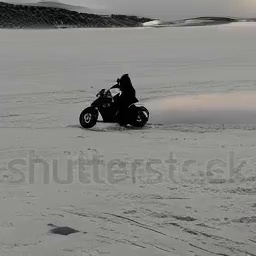}
        \includegraphics[width=0.12\textwidth]{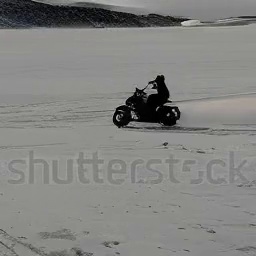}
        \includegraphics[width=0.12\textwidth]{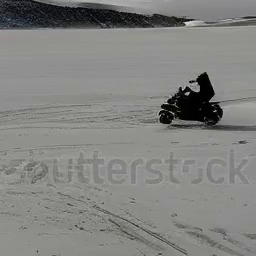}
        \includegraphics[width=0.12\textwidth]{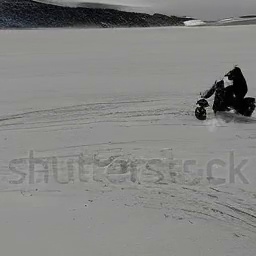}
        \caption{$p=0.3, w=6$, Uncond}
    \end{subfigure}
    \begin{subfigure}[b]{1.0\textwidth}
        \includegraphics[width=0.12\textwidth]{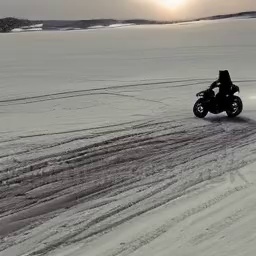}
        \includegraphics[width=0.12\textwidth]{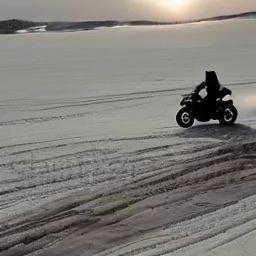}
        \includegraphics[width=0.12\textwidth]{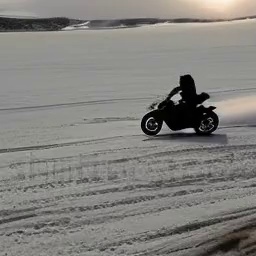}
        \includegraphics[width=0.12\textwidth]{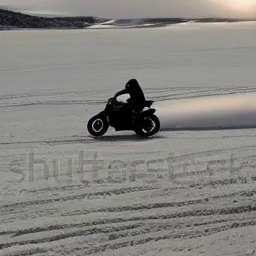}
        \includegraphics[width=0.12\textwidth]{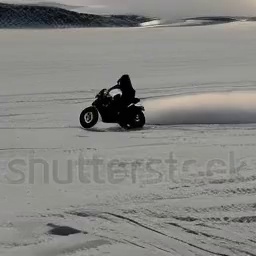}
        \includegraphics[width=0.12\textwidth]{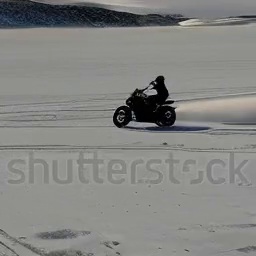}
        \includegraphics[width=0.12\textwidth]{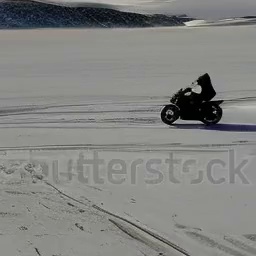}
        \includegraphics[width=0.12\textwidth]{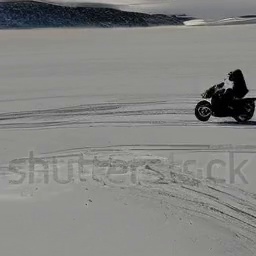}
        \caption{$p=0.5, w=6$, Cond}
    \end{subfigure}
    \begin{subfigure}[b]{1.0\textwidth}
        \includegraphics[width=0.12\textwidth]{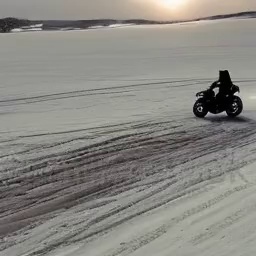}
        \includegraphics[width=0.12\textwidth]{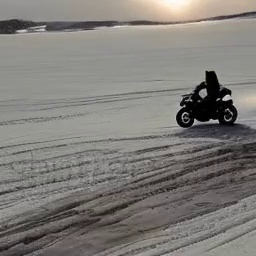}
        \includegraphics[width=0.12\textwidth]{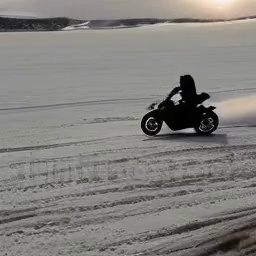}
        \includegraphics[width=0.12\textwidth]{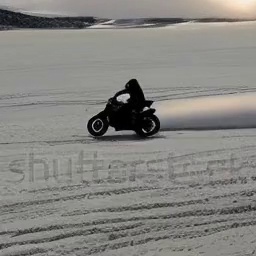}
        \includegraphics[width=0.12\textwidth]{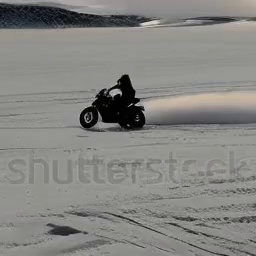}
        \includegraphics[width=0.12\textwidth]{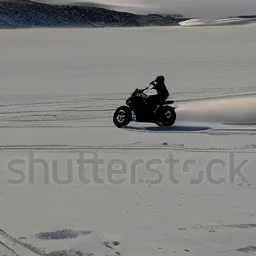}
        \includegraphics[width=0.12\textwidth]{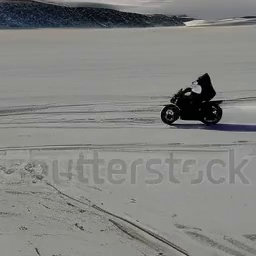}
        \includegraphics[width=0.12\textwidth]{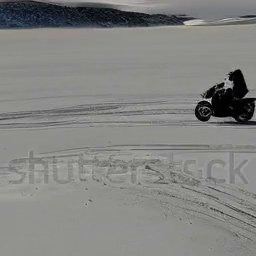}
        \caption{$p=0.5, w=6$, Uncond}
    \end{subfigure}
    \begin{subfigure}[b]{1.0\textwidth}
        \includegraphics[width=0.12\textwidth]{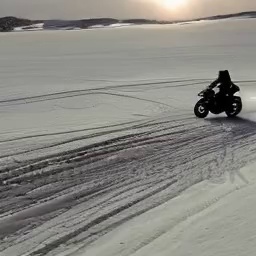}
        \includegraphics[width=0.12\textwidth]{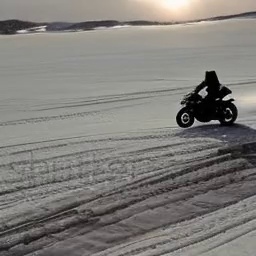}
        \includegraphics[width=0.12\textwidth]{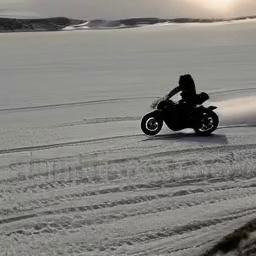}
        \includegraphics[width=0.12\textwidth]{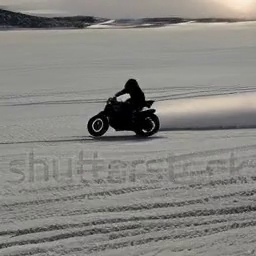}
        \includegraphics[width=0.12\textwidth]{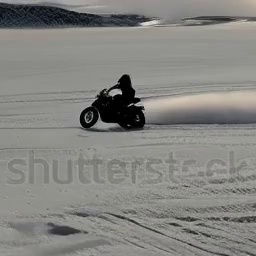}
        \includegraphics[width=0.12\textwidth]{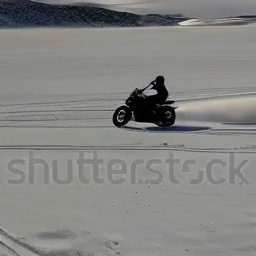}
        \includegraphics[width=0.12\textwidth]{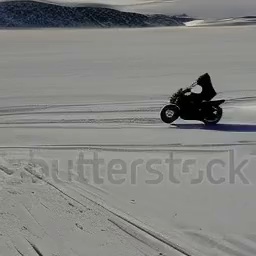}
        \includegraphics[width=0.12\textwidth]{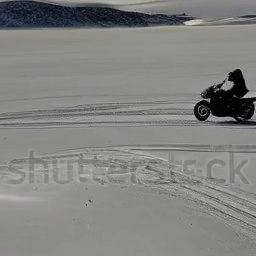}
        \caption{$p=1.0, w=6$, Uncond}
    \end{subfigure}
    \caption{``A motorcycle gliding through a snowy field" generated by ModelScope under different settings. }
    \label{fig:ms-case}
\end{figure*}

We would like to provide more qualitative results in Figure \ref{fig:sd15-case}, \ref{fig:sdxl-case}, \ref{fig:pixart-case}, \ref{fig:cog-case}, \ref{fig:ms-case}. 

\end{document}